%% file: main.tex
\documentclass[Afour,sageh,times]{sagej}
\usepackage{amsmath,amssymb,bm,booktabs,array,multirow,adjustbox,microtype,etoolbox,placeins}
\usepackage[colorlinks=true,citecolor=blue,linkcolor=blue,urlcolor=blue]{hyperref}
\allowdisplaybreaks
\usepackage{booktabs}
\usepackage{graphicx}
\usepackage{xcolor}
\usepackage{pifont}
\usepackage{array}
\usepackage{tabularx}
\usepackage{amsthm}
\usepackage{verbatim}
\usepackage{pifont}
\newcommand{\circled}[1]{%
	\textcircled{\raisebox{-0.20ex}{\scriptsize #1}}%
}
\newtheorem{Remark}{Remark}
\providecommand{\tabyes}{\ding{51}}
\providecommand{\tabno}{\ding{55}}

\usepackage{bm}
\def\journalname{xx}
\def\volumenumber{XX}
\def\issuenumber{X}
\def\startpage{1}
\def\endpage{30}

\makeatletter
\patchcmd{\ps@sagepage}{\volumenumber(\issuenumber)}{}{}{}
\makeatother

	% ...

\begin{document}
	\pagestyle{sagepage}
	\runninghead{Wang et al.}
	\title{CAPTURER: A Dexterous Aerial Manipulator for Dynamic and Precise Grasping across Diverse Environments}
	\author{Meng Wang\affilnum{1,2}, Jindou Jia\affilnum{3}, Sicheng Zhou\affilnum{4}, Zeshuai Chen\affilnum{1}, Kexin Guo\affilnum{5}, Xiang Yu\affilnum{1,6} and Lei Guo\affilnum{1,6}}
	\affiliation{\affilnum{1}School of Automation Science and Electrical Engineering, Beihang University, Beijing, China\\
		\affilnum{2}School of Electrical and Electronic Engineering, Nanyang Technological University, Singapore\\
		\affilnum{3}School of Mechanical and Aerospace Engineering, Nanyang Technological University, Nanyang, Singapore\\
		\affilnum{4}Department of Electrical and Computer Engineering, National University of Singapore, Singapore 119077, Singapore\\
		\affilnum{5}School of Aeronautic Science and Engineering, Beihang University, Beijing, China\\
		\affilnum{6}Beijing Key Laboratory of Bio-inspired Intelligent Technology of Unmanned System, China}
	\corrauth{Xiang Yu and Lei Guo, Beihang University, Beijing 100191, China.}
	\email{xiangyu\_buaa@buaa.edu.cn; lguo@buaa.edu.cn}
	\begin{abstract}
		As a representative form of embodied robotic interaction, aerial grasping requires close coordination between the aerial platform and manipulator. Most existing systems perform aerial grasping with static or slowly moving targets under fixed scenarios. Dynamic and precise grasping across different conditions introduces additional challenges, including target motion, localization drift, dynamic coupling, and external disturbances. In this work, we present CAPTURER, a dexterous aerial manipulator for centimeter-level grasping across diverse environments. To achieve the challenging goal, we first develop an onboard perception system for state estimation and target detection. Drift correction and global pose transformation are introduced to improve target localization during flight. A cooperative motion planner is proposed to coordinate the mobility of the aerial platform with the dexterity of the manipulator according to the relative target state. To maintain grasping accuracy under disturbances, we design a disturbance-rejection controller to compensate for dynamic coupling, model uncertainty, and external disturbances. The capability of CAPTURER is validated through indoor and outdoor experiments with both static and moving targets. The results demonstrate that CAPTURER achieves centimeter-level grasping accuracy for a marker pen measuring only 1.5 cm in diameter at flight speeds exceeding 1.5 m/s.
	\end{abstract}
	\keywords{Aerial manipulation, onboard perception, drift correction, cooperative planning,  robust control, dynamic and precise grasping}
	\maketitle
	
	\input{sections/01_introduction}
	\input{sections/02_system}
	\input{sections/03_perception}
	\input{sections/04_planning}
	\input{sections/05_control}
	\input{sections/07_grasping}
	\FloatBarrier
	\input{sections/08_discussion}

	\bibliographystyle{SageH}
	\bibliography{references}
\end{document}

%% file: sections/01_introduction.tex
\section{Introduction}
\subsection{Background}
Aerial robots have evolved from laboratory prototypes into increasingly capable platforms for autonomous operation in complex environments \citep{SR_PHRI}. Their high mobility has enabled applications such as autonomous drone racing \citep{limit}, navigation through dense forests \citep{Swarm}, and resilient flight under adverse weather conditions \citep{nerual_fly}. Despite these advances, most aerial robots still primarily serve as sensing and observation platforms. Performing physical interaction remains considerably more challenging, particularly when the task requires both fast motion and precise contact with the external environment.

Aerial manipulators combine the large-range mobility of unmanned aerial vehicles (UAVs) with the local operability of manipulators, enabling physical interactions in difficult-to-access and hazardous environments \citep{ado6251,BONYANKHAMSEH2018221,9462539,nature,Proximal}. Early studies have demonstrated aerial interaction using grippers mounted on multirotor platforms \citep{pioneer,Kumar2014}. Soft and compliant grippers are subsequently developed to enlarge the effective contact region and increase grasp tolerance for different target geometries \citep{9981668,npjrobotics}. Specialized systems have also achieved millimeter-level accuracy in aerial insertion tasks  \citep{10339889}. In addition, coordinated multi-robot systems have demonstrated aerial construction and manipulation over extended workspaces \citep{nature}.

However, most existing aerial systems operate under quasi-static or near-hovering conditions. Moreover, the grasping motion is often predefined or generated offline. The operating mode does not fully exploit the mobility and agility of the aerial platform during physical interaction. Many systems also rely on external localization infrastructure to maintain accurate positioning. Vision-based approaches reduce this dependence, but they commonly assume that the target is always visible within the camera's field-of-view (FoV) during operation \citep{7989751,9789094,9781617}. These assumptions are restrictive when the target moves or when the robot operates across different environments.

Other approaches use deformable or morphing aerial platforms for grasping and interaction \citep{8594330, 9196576, 10495172}. These systems exploit the vehicle body itself as part of the manipulation mechanism. However, the achievable interaction workspace and local adjustment capability are constrained by the airframe structure. Specialized mechanisms, such as permanent-magnet grippers \citep{10224341}  and delta-type manipulators \citep{10083097}, have also demonstrated dynamic grasping. In \citet{npjrobotics}, dynamic grasping of a large box is achieved using onboard cameras, while a soft gripper is employed to enlarge the effective contact area and improve grasp tolerance. However, many of these studies rely on task-specific mechanisms, relatively large targets, or predefined motion trajectories. The ability to adapt online to target motion, localization uncertainty, dynamic coupling, and environmental disturbances remains limited.

Dynamic and precise aerial grasping therefore requires more than increasing the speed of an existing aerial manipulation system. During a grasping maneuver, the target state may change continuously, while the aerial platform and manipulator move simultaneously. The robot must estimate the target state onboard, determine how motion should be distributed between the platform and manipulator, and accurately execute the resulting motion under internal and external disturbances. These requirements are strongly coupled, which become more demanding as the interaction speed increases.

In this study, our objective is to enable an aerial manipulator to autonomously perform precise grasping of static and moving targets across different operating environments. Autonomous operation is achieved without relying on external localization. The aerial platform provides large-range mobility for target approach, while the manipulator supplies local dexterity for precise adjustment. Meanwhile, the control framework maintains grasping accuracy in the presence of composite disturbances.
\begin{figure*}[!tp]
	\centering
	\includegraphics[width=0.92\linewidth,height=\textheight,keepaspectratio]{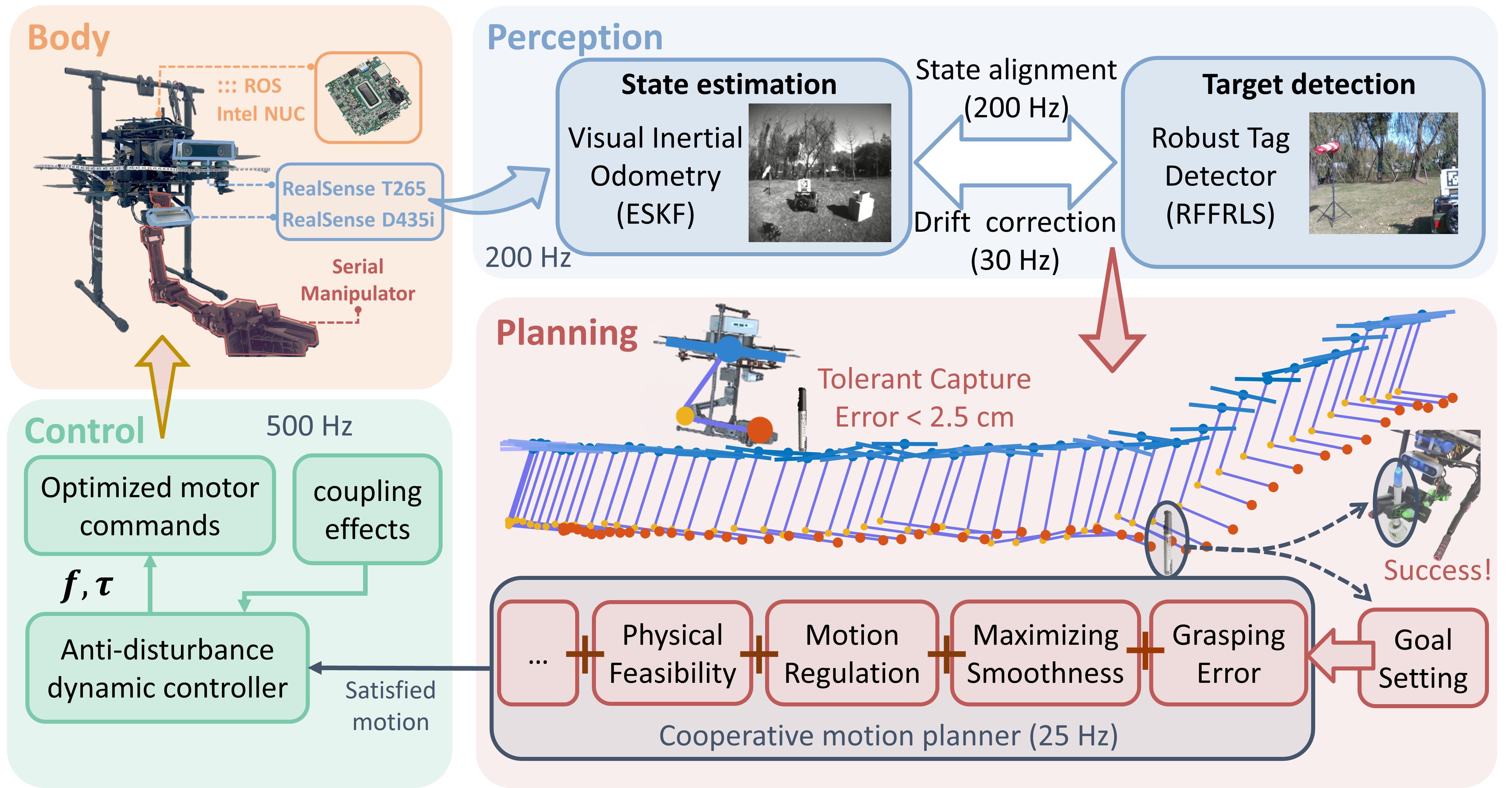}
	\caption{ Overview of CAPTURER workflow.  
		\textbf{Body}: CAPTURER integrates onboard perception, cooperative motion planning, and anti-disturbance control within a unified framework. 
		\textbf{Perception}: The perception module combines visual--inertial state estimation, target detection, drift correction, global pose transformation, motion prediction, and active target search to provide consistent robot and target states. 
		\textbf{Planning}: Based on the relative target state, the cooperative planner generates coordinated motion references for the aerial platform and manipulator while accounting for the predicted platform attitude and their different motion characteristics.
		\textbf{Control}: The dynamic controller reconstructs and compensates the strong coupling effects induced by manipulator motion and employs residual observers to handle model uncertainty and external disturbances. 
		The resulting framework supports indoor and outdoor grasping of stationary and moving targets using onboard sensing and online replanning. 
		More details, including CAD models and source code, are available on the \href{https://sites.google.com/view/capturer/\%E9\%A6\%96\%E9\%A1\%B5}{Project Site}.
		}\label{IJRR_framework}
\end{figure*}
\subsection{Problem description}
Dynamic and precise aerial grasping introduces three closely coupled technical problems.

The first problem is \textit{\textbf{perception}}. Precise grasping requires consistent estimation of both the aerial manipulator state and the target state. Unlike conventional aerial navigation, the manipulator moves relative to the floating base during operation. This motion changes the system configuration and induces additional disturbances to the aerial platform. Meanwhile, onboard localization is subject to accumulated drift. The problem becomes more difficult for a moving target because the perception system must continuously update its position while the robot itself is moving \citep{BONYANKHAMSEH2018221,9462539}. When the target leaves the FoV, measurement updates become unavailable and prediction uncertainty increases. Moreover, dynamic grasping requires a consistent target representation that remains usable by the motion planner during platform and manipulator motion.

The second problem is \textit{\textbf{planning}}. The aerial platform provides large-range mobility, whereas the manipulator provides more precise local adjustment \citep{9462539,Kim_IJRR}. Different combinations of platform and manipulator motions can satisfy the same end-effector requirement, while resulting in different remaining workspaces for subsequent adjustment. The cooperative planner must allocate motion according to their respective characteristics and the relative target state. Moreover, the attitude information of the aerial platform should be incorporated into the planning process because it directly affects the end-effector state.  

The third problem is \textit{\textbf{control}}. The manipulator motion causes coupling effects that act on both the moving platform and manipulator dynamics \citep{shabana2020dynamics,9462539}. Moreover, the coupling effects become more pronounced during dynamic operation and can significantly degrade tracking accuracy. In addition, model uncertainty and external disturbances would deteriorate the flight performance of the aerial manipulator, ultimately reducing grasping accuracy \citep{nature,Proximal}. 

Notice that the aforementioned problems are not independent. Perception provides the relative target state required for motion coordination. The cooperative planner generates the desired trajectory for both the aerial platform and manipulator. The resulting manipulator motion subsequently changes the coupling disturbance acting on the aerial platform and affects grasping accuracy. Each component must be specifically addressed to meet the requirements of dynamic and precise grasping, while the interactions must be considered at the system level.

\subsection{Contribution}
To address these coupled issues, we present \textbf{CAPTURER}, an autonomous aerial manipulator for precise grasping across diverse environments. CAPTURER is composed of a coaxial quadrotor and a rigid-link serial manipulator. Each component is specifically developed to address the requirements of dynamic and precise aerial grasping. Fig. \ref{IJRR_framework} summarizes the framework and the information exchanged between its components.

We first develop an \textbf{onboard perception framework} for dynamic aerial grasping. The framework combines state estimation and onboard target detection. A drift correction mechanism is developed to reduce accumulated localization error. Moreover, a global pose transformation further transfers target observations into a reference frame. This allows target information acquired from different robot states to be used consistently during the grasping process. With respect to moving targets, the target state is continuously updated to provide the relative information required by the cooperative planner. An active yaw search strategy is further introduced to detect targets outside the initial camera's FoV.

Next, we propose an \textbf{optimization-based cooperative motion planning method} for the aerial platform and manipulator. The planner jointly optimizes their motions according to  dynamic characteristics and the relative target motion. In comparison with the strict constraint that the flight attitude is fixed \citep{10224341,10156288}, the platform attitude is explicitly included in the optimization to improve coordination during grasping. By optimizing both motions within a unified formulation, the planner coordinates the mobility of the quadrotor with the local dexterity of the serial manipulator. The presented algorithm can continuously adapt motion allocation as the target state changes.

Finally, we present a \textbf{robust anti-disturbance controller} for dynamic manipulation. A full-state dynamic model is exploited to construct the coupling forces and torques induced by manipulator motion. A robust differentiator is used to provide the high-order signals required by the coupling model. The estimated coupling effects are compensated through a feedforward channel. Moreover, two observers are developed to compensate for residuals arising from model uncertainty and external disturbances. This design enables accurate tracking of the references generated by the cooperative planner during dynamic grasping.

The proposed scheme is verified through extensive experiments. We first assess the onboard perception system through state estimation and target detection experiments. The presented robust controller is compared with baseline methods under composite disturbances. Finally, the ability of CAPTURER is validated through real aerial grasping experiments with stationary and moving targets under both indoor and outdoor conditions. CAPTURER achieves centimeter-level grasping accuracy and successfully grasps a marker pen only 1.5 cm in diameter at speeds exceeding 1.5 m/s.

To recap, our contributions are summarized as follows:

\begin{itemize}
	
	\item We develop CAPTURER, a dexterous aerial manipulation system for dynamic and precise grasping. An onboard perception framework is developed to integrate state estimation, target detection, drift correction, global pose transformation, and active target search. The resulting perception pipeline provides consistent robot and target states during flight and reduces the dependence on external localization and initial target visibility \citep{npjrobotics,8931787}.
	
	\item An optimization-based cooperative motion planning method is proposed to jointly optimize the states of the aerial platform and the manipulator according to their dynamic characteristics and the relative target motion. The platform attitude information is explicitly incorporated into the end-effector prediction model, allowing large-range aerial mobility and local manipulator dexterity to be coordinated. Unlike pre-programmed motions \citep{9981668,9834073} and one-shot offline solutions \citep{Kumar2014,npjrobotics,10224341}, the proposed planner supports online replanning at a rate exceeding 25 Hz.
	
	\item Unlike existing methods that compensate only first-order coupling terms \citep{10083097,8047452}, a full-state disturbance rejection controller is proposed for dynamic aerial manipulation. The nonlinear coupling effects are compensated through a feedforward channel, while a robust differentiator is employed to estimate the required high-order signals. Moreover, residual observers are incorporated to estimate the remaining disturbances caused by model mismatch and external disturbances, thereby improving tracking accuracy during manipulation.
	
\end{itemize}

\section{Related work}
\subsection{Aerial grasping}

\begin{table*}[t]
	\centering
	\caption{Comparison of aerial manipulation systems.}
	\label{tab:aerial_manipulation_comparison}
	\setlength{\tabcolsep}{5pt}
	\renewcommand{\arraystretch}{1.2}
	\resizebox{\textwidth}{!}{%
		\begin{tabular}{clllllcc}
			\toprule
			\textbf{Ref.} &
			\textbf{Platform} &
			\multicolumn{2}{c}{\textbf{Target}} &
			\multicolumn{2}{c}{\textbf{Perception}} &
			\textbf{Plan} &
			\textbf{Control} \\
			\cmidrule(lr){3-4}
			\cmidrule(lr){5-6}
			\cmidrule(lr){7-7}
			\cmidrule(lr){8-8}
			& &
			Motion &
			Type &
			Robot &
			Target &
			Replan capability\textsuperscript{*} &
			Disturbance rejection\textsuperscript{\dag} \\
			\midrule
			
			\multicolumn{8}{l}{\textit{A. Hovering grasping of stationary targets}}\\
			\addlinespace[2pt]
			
			\citet{7989751} &
			Hexarotor + serial arm &
			Static &
			Cylinder &
			Mocap &
			Visual &
			\tabyes &
			\tabno \\
			
			\citet{9866833} &
			Quadrotor + folding arms &
			Static &
			Box &
			Mocap &
			Known &
			\tabno &
			\tabno \\
			
			\citet{10339889} &
			Quadrotor + serial arm &
			Static &
			Marker pen &
			Mocap &
			Mocap &
			\tabyes &
			\tabno \\
			
			\citet{10486851} &
			Quadrotor + Delta arm &
			Static &
			Block &
			Mocap &
			Mocap &
			\tabno &
			\tabyes \\
			
			\textbf{Ours} &
			\textbf{Coaxial + serial arm} &
			Static &
			\textbf{Marker pen} &
			\textbf{Visual} &
			\textbf{Visual} &
			\tabyes &
			\tabyes \\
			
			\midrule
			
			\multicolumn{8}{l}{\textit{B. Dynamic grasping of stationary targets}}\\
			\addlinespace[2pt]
			
			\citet{9981668} &
			Quadrotor + gripper &
			Static &
			Box &
			Mocap &
			Known\textsuperscript{\ddag} &
			\tabno &
			\tabno \\

			\citet{10224341} &
			Quadrotor + 1-DoF arm &
			Static &
			Iron sphere &
			Mocap &
			Known &
			\tabno &
			\tabyes \\
			
			\citet{10495172} &
			Quadrotor + morphing claws &
			Static &
			Toy doll &
			Mocap &
			Known &
			\tabno &
			\tabyes \\
			
			\citet{npjrobotics} &
			Quadrotor + soft gripper &
			Static &
			Bottle &
			Visual &
			Visual &
			\tabno &
			\tabyes \\
			
			\textbf{Ours} &
			\textbf{Coaxial + serial arm} &
			Static &
			\textbf{Marker pen} &
			\textbf{Visual} &
			\textbf{Visual} &
			\tabyes &
			\tabyes \\
			
			\midrule
			
			\multicolumn{8}{l}{\textit{C. Dynamic grasping of moving targets}}\\
			\addlinespace[2pt]
			
			\citet{10156288} &
			Quadrotor + serial arm &
			$<$ 0.1 m/s &
			Marker pen &
			Mocap &
			Mocap &
			\tabyes &
			\tabno \\
			
			\citet{10224341} &
			Quadrotor + 1-DoF arm &
			Circular motion &
			Iron sphere &
			Mocap &
			Known &
			\tabno &
			\tabyes \\
			
			\citet{npjrobotics} &
			Quadrotor + soft gripper &
			0.15 / 0.3 m/s &
			Med-kit &
			Mocap &
			Mocap &
			\tabno &
			\tabyes \\
			
			\textbf{Ours} &
			\textbf{Coaxial + serial arm} &
			\textbf{0.45 / 0.75 m/s} &
			\textbf{Marker pen} &
			\textbf{Mocap} &
			\textbf{Visual} &
			\tabyes &
			\tabyes \\
			
			\bottomrule
		\end{tabular}%
	}
	\vspace{4pt}
	\begin{minipage}{\textwidth}
		\footnotesize\raggedright
		\textsuperscript{*}: The desired motion reference is regenerated online according to the current system and target state. Tracking a precomputed trajectory  alone before grasping is not regarded as replanning.
		\textsuperscript{\dag}: A dedicated module is developed to address disturbances.
		\textsuperscript{\ddag}: Target information is provided a priori.
	\end{minipage}
\end{table*}

Specialized grippers and reconfigurable structures provide different ways to achieve aerial grasping. A low-complexity gripper mounted on a quadrotor is utilized to grasp a static object. Similarly, \citet{9834073} design a passively-triggered gripper to grasp an airborne drone. Moreover, an aerial shooting robot driven by air is presented to stick the distant object \citep{10044987}. In addition, a custom soft gripper is developed to enable flexibility in grasping different objects by increasing the contact surface \citep{9981668}. Nevertheless, a fixed and known environment (such as the pre-defined trajectory) is required in these methods \citep{04,9834073,10044987,9981668,10339889,9309359,Ryll20196DIC,10083097,nature,8059875}. Their applicability in unknown or unstructured scenes remains uncertain.

Several claw-inspired deformable mechanisms are developed to eliminate the need of a manipulator \citep{9866833,8594330,9196576,10495172}. \citet{8594330}  design a deformable quadrotor capable of controlling expansion and contraction to grasp different objects. A similar transformable aerial vehicle is described in \citet{9196576}. Moreover, a biomimetic morphing quadrotor has been explored for dynamic grasping \citep{10495172}. In an attempt to improve the success rate of aerial manipulation, the moving platform will actively slow down when approaching the object \citep{10224341}. A permanent magnet is also mounted to act as the end-effector for extending the feasible grasping area. However, the system struggles to respond in real time due to high computational cost (over 10 s). Moreover, the dependence on external positioning systems significantly limits their scalability to real-world deployment. Our previous work achieved precise end-effector operation when the aerial platform maintained a hovering position \citep{10339889,10547187}. However, the grasping trajectory is predefined and the target remains stationary.

In contrast to systems relying on specialized end-effectors or airframe deformation, CAPTURER adopts a compact multirotor platform equipped with a rigid-link serial manipulator and gripper. The aerial platform provides large-range mobility for target approach, while the manipulator supplies local motion for precise aerial grasping. At the same time, the structure introduces strong dynamic disturbances between the floating platform and the serial manipulator, making onboard perception, cooperative planning, and disturbance rejection control essential for dynamic and precise grasping.

\subsection{Onboard perception}
Onboard perception for aerial manipulation involves two closely related tasks: estimating the robot state for feedback control and detecting the target information for manipulation. 

Accurate state information is particularly critical for aerial grasping because estimation errors propagate to motion planning, trajectory tracking, and ultimately the end-effector pose. LiDAR-inertial state estimation supports accurate vehicle localization and navigation \citep{9697912,scirobotics_Ladar}, while satellite-inertial fusion provides an alternative when satellite observations are available \citep{9615117}. Visual-inertial sensing offers a compact solution and has been widely employed \citep{9102390,Swarm,npjrobotics,IJRR_Zhao}. However, visual state estimation is subject to accumulated drift, and its accuracy can vary with environmental conditions such as visual texture, illumination, and feature availability. These variations become particularly important for dynamic grasping because such drift directly affects the end-effector manipulation accuracy. Hence, a mechanism for correcting accumulated localization drift is required to maintain grasping accuracy across changing environments.

Another critical challenge in aerial grasping is target detection and motion prediction. Reliable state information is essential for cooperative planning and successful grasping, particularly when the target moves. Image-based methods use visual observations to directly guide aerial manipulators toward targets \citep{7989751,9789094,9781617}. Keypoint-based detection combined with prior object geometry has also been used for onboard target pose estimation \citep{npjrobotics}. Detection and motion prediction have further been investigated for autonomous search and interception of moving objects \citep{VRBA2022103970,rodriguezramos}. Although recurrent neural network (RNN)-based methods provide strong perception capability, their onboard implementation may require considerable computational resources and limit the perception update rate. Fiducial-based approaches such as AprilTag provide a computationally efficient alternative when the target can be instrumented with markers \citep{7759617}. However, most existing approaches assume that the target is  visible in the camera's FoV throughout the grasping process. The assumption becomes restrictive when the initial target direction is unknown. Therefore, target perception for dynamic aerial grasping should provide not only accurate and low-latency measurements, but also active target search and continuous state estimation.

To address these issues, CAPTURER employs a dual-camera configuration for state estimation and target detection, respectively. The perception pipeline incorporates drift correction, global pose transformation, and active target search. The detected target information is used for real-time drift correction, reducing accumulated localization errors of the aerial platform. Global pose transformation maps target observations into a fixed reference frame. The active yaw search strategy adjusts the sensing direction to detect targets outside the initial camera field of view, thereby relaxing the requirement for initial target visibility. Together, these components provide consistent robot and target states for dynamic grasping.

\subsection{Cooperative planning}

Unlike trajectory generation for UAVs \citep{IJRR_kummar,Champion_level,IJRR_FuZhang}, cooperative planning of aerial manipulators involves the simultaneous motion of the moving platform and manipulator. The most fundamental issue stems from the high-dimensional state space, as the degrees-of-freedom (DoFs) of the UAV and the manipulator are combined. In comparison with ground robots, the motion planning of aerial manipulators is inherently more complex because the workspace has been expanded. Both the mobility and the flexibility should be jointly considered during motion generation. Currently, the existing methods of addressing the issue can be broadly classified into two categories: model-based and optimization-based methods. 

Model-based strategies derive the cooperative trajectory using algebraic approaches, such as differential kinematics, pseudo-inversion, and null space projection. The closed-loop inverse kinematics (CLIK) algorithm is responsible for computing motion references within the control loop using differential kinematics \citep{6608869,10083097,Behavioral}. In addition, a sampling-based motion planner is proposed to tackle the task-constrained problem with the aid of the second-order CLIK algorithm \citep{8283709}. The null space-based (NSB) approach, a popular redundancy-utilized strategy, has been extended to trajectory generation \citep{7962217,LAI2022102724,7017672}. In accordance with the NSB algorithm, the planned motion reference is capable of accomplishing the primary task as well as other secondary tasks arranged in a given priority order, such as obstacle avoidance \citep{7962217}, visual servoing \citep{LAI2022102724}, and CoM alignment \citep{7017672}. Nevertheless, these methods mainly resolve cooperative motion by calculating the pseudo-inverse of the nonlinear mapping matrix, which may result in excess velocities in configuration space. An improved damped least-squares (DLS) strategy introduces a damping term to the diagonal of the Jacobian matrix for regularization \citep{8796272}, preventing aerial manipulators from approaching singular configurations. Furthermore, kinodynamic planners allow kinematic and dynamic constraints to be considered during cooperative trajectory generation \citep{9462539,9194145,IJARS}. For example, dynamics-aware search has been used to generate manipulator joint trajectories while avoiding collision with the aerial platform \citep{IJARS}.

Optimization-based methods have been increasingly investigated for cooperative trajectory generation.
Compared with model-based methods, optimization-based frameworks allow physical and task constraints to be explicitly incorporated into the planning formulation. An optimization-based planner proposed in \citet{CAO2022107607} generates the motion reference using the move-blocking scheme. In order to avoid collisions with external obstacles during the interaction phase, a quadratic programming (QP) formulation is defined to compute the desired trajectory in configuration space \citep{7762762,10943237}. Moreover, a hierarchical quadratic programming (HQP) approach is presented to execute the object grasping task within inequality constraints in \citet{10156288}. However, the attitude information of the aerial platform is not incorporated into the optimization formulation. Moreover, the aerial manipulator may encounter oscillation during maneuvering. The undesirable phenomenon leads to uncoordinated performance and a significant reduction in overall efficiency. The nonlinear model predictive control (NMPC) method considering dynamic coupling effects is developed to formulate the cooperative planning problem \citep{10224341}. However, the presented NMPC method arrives at a one-shot offline solution. In \citet{npjrobotics}, the minimum-snap trajectory generation scheme is used to compute the aerial platform trajectory offline after the target information is obtained.

Dynamic aerial grasping imposes additional requirements on cooperative planning. The motion contributions of the UAV and the manipulator should therefore be adjusted online according to the relative target state. In addition, the platform attitude affects the manipulator configuration. Since different combinations of the aerial platform and manipulator states can satisfy the same end-effector motion, simultaneous optimization may lead to oscillatory behavior. To address these requirements, a kinematic-level optimization framework is established to jointly optimize the motions of the aerial platform and manipulator while considering multiple physical constraints. Unlike previous offline solutions \citep{9981668,9834073,Kumar2014,npjrobotics,10224341}, the proposed planner supports online replanning at a rate exceeding 25 Hz, enabling real-time coordination during dynamic grasping.

\subsection{Dynamic control}
Efficient and reliable dynamic control performance is a key element to perform aerial manipulation. In comparison with bare flying platforms, the presence of strong coupling effects poses a great influence on the control capability. According to \citet{shabana2020dynamics}, the dynamic coupling disturbances caused by the operating manipulator can aggravate the aerial platform fluctuation. Moreover, due to the presence of higher-order terms in the system dynamics, the coupling effects would propagate to the acceleration of each manipulator link, deteriorating the control performance of the end-effector \citep{shabana2020dynamics}. 
\begin{figure}[!tp]
	\centering
	\includegraphics[width=0.85\linewidth,height=0.85\textheight,keepaspectratio]{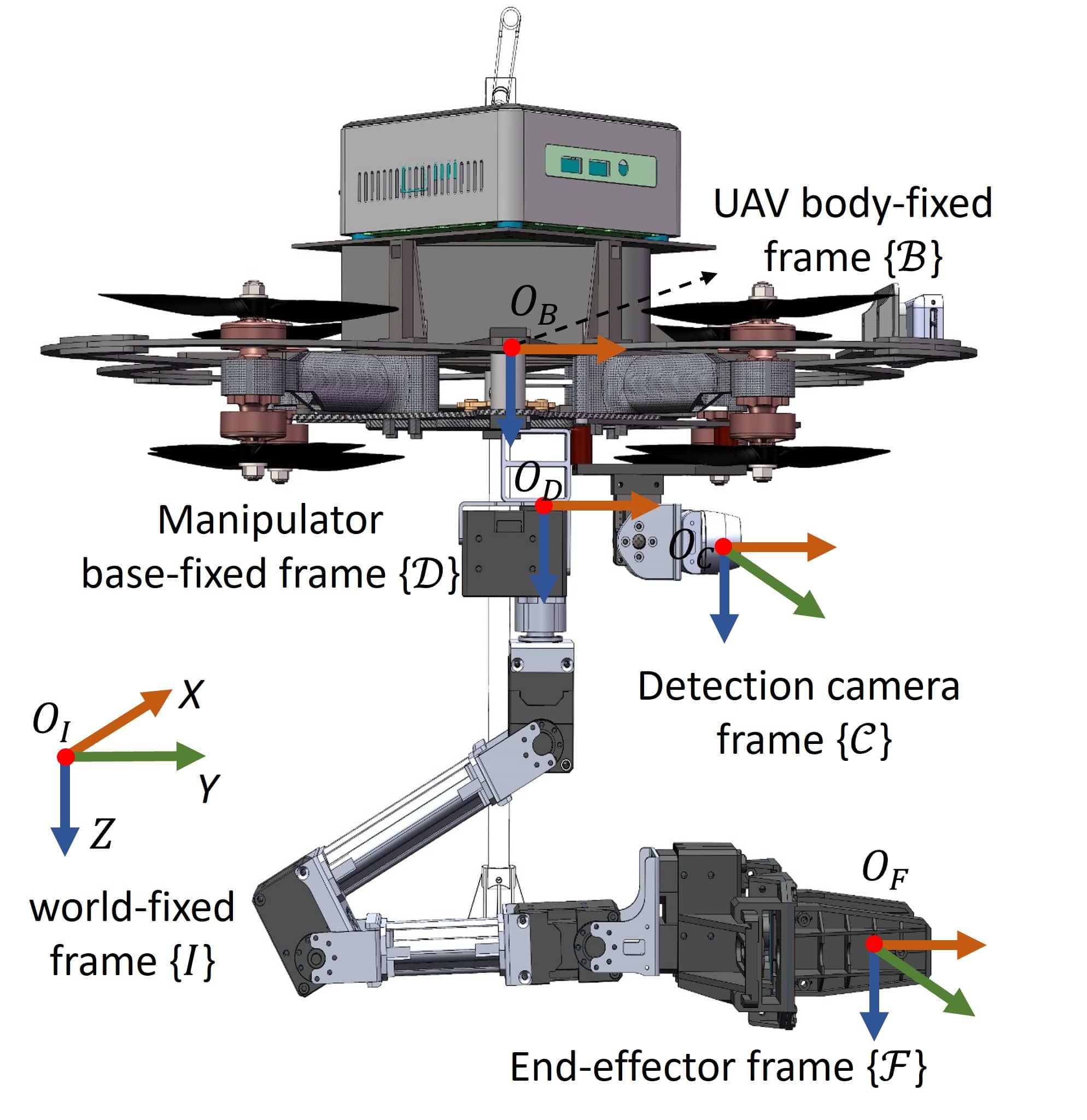}
	\caption{Schematic of CAPTURER.}\label{IJRR_frame}
\end{figure}

In \citet{7353623}, the general Euler-Lagrange equation is exploited to model the coupling effects. However, only the coupling torque generated by the displacement of the center-of-mass (CoM) is established. Therefore, the model uncertainties race up when the manipulator motion speed increases. In \citet{8924899}, a variable inertia method is adopted to establish the coupling model between the hexarotor and the 2-DoF manipulator while a H$\infty$ controller is used to suppress the modeling residuals. Similarly, a model-based estimator is exploited to handle the coupling dynamics caused by the parallel manipulator \citep{9507086}. Regarding dynamic scenarios, the generation mechanism of the coupling disturbances is presented under the rapid periodic movements of the manipulator \citep{10722859}. Although excellent control performance has been achieved using a model predictive control (MPC) method in simulation \citep{EMAMI2021106573}, avoiding the impacts caused by strong dynamic coupling disturbances remains an intractable challenge in practice. It is emphasized that the remarkable control performance is dependent on an accurate dynamical model \citep{10478625}. 

In addition to establishing precise coupling dynamics, learning-based methods have drawn growing attention in recent years \citep{CAO2022107607,IJRR_Jia,IJRR_ChenWang,jia2026physics}. In \citet{CAO2022107607}, an adaptive neural network (NN) is presented to control the motion of the aerial manipulator. In \citet{IJRR_Jia}, a bilevel learning framework is utilized to capture the shared disturbance information across different environments. Existing studies \citep{NEURIPS2018_69386f6b,10598388} exploit ordinary differential equation networks to represent the dynamical model. A reinforcement learning-based method is developed to reduce the coupling disturbances by optimizing the manipulator motion \citep{9345436}. Nevertheless, the trajectory of the manipulator is pre-calculated and cannot be re-planned online according to current system response. Moreover, the quality of feedback signals plays a pivotal role that affects the network approximation accuracy. In other words, the presence of signal noise may result in poor response, even leading to the divergence of NN estimations \citep{10547187}. 

\begin{table}[tbp]
	\centering
	\caption{\textbf{Nomenclature.}}
	\label{Nomenclature}
	\renewcommand{\arraystretch}{1.22}
	\setlength{\tabcolsep}{3pt}
	\small
	\begin{tabularx}{\columnwidth}{
			@{}>{\raggedright\arraybackslash}p{0.27\columnwidth}
			>{\raggedright\arraybackslash}X@{}
		}
		\hline
		\hline
		Symbols & Physical interpretations \\
		\hline
		
		$m_s\in\mathbb{R}$
		& Total mass of the aerial manipulator \\
		
		$\boldsymbol{J}\in\mathbb{R}^{3\times3}$
		& Inertia matrix of the aerial platform \\
		
		$p_b\in\mathbb{R}^{3}$
		& The aerial platform position in $\{\mathcal{I}\}$ \\
		
		$p_e\in\mathbb{R}^{3}$
		& The end-effector position in $\{\mathcal{I}\}$ \\
		
		$p_{ed}\in\mathbb{R}^{3}$
		& The end-effector position relative to the manipulator base, expressed in $\{\mathcal{D}\}$ \\
		
		$v_b\in\mathbb{R}^{3}$
		& The UAV linear velocity in $\{\mathcal{I}\}$ \\
		
		$v_e\in\mathbb{R}^{3}$
		& The end-effector linear velocity in $\{\mathcal{I}\}$ \\
		
		$\omega_b\in\mathbb{R}^{3}$
		& The UAV angular velocity in $\{\mathcal{B}\}$ \\
		
		$\Omega_b\in\mathbb{R}^{3}$
		& The UAV angular velocity in $\{\mathcal{I}\}$ \\
		
		$\Omega_e\in\mathbb{R}^{3}$
		& The end-effector angular velocity in $\{\mathcal{I}\}$ \\
		
		$\boldsymbol{R_e^d}\in SO(3)$
		& Rotation matrix from $\{\mathcal{F}\}$ to $\{\mathcal{D}\}$ \\
		
		$\boldsymbol{R_d^b}\in SO(3)$
		& Rotation matrix from $\{\mathcal{D}\}$ to $\{\mathcal{B}\}$ \\
		
		$\boldsymbol{R_b}\in SO(3)$
		& Rotation matrix from $\{\mathcal{B}\}$ to $\{\mathcal{I}\}$ \\
		
		$\boldsymbol{R_e}\in SO(3)$
		& Rotation matrix from $\{\mathcal{F}\}$ to $\{\mathcal{I}\}$ \\
		
		$q\in\mathbb{R}^{4}$
		& The manipulator joint position \\
		
		$\boldsymbol{J_{t}}(q)\in\mathbb{R}^{3\times4}$
		& Translational part of the manipulator Jacobian \\
		
		$\boldsymbol{J_{r}}(q)\in\mathbb{R}^{3\times4}$
		& Rotational part of the manipulator Jacobian \\
		
		$e_3=[0,0,1]^{\top}$
		& Unit vector along the $Z$-axis \\
		
		\hline
	\end{tabularx}
\end{table}

In contrast to learning-based approaches, another basic perspective is the utilization of disturbance observer (DO). The studies in \citet{8047452,9309359} introduce a singular perturbation scheme to ensure the flight stability of the moving platform while the DO is developed to directly estimate unknown disturbances. In \citet{10083097}, the first-order coupling terms are directly compensated while two extended state observers (ESOs) are introduced to estimate the remaining disturbances. However, the neglected higher-order terms have a detrimental impact on control accuracy, particularly in dynamic operations. The reported experiments are conducted at motion speeds below 0.1 m/s, where the influence of neglected higher-order coupling terms is relatively limited. In addition, the same regulating mechanism is used to adjust different observed states \citep{10083097}.

Different from approaches that rely primarily on partial coupling information \citep{8047452,9309359,10083097}, CAPTURER combines model-based coupling compensation with residual disturbance estimation. A full-state dynamical model is established to describe the strong coupling effects. Moreover, a robust differentiator is designed to reconstruct the high-order signals required by the coupling model \citep{SEEBER2021109858}, and the modeled coupling forces and torques are directly compensated through a feedforward channel. Two residual observers are subsequently introduced to estimate residuals caused by model uncertainty and external wind disturbances, thereby improving the disturbance rejection capability during manipulation.

\begin{figure}[tp]
	\centering
	\includegraphics[width=0.98\linewidth,height=\textheight,keepaspectratio]{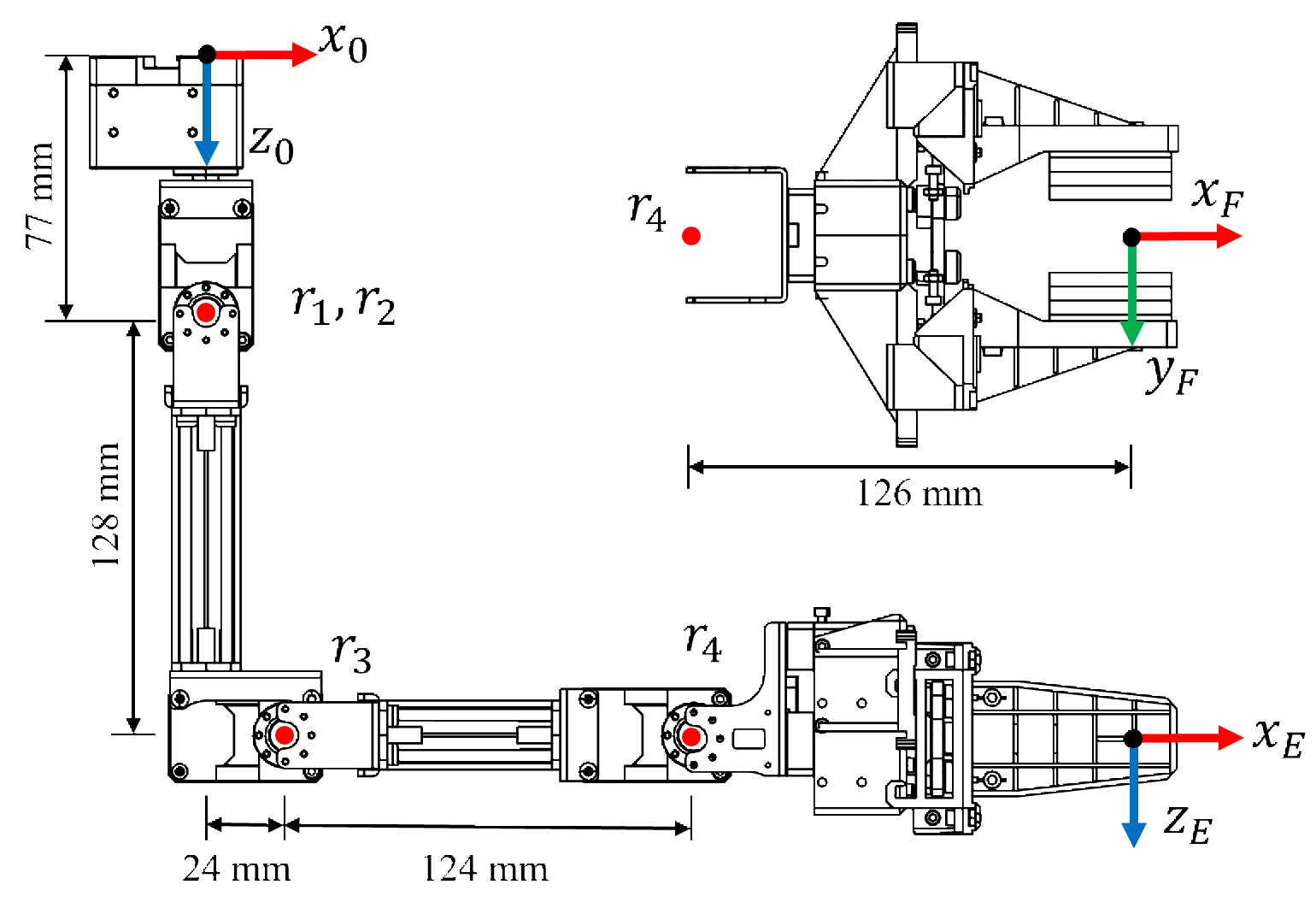}
	\caption{Schematic of the serial manipulator.}\label{IJRR_PoE}
\end{figure}

\begin{figure}[tp]
	\centering
	\includegraphics[width=0.98\linewidth,height=\textheight,keepaspectratio]{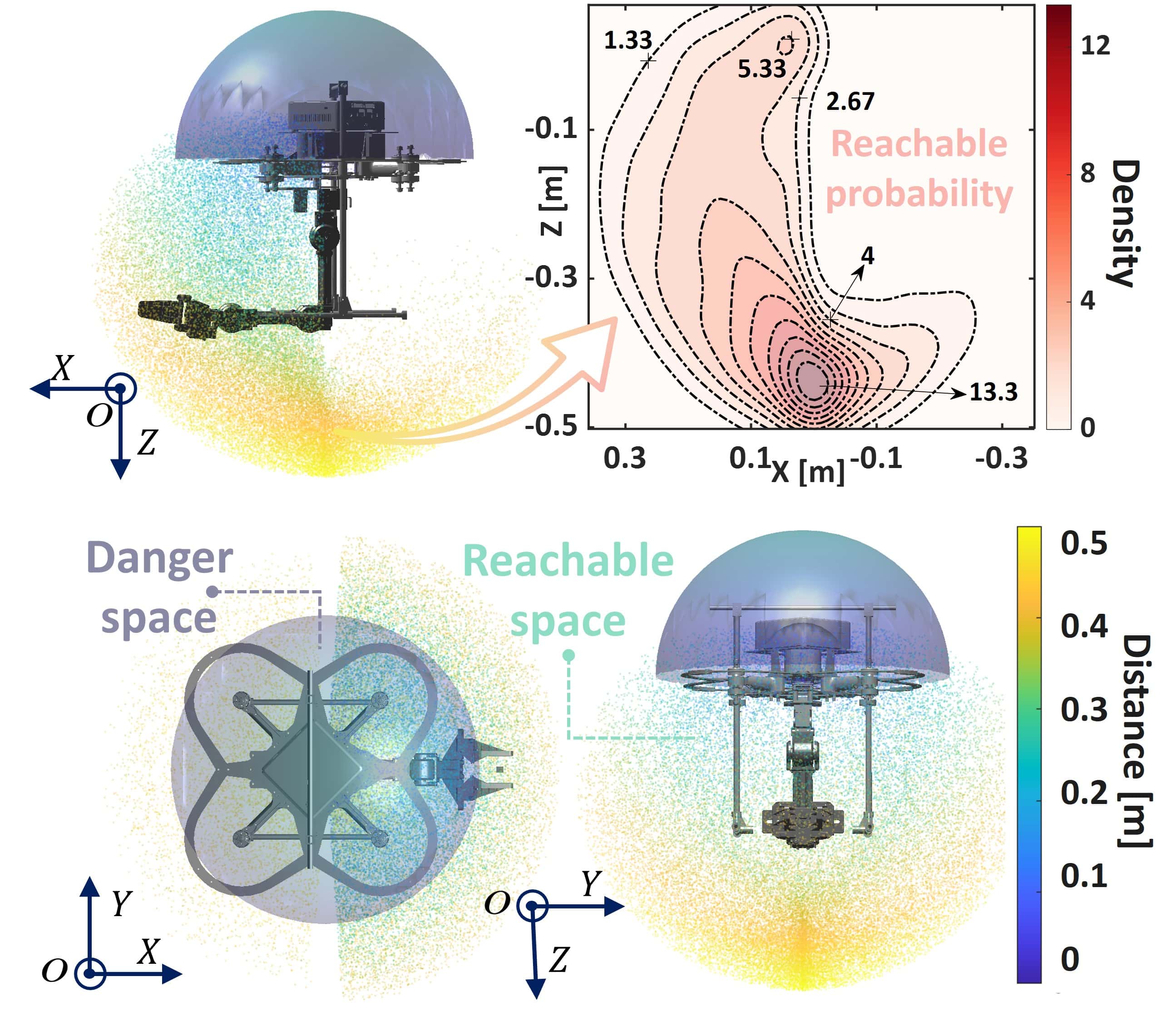}
	\caption{Visualization
		of the operational capability.}\label{IJRR_reachable}
\end{figure}

%% file: sections/02_system.tex
\section{Preliminaries}\label{m_Overview}

In this section, the system model is presented. Unlike conventional UAVs or robotic arms, the aerial manipulator exhibits distinct characteristics arising from its coupled structure \citep{9929266}. 

As shown in Fig. \ref{IJRR_frame}, five frames are defined to describe the kinematics: the world-fixed frame $\{\mathcal{I}\}$, the body-fixed frame $\{\mathcal{B}\}$, the detection camera frame $\{\mathcal{C}\}$, the manipulator base-fixed frame $\{\mathcal{D}\}$, and the end-effector frame $\{\mathcal{F}\}$. The main physical symbols used in the following derivations are shown in Table \ref{Nomenclature}.


\subsection{Kinematics}\label{kinematics}
Based on the coordinate frames defined above, the overall kinematics can be formulated as:
\begin{equation}
	\left\{
	\begin{matrix}
		p_e=p_b+\bm{R_b}\left(p_o+\bm{R_d^b}p_{ed}\right)\\[4pt]
		\bm{R_e}=\bm{R_b}\bm{R_d^b}\bm{R_e^d}
	\end{matrix}
	\right.,\label{kinematic_model}
\end{equation}
where $p_o\in\mathbb{R}^{3}$ denotes the constant position offset from $\{\mathcal{B}\}$ to $\{\mathcal{D}\}$, expressed in $\{\mathcal{B}\}$. The offset can be directly obtained by extrinsic calibration. Since $\{\mathcal{B}\}$ and $\{\mathcal{D}\}$ are aligned, $\bm{R_d^b}=\bm{I_3}$.

Differentiating Eq. \eqref{kinematic_model}, we have
\begin{equation}
	\left\{
	\begin{matrix}
		v_e=v_b-\bm{R_b}\left(p_o+p_{ed}\right)^\times\omega_b+\bm{R_b}\bm{J_t}(q)\dot{q}\\[4pt]
		\Omega_e=\Omega_b+\bm{R_b}\bm{J_r}(q)\dot{q}
	\end{matrix}
	\right.\label{kinematic_V2}
\end{equation}
where $(\cdot)^{^\times}$ represents the skew-symmetric operator. Eq. \eqref{kinematic_V2} shows that the end-effector motion depends jointly on the aerial platform motion and manipulator joint motion. 

For the manipulator, the product-of-exponentials (PoE) formulation is adopted to describe its forward kinematics. The PoE formulation provides a compact geometric representation \citep{kevin2017}. Only the initial transformation matrix $\bm{T_0}\in SE(3)$ and the screw-axis parameters of each joint are required. Let $\vartheta_i\in\mathbb{R}^{3}$ be the unit rotation axis of the $i$th joint, and $r_i\in\mathbb{R}^{3}$ be the position of an arbitrary point on the corresponding joint axis, where $i=1,\ldots,4$. The forward kinematics  can be expressed as:
\begin{subequations}
	\begin{align}
		\nu_i &= -\vartheta_i\times r_i, \label{eq:prelim_poe:a}\\
		\widehat{\bm\xi}_i &= \begin{bmatrix}\vartheta_i^\times & \nu_i \\ \mathbf{0}_{1\times3} & 0\end{bmatrix}, \label{eq:prelim_poe:b}\\
		\bm{T}^D_F(q) &= \left(\prod_{i=1}^{4}e^{\widehat{\bm\xi}_i q_i}\right)\bm{T}_0, \label{eq:prelim_poe:c}
	\end{align}
	\label{eq:prelim_poe}
\end{subequations}
where $\bm{T^D_F}\in SE(3)$ is the transformation matrix from $\{\mathcal{F}\}$ to $\{\mathcal{D}\}$. The  manipulator system parameters are listed in Table \ref{POE}. Fig. 4 shows the reachable workspace and reachability distribution of the end-effector.

Our previous work has shown that the kinematic coupling can significantly amplify the propagation of floating-base errors to the end-effector \citep{10339889}. In particular, even a small attitude deviation of the aerial platform can induce a considerable end-effector position error when the manipulator is extended. The error propagation directly reduces end-effector tracking accuracy and can compromise precise grasping.

\begin{table}[htbp]
	\centering
	\caption{\textbf{The manipulator system parameters.}}
	\label{POE}
	\renewcommand{\arraystretch}{1.25}
	\setlength{\tabcolsep}{4pt}
	\small
	
	\begin{tabularx}{\columnwidth}{
			@{}c
			>{\centering\arraybackslash}p{0.23\columnwidth}
			>{\centering\arraybackslash}p{0.30\columnwidth}
			>{\centering\arraybackslash}X@{}
		}
		\hline
		\hline
		Joint &
		$\vartheta_i$ &
		$r_i$ (mm) &
		Mass (g) \\
		\hline
		\noalign{\vskip 2pt} 
		1 &
		$\begin{bmatrix}0&0&1\end{bmatrix}^{\top}$ &
		$\begin{bmatrix}0&0&77\end{bmatrix}^{\top}$ &
		210 \\[2pt]
		
		2 &
		$\begin{bmatrix}0&1&0\end{bmatrix}^{\top}$ &
		$\begin{bmatrix}0&0&77\end{bmatrix}^{\top}$ &
		146 \\[2pt]
		
		3 &
		$\begin{bmatrix}0&1&0\end{bmatrix}^{\top}$ &
		$\begin{bmatrix}24&0&205\end{bmatrix}^{\top}$ &
		138 \\[2pt]
		
		4 &
		$\begin{bmatrix}0&1&0\end{bmatrix}^{\top}$ &
		$\begin{bmatrix}148&0&205\end{bmatrix}^{\top}$ &
		336 \\
		
		\hline
	\end{tabularx}
\end{table}

\subsection{Dynamics}\label{}
Combined with the kinematics described in Section \ref{kinematics}, the UAV dynamics can be portrayed as:
\begin{equation}
	\left\{\begin{matrix}{\dot{p}}_b=v_b\\{\dot{R}_{\bm{b}}}=\bm{R_b}\omega_b^\times\\m_s{\dot{v}}_b=m_sge_3-fb_3+F_m\\
		\textbf{\textit{J}}{\dot{\omega}}_b=-\omega_b^\times\textbf{\textit{J}}\omega_b+\tau+\tau_m\\\end{matrix}\right.,\label{Model}
\end{equation}
\begin{figure*}[htp]
	\centering
	\includegraphics[width=\linewidth,height=\textheight,keepaspectratio]{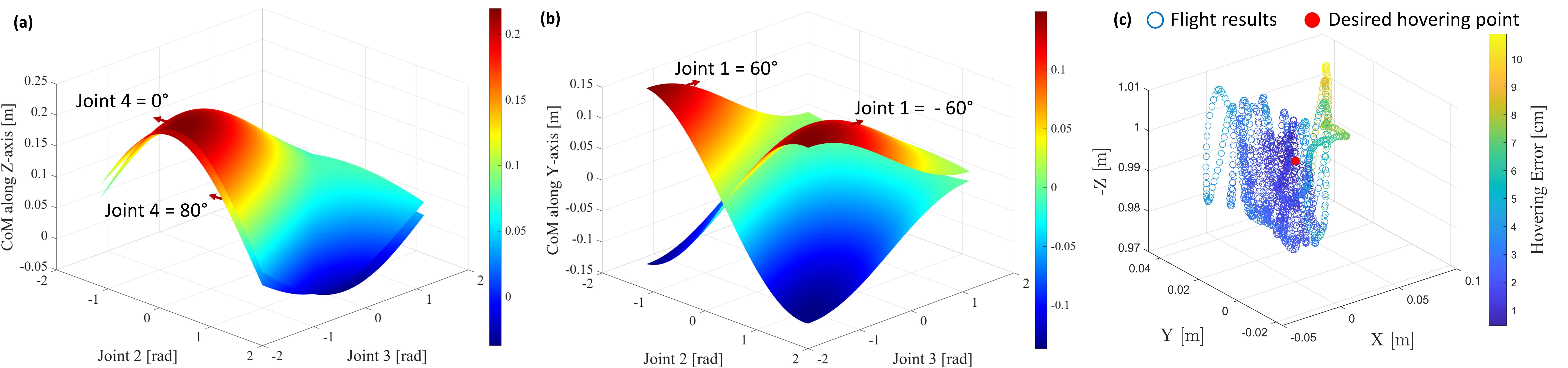}
	\caption{(a) CoM shift along $Z$-axis. (b) CoM shift along $Y$-axis. (c) The aerial platform is oscillating when the manipulator moves.}\label{IJRR_CoM}
\end{figure*}
where $b_3=\textbf{\textit{R}}_be_3$. $f\in\mathbb{R}$ is the total thrust generated by eight propellers and $\tau\in\mathbb{R}^{3}$ is the control moment of the rotational loop.

On the basis of \citet{shabana2020dynamics}, the strong dynamic coupling disturbances are completely determined by the system states and their derivatives. Specifically, the dynamic coupling effects can be divided into two components: the rotation motion of the aerial platform and the relative movement of the manipulator with respect to its base. The detailed expression is formulated as:

\begin{equation}
	\begin{aligned}
		F_m=&-m_m\bm{R}_{b}\Big(\omega_b^\times\left(\omega_b^\times p_{bm}^B\right)+\dot{\omega}_b^\times p_{bm}^B+2\omega_b^\times\dot{p}_{bm}^B\\
		&+\ddot{p}_{bm}^B\Big),
		\label{coupling_force}
	\end{aligned}
\end{equation}
\begin{equation}
	\begin{aligned}
		\tau_m &= -\bm{J_{bm}^{B}}\dot{\omega}_b - \omega_b^{\times} \bm{J_{bm}^{B}}{\omega}_b + m_m (p_{bm}^{B})^{\times} \bm{R_b}^{\top}\left(g e_3 - \ddot{p}_b\right) \\
		&\quad-\dot{\bm{J}}_{\bm{bm}}^{\bm{B}}{\omega}_b-\omega_b^\times\bm{L}_{\bm{m}}^{\bm{B}}-\dot{\bm{L}}_{\bm{m}}^{\bm{B}},
		\label{coupling_torque}
	\end{aligned}
\end{equation}
where $F_m\in\mathbb{R}^3$ and $\tau_m\in\mathbb{R}^3$ denote the dynamic coupling force and torque, respectively. $m_m\in\mathbb{R}$ denotes the mass of the manipulator, and $p_{bm}^B\in\mathbb{R}^3$ denotes its CoM position with respect to $\{\mathcal{B}\}$. $\textbf{\textit{J}}_{\textbf{\textit{bm}}}^\textbf{\textit{B}}\in\mathbb{R}^{3\times3}$ and $\textbf{\textit{L}}_\textbf{\textit{m}}^\textbf{\textit{B}}\in\mathbb{R}^3$ denote the inertia tensor and angular momentum of the manipulator expressed in $\{\mathcal{B}\}$, respectively. A detailed analysis of the dynamic coupling disturbances is provided in our previous work \citep{10547187}.

\subsection{CoM shift analysis}\label{}
Unlike conventional aerial vehicles with a fixed mass distribution, the aerial manipulator experiences continuous mass redistribution during manipulation. The CoM of the overall system continuously shifts with manipulator joint motion, while the grasped object causes an additional displacement. The shift directly changes the moment generated by the total thrust with respect to the system CoM and therefore introduces additional coupling torque. The CoM position of the entire system $p_c\in\mathbb{R}^{3}$ expressed in $\{\mathcal{B}\}$ can be denoted as:

\begin{equation}
	p_c=\frac{1}{m_s}
	\sum_{i=1}^{n}
	m_i
	\left(
	p_o+
	\bm{R_d^b}p_{di}^{d}
	\right),
	\label{CoM_eq}
\end{equation}
where $m_{i}\in\mathbb{R}$ is the mass of the $i$-th manipulator link and $p_{di}^{d}\in\mathbb{R}^{3}$ denotes the position of the CoM of the $i$-th link expressed in $\{\mathcal{D}\}$.

Eq. \eqref{CoM_eq} shows that the CoM shift is directly determined by the manipulator configuration. Different joints contribute differently to the CoM shift because their motions redistribute the link masses along different directions. As shown in Fig. \ref{IJRR_PoE}, the first joint mainly affects the CoM position along the $Y$-axis, whereas Joints 2, 3 and 4 primarily change the CoM distribution along the $X$-and $Z$-axes. The evaluation process is illustrated in Figs. \ref{IJRR_CoM}(a-b).

\subsection{Coupling effects analysis}\label{coupling_analysis}
According to Eqs.~\eqref{coupling_force} and \eqref{coupling_torque}, the manipulator motion induces additional coupling disturbances. In particular, the higher-order coupling terms become more significant as the manipulator motion increases. 
\begin{figure}[tbp]
	\centering
	\includegraphics[width=0.98\linewidth,height=\textheight,keepaspectratio]{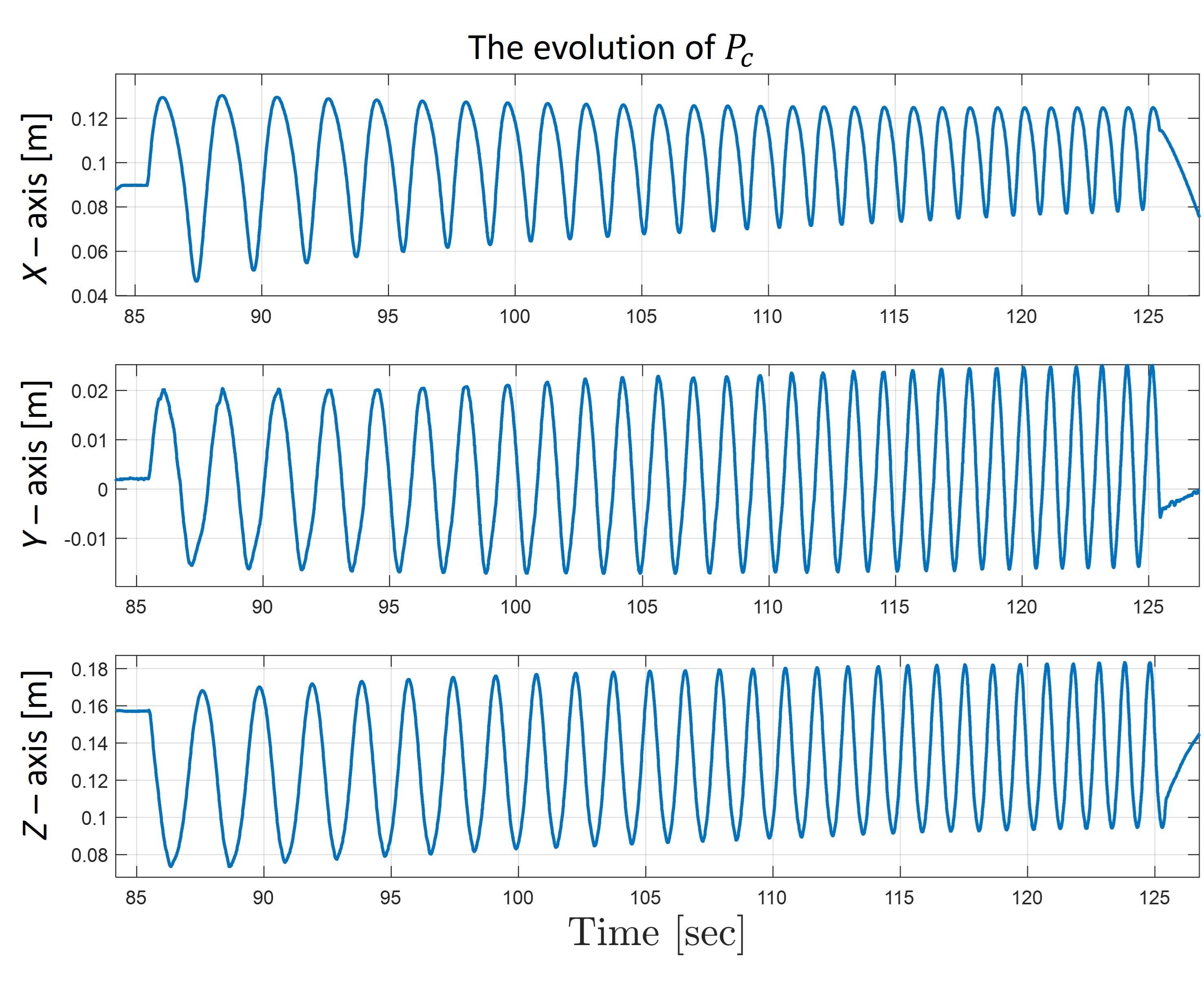}
	\caption{The evaluation of CoM position during manipulation motion.}\label{IJRR_Pc}
\end{figure}

In this section, a real experiment is conducted to  quantify the influence of the induced coupling disturbances. The aerial platform maintains a hover, while the manipulator tracks a prescribed trajectory. The desired end-effector trajectory is defined as:
$$
\left[0.28+\frac{R\sin(\omega t)}{\sqrt5},\frac{R\sin(\omega t)}{\sqrt{5/4}},0.105+R\cos(\omega t)\right]^{\top} m.
$$
\begin{figure*}[tbp]
	\centering
	\includegraphics[width=0.95\linewidth,height=\textheight,keepaspectratio]{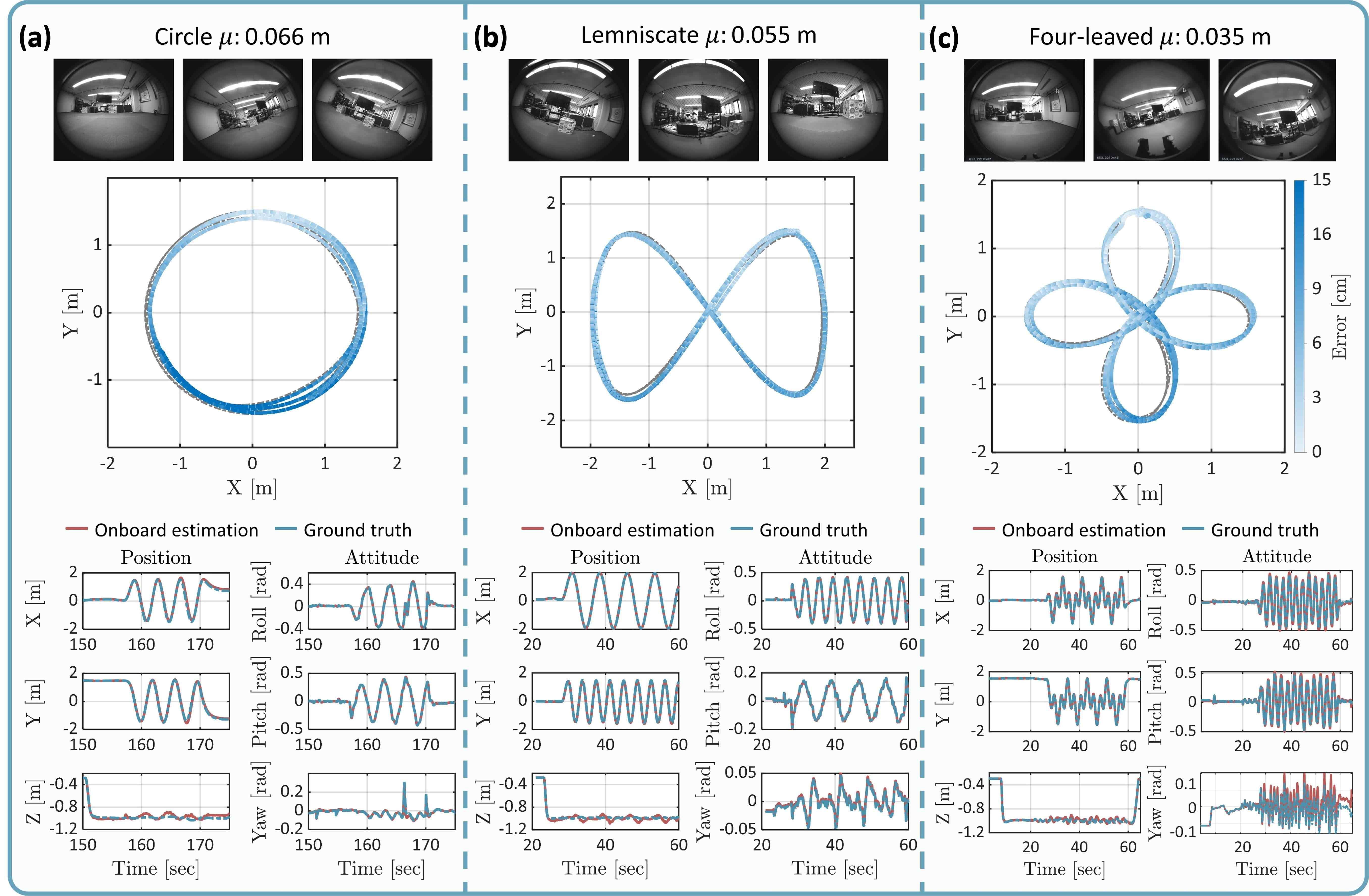}
	\caption{{Performance comparison between state estimation via onboard sensors and ground truth in three trajectories.} (a) Circle trajectory. (b) Lemniscate trajectory. (c) Four-leaf trajectory.
	}\label{IJRR_estimation}
\end{figure*}

The reference is a circular trajectory defined in both the vertical and horizontal planes. The primary objective is to sufficiently excite the dynamics of all manipulator joints \citep{631234}. Taking into account the reachable space of the manipulator illustrated in Fig. \ref{IJRR_reachable}, the circular trajectory radius is set to $R$ = 0.1 m, basically encompassing the range of dynamic operations. Meanwhile, the angular frequency progressively increases from 2.5 to 4.5 rad/s to ensure that the critical nonlinear dynamics can be fully excited. The nonlinear baseline controller is used as the dynamic controller for the aerial platform. Details are provided in Section  \ref{CC}.
As displayed in Fig. \ref{IJRR_CoM}(c), the aerial platform exhibits a significant position  offset of about 10 cm from the desired point during manipulator motion. Moreover, the state evolution of the CoM is illustrated in Fig. \ref{IJRR_Pc}. 

%% file: sections/03_perception.tex
\section{Onboard perception design}
Precise aerial grasping requires both the robot state for flight control and the target information for motion coordination. CAPTURER combines visual–inertial state estimation with fiducial-based target detection. The target measurements are transformed into a global frame and filtered to predict the target motion. When the target height is known, the target measurement can also be used to correct state drift.

\subsection{State estimation}\label{m_Accurate_state}

The realization of aerial manipulation in dynamic environments necessitates accurate and high-frequency onboard state estimation. While laser radar systems provide high-fidelity positioning, the significant payload limits scalability on compact aerial platforms \citep{scirobotics_Ladar}. Consequently, we utilize a commercial tracking camera T265 to provide a 200 Hz state feedback. However, in the cascaded control architecture, the attitude loop of the aerial platform operates at 500 Hz, rendering the camera's raw update rate insufficient for high-bandwidth rotational regulation. Furthermore, the communication latency between the host computer and the flight controller introduces phase lag.

To enhance perception performance, an Error-State Kalman Filter (ESKF) is implemented to fuse the camera's information with 500 Hz inertial data from an onboard MPU9050 IMU \citep{quaternionkinematicserrorstatekalman}.  The specific disturbance characteristics of aerial manipulators should be considered in the fusion process. Unlike standard quadrotors, the inertial signals of aerial manipulators are contaminated by dynamic coupling disturbances induced by the manipulator motion. Moreover, the yaw state estimation requires additional consideration since it is sensitive to electromagnetic perturbations from the onboard actuators and electronics. We assign a higher confidence weight to the vision-based yaw to ensure directional stability. The algorithmic details can be found in source code.

A direct comparison of the state estimation with respect to ground truth is provided to substantiate the usability and accuracy of the presented state estimation algorithm. Three agile trajectories are scheduled, including a lemniscate trajectory with a 4 m amplitude at speeds up to 3.2 m/s, a circular trajectory with a radius of 1.5 m at speeds up to 2.5 m/s, and a four-leaf trajectory with $T = 8$ s. Meanwhile, the manipulator is commanded to track a circular trajectory, mimicking  real-world dynamic operations. The fusion performance is assessed based on the deviation between the filtering results and ground truth. The mean absolute estimation error is defined as the metric. The state estimation results across three trajectories are provided in Fig. \ref{IJRR_estimation}.
Specifically, the mean state estimation errors are measured as 0.066 m for the circular trajectory, 0.055 m for the lemniscate trajectory, and 0.035 m for the four-leaf trajectory. The experimental data indicate that the proposed solution demonstrates reliable estimation performance.

\begin{figure*}[htbp]
	\centering
	\includegraphics[width=0.95\linewidth,height=\textheight,keepaspectratio]{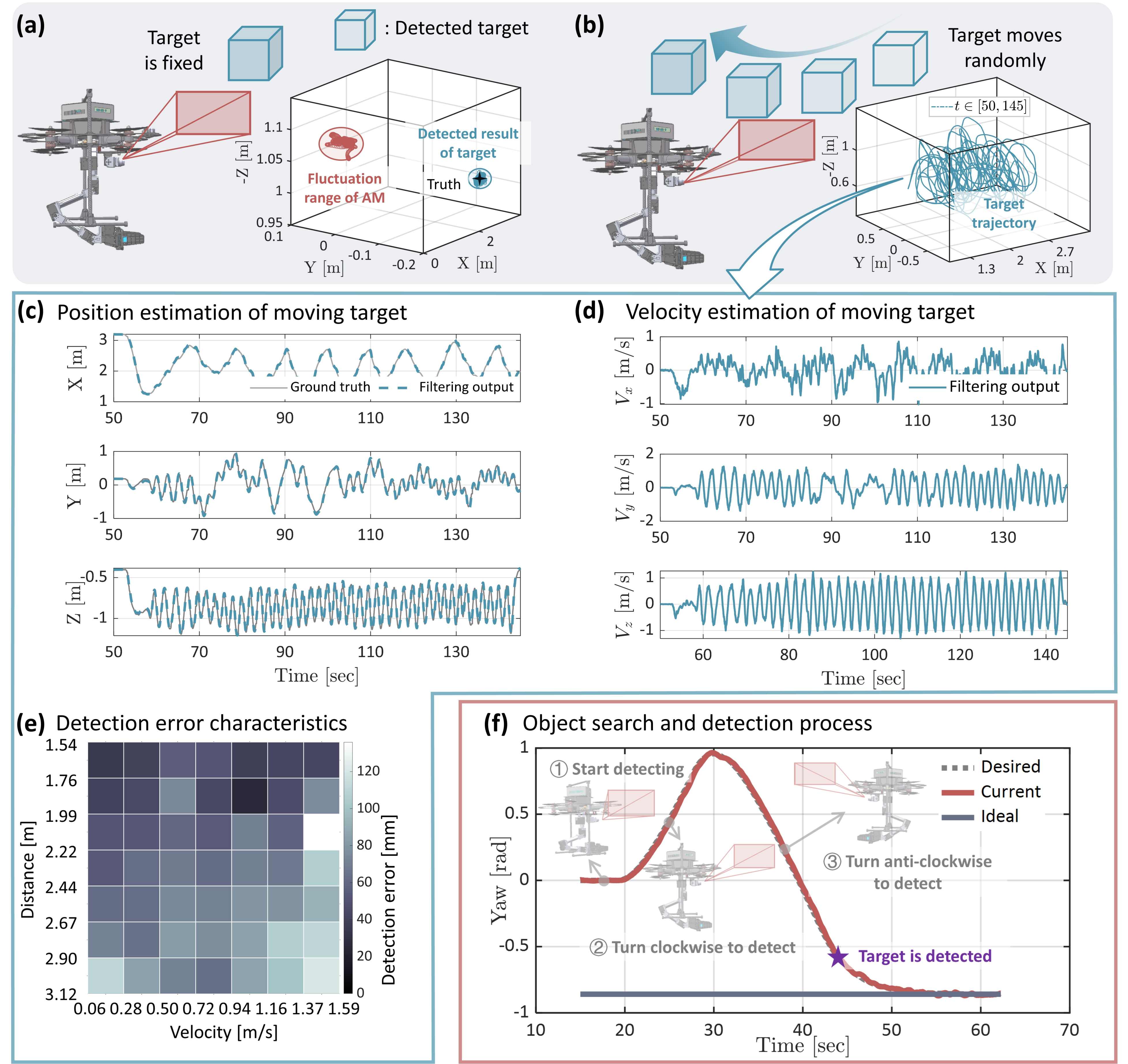}
	\caption{(a) CAPTURER detects a fixed target during flight. (b) CAPTURER
		detects a target that is moving randomly. (c) The position estimation results of the moving target. (d) The predicted velocity output of the moving
		target. (e)  Detection error characteristics. The horizontal axis
		represents the moving velocity of the target and the vertical axis denotes the distance between the
		detected target and CAPTURER. (f) The process of searching for a target outside the camera's FoV.}\label{IJRR_Detection}
\end{figure*}

\subsection{Target detection and motion prediction}\label{m_ODMP}
Different from object detection in fixed-base robotics, the base of aerial manipulator fluctuates. The platform motion significantly increases the difficulty of target detection, especially during dynamic aerial manipulation. To address these constraints, the AprilTag feature is used to enable 6-DoF state recognition from a single RGB image \citep{7759617}. The state information of the aerial platform is incorporated to provide a global pose estimation of the detected objects. The homogeneous transformation formula is expressed as $\bm{T^{O}_{I}}=\bm{T^{B}_{I}T^{C}_{B}T^{O}_{C}}$. The equation indicates that the position of the captured object in $\{\mathcal{I}\}$ is  determined by three elements: the UAV state in $\{\mathcal{I}\}$ ($\bm{T^{B}_{I}}$),  the state of the detection camera in  $\{\mathcal{B}\}$ ($\bm{T^{C}_{B}}$), and  the target state in $\{\mathcal{C}\}$ ($\bm{T^{O}_{C}}$).

Let $x_t=[p_t^\top,\dot{p}_t^\top,\ddot{p}_t^\top]^\top\in\mathbb{R}^{9}$. $p_t\in\mathbb{R}^{3}$ is the position portion of $\bm{T^{O}_{I}}$. The transition dynamics of the grasped object over a short-term horizon can be described as a discrete stochastic linear system of a point mass. The resulting formulation is expressed as:
\begin{align}
	x_{t,k+1}&=\bm{A_t}x_{t,k}+\bm{G_t}\xi_k,\qquad z_k=\bm{C_t}x_{t,k}+\zeta_k,\\
	\bm{A_t}&=\begin{bmatrix}\bm{I_3}&\Delta t\bm{I_3}&\frac12\Delta t^2\bm{I_3}\\\bm{0_3}&\bm{I_3}&\Delta t\bm{I_3}\\\bm{0_3}&\bm{0_3}&\bm{I_3}\end{bmatrix},\quad
	\bm{G_t}=\begin{bmatrix}\bm{0_3}\\\bm{0_3}\\\bm{I_3}\end{bmatrix},\\ \bm{C_t}&=[\bm{I_3}\;\bm{0_3}\;\bm{0_3}],
\end{align}
where $z_k\in\mathbb{R}^3$ represents the observed position of the target. $\Delta t\in\mathbb{R}$ is the propagation period. $\xi_k\in\mathbb{R}^3$ and $\zeta_k\in\mathbb{R}^3$ denote the process noise and measurement noise, respectively, which are assumed to follow a Gaussian distribution. 

By virtue of the established motion equation, the future state information of the captured object can be predicted using our filtering algorithm. In order to ensure robustness and consistency of the state estimation, the filtered state $x_t$ is updated at each step when a new measurement is available. If the state estimation is deemed invalid (either during the first iteration of the filtering process or due to a previously invalid detection), the filtering algorithm is reinitialized using the last measurement. Notably, the state estimation is invalidated if any of the following conditions occur:

$\bullet$ The number of initialization measurements falls below the specified threshold. In our work, the number threshold is set as 15. 

$\bullet$ The time interval since the last update exceeds the specified time threshold. In our work, the time threshold is set as 3~s. 

Three hierarchical trials are arranged to assess the capacity of the presented target detection and motion prediction scheme.

\textit{1) Stationary target detection}. The state information of a stationary target is detected while the aerial manipulator is required to hover. Due to the presence of multiple disturbances, the aerial manipulator experiences fluctuation. As shown in Fig. \ref{IJRR_Detection}(a), the state of the detected target can be accurately estimated even if the aerial manipulator fluctuates. The maximum error $\epsilon$, mean absolute error $\mu$, and standard deviation $\varrho$ are denoted as the evaluation indices. By resorting to the defined indices, the measures of $\epsilon$, $\mu$, and $\varrho$ are 1.6 cm, 0.77 cm, and 0.34 cm, respectively. The motion capture system (NOKOV) provides the ground truth of the identified target for verifying estimation accuracy. 

\textit{2) Dynamic motion prediction}. The objective of Trial II is to assess the capability of predicting the motion information when the target moves randomly. Fig. \ref{IJRR_Detection}(b) highlights the state prediction capability of our filtering scheme. Compared to static target detection, additional challenges are involved when predicting the motion of a moving target. The incorporation of a forgetting factor in the motion prediction process improves the ability of tracking  dynamic objects. The predicted state information of the dynamic target, including its position and velocity, is illustrated in Figs. \ref{IJRR_Detection}(c-d). Regarding the position estimation, the measure of $\epsilon$ is 6.60 cm even if the object moves randomly.  As can be seen in Fig. \ref{IJRR_Detection}(e), the detection distance and the motion velocity of the target have a significant influence on the state prediction capability for the dynamic object. It is worth noting that the position accuracy may degrade rapidly as the distance between the target and our aerial manipulator increases, indicating that the manipulation range should be constrained.

\textit{3)  Active yaw search and localization}.  The aim of Trial III is to examine the efficacy of actively detecting the target. The aerial manipulator is endowed with the capability of turning counterclockwise to search  for the target in the left half plane when the target is not detected in the right half plane. Once the target is firmly in the camera's FoV, the expected yaw signal is promptly calculated based on the predicted target information to ensure that the target is in the center of FoV. Moreover, a first-order filter approach is used to smooth the desired yaw output after detecting the target. For comparison, the ideal yaw signal is derived on the basis of the target's position in $\{\mathcal{I}\}$ provided by the motion capture system. Fig. \ref{IJRR_Detection}(f) illustrates the entire process of detecting the target. The yaw detection error is less than 0.3$^{\circ}$.

\subsection{Vertical drift correction}\label{m_S_dri}
Different from autonomous exploration using onboard sensors \citep{Swarm,9102390}, it is nontrivial to grasp a pen-sized object in unstructured environments. The primary obstacle stems from the drift error caused by variations in lighting and texture \citep{7747236}. The drift error refers to an inevitable discrepancy between the actual position and the perceived position determined by onboard sensors. With respect to those tasks that do not require precise operations, the positioning drift is generally overlooked \citep{Champion_level,abh1221,9102390}. For instance, a static box can be successfully grasped with the assistance of a soft gripper with decimeter-level error tolerance \citep{npjrobotics}. However, this strategy is difficult to apply to active grasping of a pen-sized target, given the stringent accuracy requirement. In other words, the drift error must be tackled to enable precise aerial grasping. Therefore, a target-aided state correction strategy is developed to handle the drift across different scenarios. As illustrated in Fig. \ref{IJRR_frame}, the position of the grasped target in the world-fixed frame can be formulated as:
\begin{equation}
	p_{t} = p_b+\bm{R_b}(p_c^{b}+\bm{R_c}p_t^{c}),
	\label{correction}
\end{equation} 
where  $p_c^{b}\in\mathbb{R}^{3\times1}$ represents the constant position of the detection camera in $\{\mathcal{B}\}$, which can be precisely obtained by extrinsic calibration. $\bm{R_c}\in SO(3)$ denotes the attitude rotation matrix from $\{\mathcal{C}\}$ to $\{\mathcal{B}\}$, which is determined by the pitch angle of the camera. The calculation accuracy of  $\bm{R_c}$ can be guaranteed since the pitch angle is controlled by the servo actuator XL330-M228 with 0.088$^{\circ}$ resolution. Moreover,  $p_t^{c}\in\mathbb{R}^{3\times1}$ is the position of the grasped target in $\{\mathcal{C}\}$. Therefore, Eq. (\ref{correction}) can be utilized to compensate the drift error  if the actual height of the captured target is known. The presented hypothesis is reasonable in practical aerial manipulation tasks, particularly when interacting with infrastructure, workpieces, or task objects placed at prescribed locations. The target height provides an absolute geometric constraint for correcting the vertical localization drift of the aerial platform. Subsequently, the estimated flight altitude of the aerial platform can be retrieved as:
\begin{equation}
	\hat{p}_b(z) = p_{t}(z) - \bm{R_b}\left(p_c^{b}+\bm{R_c}p_t^{c})\right)(z),
	\label{corr_esti}
\end{equation} 
where $p_{t}(z)\in\mathbb{R}$ is defined as the vertical drift correction ratio. Therefore, the ratio $\upsilon = \frac{\hat{p}_b(z)}{{p}_b(z)}$ represents the magnitude of the correction ratio. Fig. \ref{IJRR_grasp_V2_5}(a) shows the estimation results of drift errors over different time periods when capturing the pen-sized object in an open lawn.
Unlike previous studies \citep{npjrobotics,10328688}, the developed state correction process significantly reduces the reliance on absolute state information of the aerial manipulator, thereby avoiding the cost and mobility limitations associated with external localization systems. Moreover, the same principle can be directly extended to the horizontal directions.

%% file: sections/04_planning.tex
\section{Cooperative motion planning}\label{CMP}
Although existing studies have demonstrated cooperative trajectory generation, two limitations remain \citep{10224341,npjrobotics,10156288,CAO2022107607,6608869,Behavioral,8283709,7962217,LAI2022102724,7017672,8796272,9194145,IJARS,7762762}. First, the motion trajectory is calculated offline after obtaining the target position. The offline optimization strategy reduces the dexterity of aerial manipulators and makes it problematic in responding to the changing target \citep{10224341,npjrobotics}. More importantly, they lack a comprehensive analysis of motion characteristics and incur a heavy computational overhead \citep{10224341,CAO2022107607,6608869,Behavioral,8283709,7962217,LAI2022102724,7017672,8796272,9194145,IJARS,7762762}.

Attempting to fill this gap, we formulate a cooperative planning strategy to simultaneously consider both the mobility of the aerial platform and the flexibility of the manipulator. Instead of relying on manual setup of the interaction process, the grasping trajectory is generated online adaptively according to real-time motion states in our work. Specifically, a kinematic-level optimization-based predictive model is formulated with a 12-dimensional motion state and a 6-dimensional control input. In addition, multiple practical constraints are incorporated into the optimization formulation. The state propagation is first established as follows. 

Let $w_k=\left[p_b^{\top},\dot{p}_b^{\top},p_{ed}^{\top},\dot{p}_{ed}^{\top}\right]_k^{\top}\in\mathbb{R}^{12\times1}$ denote the predicted state at the $k$-th step. Its discrete-time propagation is expressed as
\begin{equation}
	\left\{
	\begin{aligned}
		w_{k+1}&=\bm{A_p}w_k+\bm{B_p}u_k,\\
		o_{k+1}&=\bm{C_p}w_{k+1},
	\end{aligned}
	\right.
\end{equation}
where $u_k=\left[\ddot{p}_b^{\top},\ddot{p}_{ed}^{\top}\right]_k^{\top}\in\mathbb{R}^{6\times1}$ is the optimized control input. It consists of the acceleration of the aerial platform in the world-fixed frame and that of the end-effector relative to the manipulator base. The predicted output is given by
\begin{equation}
	o_{k+1}
	=
	\left[
	(y^{pu})^{\top},
	(y^{vu})^{\top},
	(y^{pm})^{\top},
	(y^{vm})^{\top}
	\right]_{k+1}^{\top}
	\in\mathbb{R}^{12\times1},
\end{equation}
where $y^{pu}\in\mathbb{R}^{3}$ and $y^{vu}\in\mathbb{R}^{3}$ denote the predicted position and velocity of the aerial platform in the world-fixed frame, while $y^{pm}\in\mathbb{R}^{3}$ and $y^{vm}\in\mathbb{R}^{3}$ denote those of the end-effector relative to the manipulator base. The propagation matrices are provided in Appendix~\ref{DPM}.

Based on the predicted states, the end-effector position in the world-fixed frame is calculated as
\begin{equation}
	\hat{p}_e(t+k)
	=
	y^{pu}(t+k)
	+
	\bm{R_b}(t+k)
	\left(
	p_o+y^{pm}(t+k)
	\right).
\end{equation}

Previous planning methods \citep{10224341,10156288} assume that the attitude rotation matrix remains constant over the prediction horizon, i.e., $\bm{R_b}(t+k)=\bm{R_b}(t)$. Such an assumption neglects the rotational motion of the aerial platform during prediction. To account for this motion, the measured angular velocity $\omega_b(t)$ is used to propagate the platform attitude over the prediction horizon and is updated at each replanning cycle. The attitude propagation is expressed as
\begin{equation}
	\bm{R_b}(t+k)
	=
	\bm{R_b}(t+k-1)
	\exp\!\left(
	\delta t\,\omega_b^\times(t)
	\right),
\end{equation}
where $\delta t>0$ denotes the prediction interval. The formulation explicitly incorporates the rotational motion of the aerial platform into the predicted end-effector position. The end-effector tracking errors over the prediction horizon are defined and stacked as:
\begin{subequations}
	\begin{align}
		&e_p(t+k)
		=
		p_e^d(t+k)-\hat{p}_e(t+k),\\
		&E_P
		=
		\left[
		e_p^{\top}(t+1),
		\ldots,
		e_p^{\top}(t+N)
		\right]^{\top}
		\in\mathbb{R}^{3N\times1}.
	\end{align}
\end{subequations}

The corresponding tracking cost is defined as:
\begin{equation}
	Q_1
	=
	\Vert E_P\Vert^{2}_{\bm{\Lambda_1}}
	=
	E_P^{\top}\bm{\Lambda_1}E_P,
	\label{Q1V1}
\end{equation}
where $\bm{\Lambda_1}\in\mathbb{R}^{3N\times3N}$ is a positive-definite weighting matrix.

When no additional feasibility penalty or inequality constraint is activated, the above tracking cost retains an unconstrained  optimization form. However, tracking accuracy alone does not account for motion smoothness, cooperative motion allocation, or physical feasibility. Therefore, specific constraints are introduced in the following formulation.

\noindent{{\textbf{Maximizing control smoothness:}}}
To avoid abrupt variations in the optimized acceleration commands, the increment in the control input is penalized over the prediction horizon. The corresponding cost function is defined as:
\begin{equation}
	Q_2
	=
	\Vert \Delta u\Vert^{2}_{\bm{\Lambda_2}}
	=
	\Delta u^{\top}\bm{\Lambda_2}\Delta u,
	\quad
	\Delta u\in\mathbb{R}^{6N\times1},
	\label{Q2}
\end{equation}
where $\bm{\Lambda_2}\in\mathbb{R}^{6N\times6N}$ is the weighting matrix and $\Delta u$ denotes the stacked increments of the control input over the prediction horizon. The  cost suppresses rapid changes in the optimized acceleration commands and improves the smoothness of the generated cooperative motion.

\noindent{{\textbf{Control action regulation:}}} 
A practical optimization problem has been constructed to guide the model outputs from the above two aspects: minimizing tracking error and maximizing control smoothness \citep{limit}. Nevertheless, a critical drawback remains in this formulation. 
The optimization solution does not account for the substantial differences in motion characteristics between the aerial platform and the manipulator. Therefore, a dynamic parameter $\kappa$ is introduced to effectively regulate the state variables.
\begin{equation}
	\kappa = \left\{  
	\begin{aligned} 
		1, \quad &\Vert d(h_{i})\Vert > d_{0}  \\  
		0, \quad &\Vert d(h_{i})\Vert \leq d_{0} 
	\end{aligned} 
	\right.  ,
\end{equation}
where $d(h_{i})$ denotes the current distance between the end-effector and the grasped target. $d_0\in\mathbb{R}$ represents the critical distance, which depends on the state of the aerial manipulator and the relative motion. Moreover, the weighting matrix $\bm{\Lambda_{2}}$ is updated online from the perspective of behavioral adaptation, which can be defined as $\bm{\Lambda_{2}} = \kappa\bm{\Lambda_{2}^{u}} + (1-\kappa)\bm{\Lambda_{2}^{m}}$. $\bm{\Lambda_{2}^{u}}\in\mathbb{R}^{6N\times6N}$ and $\bm{\Lambda_{2}^{m}}\in\mathbb{R}^{6N\times6N}$ represent the weighting matrices corresponding to different capture stages. The involvement of $\kappa$ signifies that when the aerial manipulator is far away from the target, the state of the aerial platform is prioritized and the manipulator remains in its initial configuration. Once the target enters the effective workspace, the flexibility of the manipulator is  harnessed to capture the target while the aerial platform maintains its motion velocity. 

\noindent{{\textbf{Motion regularization:}}} 
Although tracking accuracy and action smoothness are considered in the preceding cost, the predicted state increment of the aerial platform and manipulator are not explicitly constrained. As illustrated in Fig. \ref{IJRR_reachable}, similar end-effector tracking performance can be achieved by different combinations of the platform and manipulator motions. Without additional  motion regularization, unnecessary oscillatory behavior between the two components may be generated. Let
$\Delta y_p(t+k)
=
\left[
\left(\Delta y^{pu}(t+k)\right)^{\top},
\left(\Delta y^{pm}(t+k)\right)^{\top}
\right]^{\top}
\in\mathbb{R}^{6\times1}$,
where $\Delta y^{pu}(t+k)$ and $\Delta y^{pm}(t+k)$ denote the predicted position increments of the aerial platform and end-effector at the $k$-th prediction step, respectively. 

In order to improve the continuity of the cooperative trajectory, the position increments over the prediction horizon are stacked as:
\begin{equation}
	\Delta Y_p
	=
	\left[
	\Delta y_p^{\top}(t+1),
	\ldots,
	\Delta y_p^{\top}(t+N)
	\right]^{\top}
	\in\mathbb{R}^{6N\times1}.
\end{equation}
The corresponding motion regularization cost is defined as:
\begin{equation}
	Q_3
	=
	\Vert \Delta Y_p \Vert^{2}_{\bm{\Lambda_3}}
	=
	\Delta Y_p^{\top}
	\bm{\Lambda_3}
	\Delta Y_p,
\end{equation}
where $\bm{\Lambda_3}\in\mathbb{R}^{6N\times6N}$ is the weighting matrix. Through the developed regularization process, excessive variations in the cooperative motion can be suppressed.

\noindent{{\textbf{Dynamic feasibility:}}} For the sake of feasibility, dynamic constraints should be satisfied throughout the entire prediction horizon. In our work, the amplitudes of motion velocity and acceleration are limited by imposing a penalty if they exceed corresponding physical thresholds. Specifically, the following feasible region is incorporated as:
\begin{subequations}
	\begin{align}
		&Q_{4}^{vu} = \sum_{k=1}^{N}\max\{ ||y^{vu}(t+k)||^{2}-{{\dot{p}}_{b\max}}^{2},0\},\\
		&Q_{4}^{vm} = \sum_{k=1}^{N}\max\{ ||y^{vm}(t+k)||^{2}-{{\dot{p}}_{ed\max}}^{2},0\},\\
		&Q_{4}^{au} = \sum_{k=1}^{N}\max\{ ||\ddot{p}_{b}(t+k)||^{2}-{{\ddot{p}}_{b\max}}^{2},0\},\\
		&Q_{4}^{am} = \sum_{k=1}^{N}\max\{ ||\ddot{p}_{ed}(t+k)||^{2}-{{\ddot{p}}_{ed\max}}^{2},0\},\\
		&Q_4 = \lambda_{4,u}(Q_{4}^{vu} + Q_{4}^{au}) + \lambda_{4,m}(Q_{4}^{vm} + Q_{4}^{am}),
	\end{align}\label{dfV1}
\end{subequations}
where $\dot{p}_{b\max}\in\mathbb{R}$ and $\ddot{p}_{b\max}\in\mathbb{R}$ denote the maximum allowable velocity and acceleration magnitudes of the aerial platform, respectively. Similarly, $\dot{p}_{ed\max}\in\mathbb{R}$ and $\ddot{p}_{ed\max}\in\mathbb{R}$ denote the corresponding maximum allowable velocity and acceleration magnitudes of the end-effector. We directly sum $Q_{4}^{vu}$ and $Q_{4}^{au}$ due to the fact that they share similar magnitudes \citep{Swarm}. In line with the same principle, $Q_{4}^{vm}$ and $Q_{4}^{am}$ are directly summed. Two weighting parameters ($\lambda_{4,u}$ and $\lambda_{4,m}$) are utilized to regulate the dynamic cost function, considering the significant difference in the motion characteristics between the moving platform and the manipulator.

\noindent{{\textbf{Kinematic feasibility:}}} Unlike conventional fixed-base manipulators, the manipulator is mounted beneath the aerial platform. As shown in Fig. \ref{IJRR_reachable}, the reachable workspace of the end-effector is constrained by the propellers, outer frame, and joint configuration.  In order to ensure both effectiveness and safety,  a kinematic feasibility penalty is introduced as:
\begin{equation}
	Q_5
	=
	\lambda_{5}
	\sum_{k=1}^{N}
	\min\left\{
	d\!\left(y^{pm}(t+k)\right)-d_m,
	0
	\right\}^{2},
	\label{kfmaV1}
\end{equation}
where $\lambda_{5}>0$ is the weighting parameter and $d_m>0$ denotes the prescribed safety margin. The function $d(y^{pm}(t+k))$ represents the minimum distance between the end-effector position and the workspace boundary. The penalty term maintains zero when the manipulator operates in the safe space. Furthermore, the cost term increases rapidly to avoid collision once the manipulator approaches the boundary of the reachable workspace.

\noindent{{\textbf{Actuators limitation:}}}  The increment of control actions is restricted to a feasible region for safe manipulation. An additional constraint on actuators is considered as follows:
\begin{equation}
	\Delta u_{\min}\preceq \Delta u \preceq \Delta u_{\max},
\end{equation}
where $\Delta u_{\min}\in\mathbb{R}^{6N\times1}$ and $\Delta u_{\max}\in\mathbb{R}^{6N\times1}$ represent the minimum and maximum allowable increment of control actions. 

The overall cooperative motion planning problem is formulated as
\begin{equation}
	\min_{u}Q=\min_{u}(Q_1+Q_2+Q_3+Q_4+Q_5)
	\label{overall_optimization}
\end{equation}
subject to
\begin{equation}
	\left\{\begin{matrix}w_{k+1}=\bm{A_p}w_k+\bm{B_p}u_k\\o_{k+1}=\bm{C_p}w_{k+1}\\
		\Delta u_{min}\preceq \Delta u \preceq \Delta u_{max},
	\end{matrix}\right.,
\end{equation}
$Q_1$ minimizes the end-effector tracking error, $Q_2$ penalizes the control actions, $Q_3$ improves the continuity of the cooperative motion, while $Q_4$ and $Q_5$ account for dynamic and kinematic feasibility, respectively. The actuator limits are explicitly imposed through the bounds on $\Delta u$.  Moreover, slack variables are introduced for the dynamic and kinematic feasibility constraints to retain the optimization structure in the online implementation. Finally, the resulting constrained optimization problem is implemented in C++11 and solved using the qpOASES library.

The prediction horizon is set to $N=15$ with a discretization interval of $\delta t=0.07$~s. The optimization can be solved online on the onboard Intel NUC11PAH at a frequency of 25~Hz, enabling real-time grasping. Subsequently, the continuous velocity signal of the aerial manipulator is obtained by integrating the optimized control actions. The procedure ensures that the manipulator joints operate smoothly, which is beneficial for executing dynamic tasks on a moving aerial platform. The effectiveness of the proposed cooperative planning method is further verified through aerial grasping experiments under different operating conditions in Section~\ref{AGADE}. Fig. \ref{IJRR_planning_time} shows the computation time of the proposed cooperative planning method during aerial grasping.

\begin{Remark}
	While the feasibility has been penalized in Eqs. (\ref{dfV1}) and (\ref{kfmaV1}), there is still a possibility of generating an infeasible solution. Therefore, an iterative refinement approach \citep{9102390} is employed to check the planned trajectory in this work. If the optimized solution lies outside the safe boundary, the corresponding penalty weighting parameters ($\lambda_{4,u}$, $\lambda_{4,m}$, and $\lambda_{5}$) are increased to solve the optimization problem again. Consequently, the safety of the optimized solution is enhanced with each iteration, ensuring that the aerial manipulator satisfies both kinematic and dynamic constraints across the entire phase. 	
\end{Remark}

%% file: sections/05_control.tex
\section{Anti-disturbance dynamic controller}\label{m_controller}

\subsection{Estimation of dynamic coupling effects using a robust differentiator}
Eqs. \eqref{coupling_force} and \eqref{coupling_torque} show that the dynamic coupling disturbances depend on the system inertial parameters, states, and their derivatives. Nevertheless, it is intractable to accurately get the coupling effects even if the detailed analytical expressions are provided in Eqs. (\ref{coupling_force})-(\ref{coupling_torque}). A critical challenge lies in accurately estimating the coupling force and torque of aerial manipulators. The direct measurement of high-order state information is extremely difficult for normal actuators. To intentionally circumvent the issue, only the first-order term of the dynamic coupling effects is typically taken into account \citep{10083097,7353623}. Nevertheless, the neglected higher-order terms play a significant role when performing dynamic operations. 

As shown in Fig. \ref{IJRR_PoE}, a serial manipulator driven by five XM430-W350 servo actuators is adopted to achieve aerial operations. The joint position $q$ and joint velocity $\dot{q}$ are obtained from encoder feedback at resolutions of 0.088~$^{\circ}$ and 0.229~rpm, respectively. By recalling Eqs. (\ref{coupling_force})-(\ref{coupling_torque}), all variables associated with the coupling disturbances are measurable except for ${\ddot{p}}_{bm}^B$ and $\dot{\omega}_{b}$. Therefore, the estimation of coupling disturbances can be reconstructed as a problem of precisely extracting unmeasured signals. Driven by the above analysis, a robust differentiator (RD) based on super-twist theory is adopted to acquire unmeasured signals, which can be formulated as:
\begin{equation}
	\left\{\begin{matrix}\dot{g}_1=\frac{k_{g1}}{k_{g3}}\sqrt{\exp(k_{g3}^2\left|s-g_1\right|)-1}\cdot 
		\operatorname{sgn}(s-g_1)+g_2\\
		\dot{g}_2=k_{g2}\exp(k_{g3}^2\left|s-g_1\right|)\cdot \operatorname{sgn}(s-g_1)\\
		\dot{s}=g_2\\\end{matrix}\right.,
\end{equation}
with auxiliary variables $g_1\in\mathbb{R}$ and $g_2\in\mathbb{R}$. $\dot{s}\in\mathbb{R}$ is the required differential signal. The principle of tuning parameters $k_{g1}\in\mathbb{R}$,$ k_{g2}\in\mathbb{R}$, and $k_{g3}\in\mathbb{R}$ is detailed in \citet{SEEBER2021109858}. 
For a signal $s$ with a bounded second derivative, the real-time data processing approach provides a derivative estimate that converges uniformly to the true derivative within a known finite time. The prior knowledge of convergence time provides a potential in enhancing the dynamic performance by adjusting the controller turning time in advance. 

The estimation performance of the RD is compared with two commonly used differentiation approaches. First, the general discrete differentiator (DD) is chosen as a comparison, which can be expressed as $\dot{s}\left(t\right)=\left({s\left(t\right)-s\left(t-3t_s\right)}\right)/{3t_s}$
with sampling period $t_s$. Apart from the differentiator approach, the popular Luenberger observer (LO) is selected to estimate the signal derivative within the assumption that the CoM motion of the manipulator is linearly driven. Specifically, the unmeasured signal ${\ddot{p}}_{bm}^B$ is defined as the observed state. 

\begin{figure}[tbp]
	\centering
	\includegraphics[width=0.98\linewidth,height=\textheight,keepaspectratio]{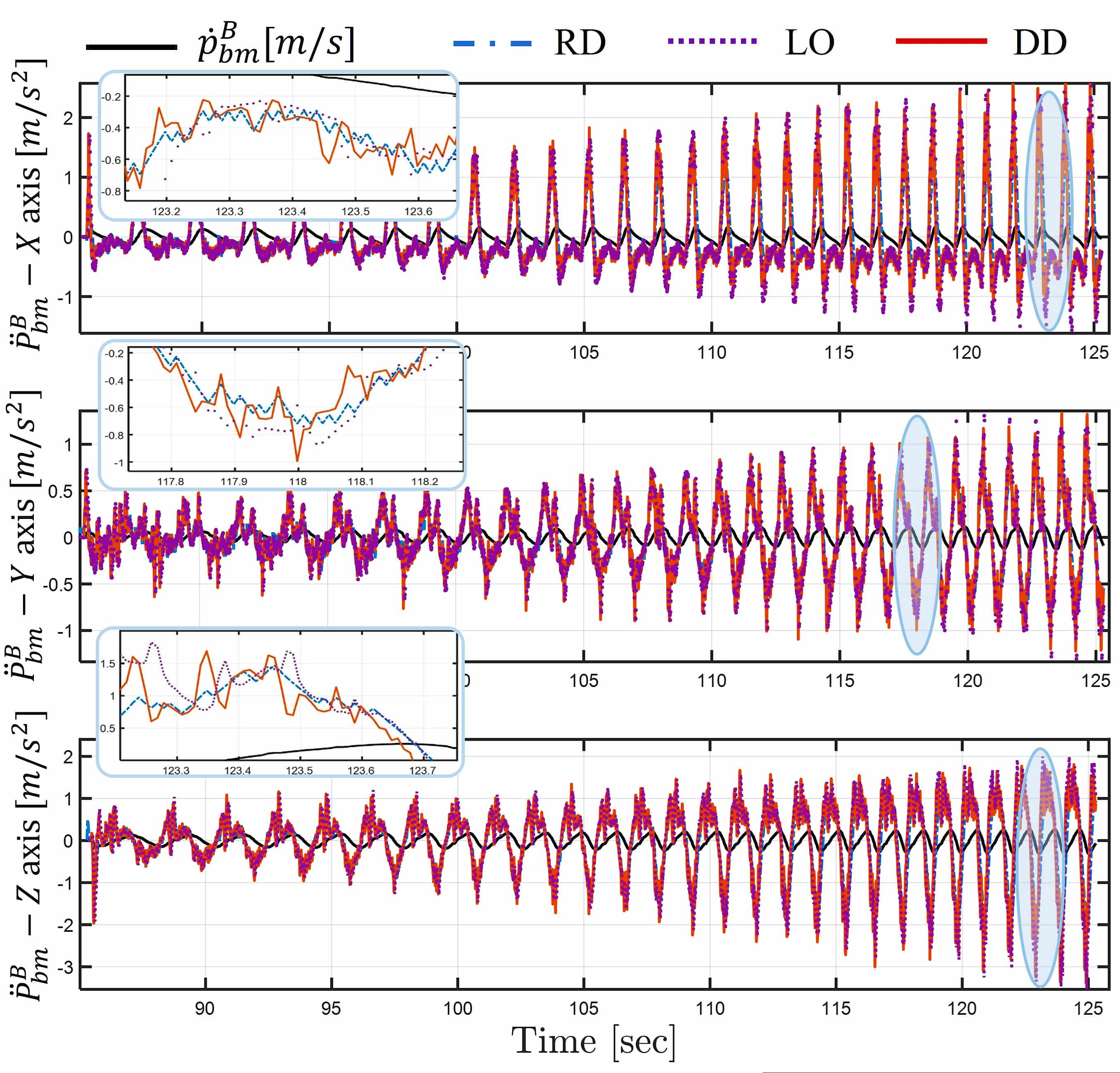}
	\caption{Estimation of $\ddot{P}_{bm}^{B}$ using different differentiation methods.}\label{IJRR_ddot_Pbm}
\end{figure}

The three approaches are evaluated while the manipulator tracks the three-dimensional periodic trajectory defined in Section~\ref{coupling_analysis}. The corresponding estimates of $\ddot{p}_{bm}^{B}$ are shown in Fig.~\ref{IJRR_ddot_Pbm}. Although the signal is smoothed by averaging several recent samples, the resulting derivative estimation remains noisy. As shown in Fig.~\ref{IJRR_ddot_Pbm}, the employed RD yields an accurate derivative estimation with reduced noise sensitivity. The performance of the LO can approach that of the RD after careful pole tuning. Driven by comprehensive analysis, RD is more suitable for estimating the required high-order signals on account of its provable finite-time convergence property and explicit parameter-selection guidelines \citep{SEEBER2021109858}. Similarly, the unmeasured signal ${\dot{\omega}}_b$ can be precisely acquired in the same data processing scheme.

\begin{Remark}
	Two alternative approaches can be used to obtain ${\ddot{p}}_{bm}^B$. One is that the signal $\ddot{q}$ is obtained by filtering $\dot{q}$. Then, ${\ddot{p}}_{bm}^B$ is calculated using the deduced second-order nonlinear mapping relation between $\ddot{q}$ and ${\ddot{p}}_{bm}^B$. Another is that the signal ${\dot{p}}_{bm}^B$ is directly processed to get its derivative, which is exploited in our study. The second approach is adopted because filtering errors in \(\ddot q\) may be amplified through the second-order nonlinear mapping.
\end{Remark}

\begin{figure}[tbp]
	\centering
	\includegraphics[width=0.98\linewidth,height=\textheight,keepaspectratio]{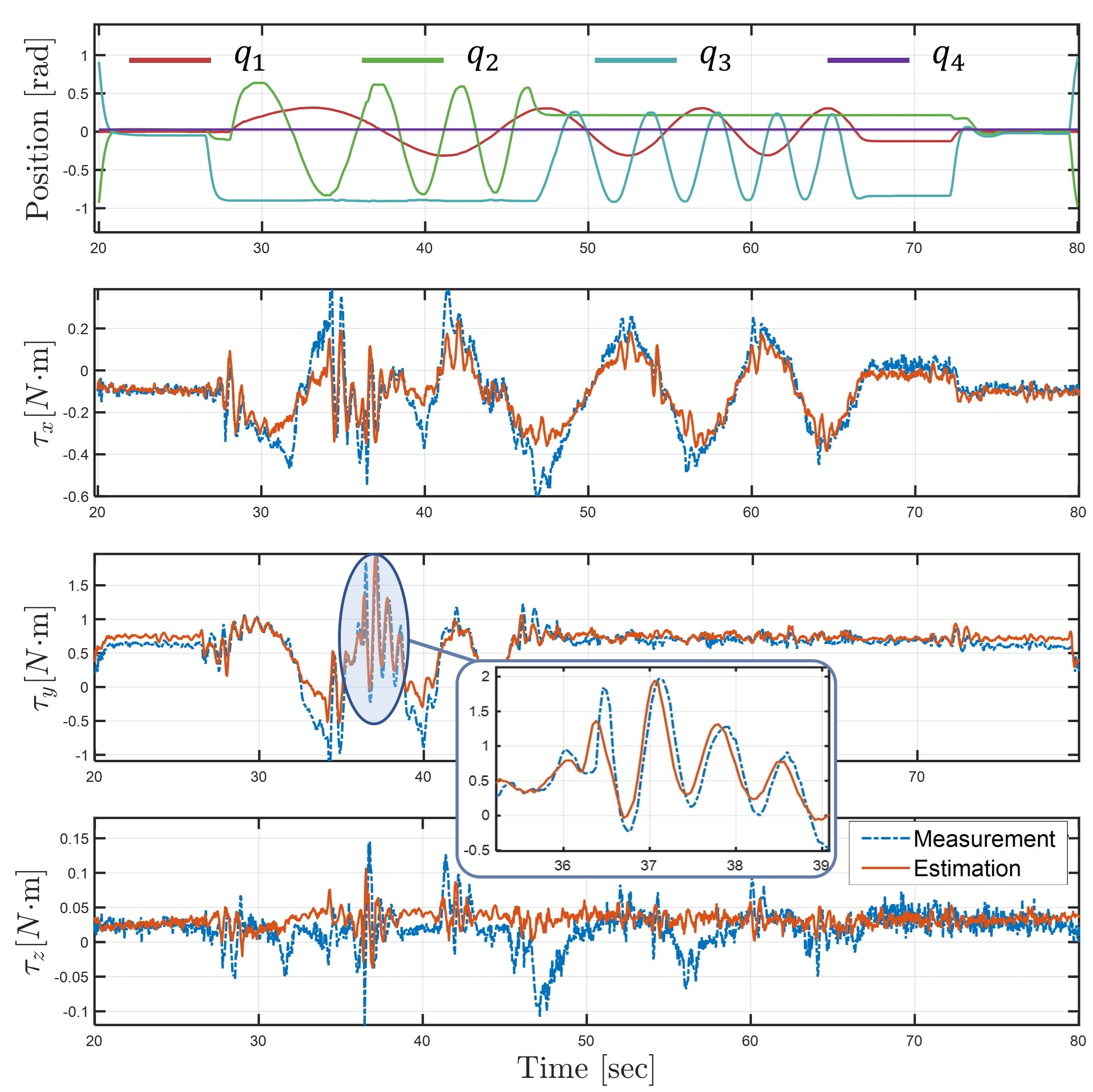}
	\caption{Estimation results of the coupling torque.}\label{IJRR_coupling_estimation}
\end{figure}
A further experiment is conducted to evaluate the reconstruction of the coupling disturbances during manipulator motion. The manipulator is commanded to track the prescribed trajectory, and the corresponding joint motions are shown in Fig.~\ref{IJRR_coupling_estimation}. A torque sensor (HPS-FT025) with 0.001 N$\cdot$m resolution is installed at the base of the manipulator to measure the coupling effects. As can be seen in Fig. \ref{IJRR_coupling_estimation}, the established full-state dynamic coupling model effectively captures the critical nonlinear disturbances by resorting to the exploited data processing method. Therefore, the estimated coupling effects can be directly fed into the control loop to improve the tracking performance of the aerial platform. 

Nevertheless, discrepancies remain between the estimated and measured coupling disturbances. These errors mainly arise from two sources. First, the measuring point of the torque sensor cannot be perfectly aligned with the nominal manipulator-base origin because of installation tolerances. Moreover, accurately acquiring the inertia parameters proves to be a challenging task. The assumption that the simulation model perfectly aligns with reality is infeasible in practical engineering due to the presence of motors, reducers, and other electronic components \citep{7158852,9919418}. The estimated parameters obtained via dynamic parameter identification methods cannot be guaranteed to faultlessly match the real values \citep{8772145,9919418}. Moreover, the downwash airflow caused by the rotation of the propeller blades can also induce external disturbances to the manipulator. Therefore, considering the estimation error induced by inertial parameter uncertainties and external disturbances, the coupling effects can be rewritten as:
\begin{equation}
	F_{m} = \hat{F}_{m} + \Delta_{F}, 	\tau_{m} = \hat{\tau}_{m} + \Delta_{\tau},\label{established_coupling}
\end{equation}
where $\hat{F}_m\in\mathbb{R}^3$ and $\hat{\tau}_m\in\mathbb{R}^3$ denote the modeled coupling force and torque used for feedforward compensation. $\Delta_F\in\mathbb{R}^3$ and $\Delta_\tau\in\mathbb{R}^3$ represent the corresponding residual disturbances, whose first-order derivatives are assumed to be bounded \citep{Proximal, Kim_IJRR}. To further suppress these uncompensated effects, two residual disturbance observers are introduced in the subsequent controller design.

\subsection{UAV dynamic controller}
Using the estimated coupling disturbances for feedforward compensation, the proposed robust controller for the aerial platform consists of three components: a DO-based control scheme in the translational loop, an attitude generation unit using Hopf Fibration \citep{Hopf}, and a NDO-based geometric control scheme in the rotational loop.

\subsubsection{Translational loop}
A DO-based backstepping controller is developed for the translational dynamics.

\textit{Step 1:} Let $p_b^d\in\mathbb{R}^3$ denote the desired trajectory of the aerial platform in $\{\mathcal{I}\}$. $z_1=p_b-p_b^d$ is the position tracking error. Consider the following quadratic form:
\begin{equation}
	V_1=\frac{1}{2}z_1^{\top}z_1.
\end{equation}

Subsequently, the derivative of $V_1$ can be obtained as:
\begin{equation}
	\dot{V}_1=z_{1}^{\top}\left(v_b-{\dot{p}}_b^d\right).
\end{equation}

\textit{Step 2:} Let $z_2=v_b-\varpi_1$. $\varpi_1=-\bm{k_1}z_1+{\dot{p}}_b^d$ represents the designed stabilizing parameter and $\bm{k_1}\in\mathbb{R}^{3\times3}$ is the positive diagonal gain matrix. This step aims at forcing $v_b$ to track $\varpi_1$. By substituting $\varpi_{1}$ into ${\dot{V}}_1$, we have 
\begin{equation}
	{\dot{V}}_1=z_1^{\top}z_2+z_1^{\top}\varpi_1-z_1^{\top}{\dot{p}}_b^d.
\end{equation}

The following Lyapunov function is chosen as:
\begin{equation}
	V_2=V_1+\frac{1}{2}z_2^{\top}z_2. \label{V2}
\end{equation}

Subsequently, differentiating Eq. (\ref{V2}) gives 
\begin{equation}
	\begin{aligned} 
		{\dot{V}}_2 &={\dot{V}}_1+z_2^{\top}{\dot{z}}_2\\ &=-z_1^{\top}\bm{k_1}z_1+z_2^{\top}\left(z_1+ge_3-\frac{fb_3}{m_s}+\frac{F_m}{m_s}-\dot{\varpi}_1\right).
	\end{aligned} 
\end{equation}

By resorting to Eq. \eqref{established_coupling}, the translational controller can be formulated as:
\begin{equation}
	f{b_3}={m_s}\left(ge_3+\bm{k_2}z_2+z_1-\dot{\varpi}_1\right)+{{\hat{F}}_m+{\hat{\Delta}}_F},
	\label{f_total}
\end{equation}
where ${\textbf{\textit{k}}}_2\in\mathbb{R}^{3\times3}$ is the positive diagonal gain matrix and ${\hat{\Delta}}_F$ is the estimation of $\Delta F$. 

Let $\tilde{\Delta}_{F}=\Delta_F-\hat{\Delta}_F$ be the estimation error of the DO. It can be verified that
\begin{equation}
	{\dot{V}}_2=-z_1^{\top}\bm{k_1}z_1-z_2^{\top}\bm{k_2}z_2+\frac{z_2^{\top}\tilde{\Delta}_F}{m_s}.
\end{equation}

\textit{Step 3:} By incorporating auxiliary variable $\epsilon$, the DO to estimate $\Delta_F$ can be constructed as:
\begin{equation}
	\left\{\begin{matrix}{\hat{\Delta}}_F=m_s\left(\epsilon+\bm{L_p}v_b\right)\\\dot{\epsilon}=-\bm{L_p}\left(\epsilon+\bm{L_p}v_b\right)+\bm{L_p}\left(\frac{fb_3}{m_s}-ge_3-\frac{{\hat{F}}_m}{m_s}\right)\\\end{matrix}\right.,
\end{equation}
where $\bm{L_p}\in\mathbb{R}^{3\times3}$ denotes the positive observer gain matrix. By integrating Eq. (\ref{f_total}), the estimation error dynamics can be formulated as:
\begin{equation}  
	\begin{aligned} 
		\dot{\tilde{\Delta}}_{F}& = \dot{\Delta}_{F} - \dot{\hat{\Delta}}_{F} = \dot{\Delta}_{F}-m_{s}\left(\dot{\epsilon}+\bm{L_p}\dot{v}_b\right) \\
		& = \dot{\Delta}_{F} + \bm{L_p}\hat{\Delta}_{F} - \bm{L_p}\Delta_{F} = \dot{\Delta}_{F} - \bm{L_p}{\tilde{\Delta}}_{F}.
	\end{aligned}  
\end{equation}

Consider the following quadratic form:
\begin{equation}
	V_3=V_2+\frac{1}{2}\tilde{\Delta}_{F}^{\top}\tilde{\Delta}_{F}.
	\label{V3}
\end{equation}

On the basis of Eq. (\ref{V2}), the derivative of $V_3$ can be derived as:
\begin{equation}
	{\dot{V}}_3=-z_1^{\top}\bm{k_1}z_1-z_2^{\top}\bm{k_2}z_2+\frac{z_2^{\top}\tilde{\Delta}_{F}}{m_s}+\tilde{\Delta}_{F}^{\top}\left(\dot{\Delta}_{F} - \bm{L_p}{\tilde{\Delta}}_{F}\right).
\end{equation}

Using Young's inequality, it can be achieved that
\begin{subequations}
	\begin{align} 
		\frac{z_2^{\top}\tilde{\Delta}_{F}}{m_s} &\leq \frac{1}{2}z_2^{\top}z_{2}+\frac{1}{2m_s^2}\tilde{\Delta}_{F}^{\top}\tilde{\Delta}_{F},  \\
		\tilde{\Delta}_{F}^{\top}\dot{\Delta}_{F} &\leq \frac{1}{2}\dot{\Delta}_{F}^{\top}\dot{\Delta}_{F} + \frac{1}{2}\tilde{\Delta}_{F}^{\top}\tilde{\Delta}_{F}.
	\end{align}
\end{subequations}

It can be rendered as:
\begin{equation}
	{\dot{V}}_3\leq-\beta_1z_1^{\top}z_1-\beta_2z_2^{\top}z_2-\beta_3\tilde{\Delta}_{F}^{\top}\tilde{\Delta}_{F}+\varepsilon_{1},
\end{equation}
where $\beta_1=\lambda_m(\bm{k_1})$, $\beta_2=\lambda_m(\bm{k_2})-\frac{1}{2}$, $\beta_3=\lambda_m(\bm{L_p})-\frac{1}{2}-\frac{1}{2m_s^2}$, and $\varepsilon_1$ is the upper bound of $\frac{1}{2}\dot{\Delta}_{F}^{\top}\dot{\Delta}_{F}$. $\lambda_m(\cdot)$ represents the minimum
eigenvalue of the matrix. By combining Eq. (\ref{V3}), we have
\begin{equation}
	\begin{aligned} 
		&0\leq V_3\le\frac{\varepsilon_1}{\sigma_1}+\left[V_3\left(0\right)-\frac{\varepsilon_1}{\sigma_1}\right]e^{-\sigma_1t}, \forall t>0\\
		&\sigma_1=2\min\{\beta_1,\beta_2,\beta_3\}.
	\end{aligned} 
\end{equation}

Therefore, all signals in the translational loop are uniformly ultimately bounded.

\subsubsection{Attitude generation using Hopf Fibration} Based on the cascade control principle, the differential flatness theory with Hopf Fibration is utilized to generate reference signals of the rotational loop \citep{Hopf}. Let $\psi_d$ be the desired yaw angle, which is generated by the object detection module. Then, the desired attitude quaternion $q_{d}=q_{u}\otimes q_{\psi_d}$ and angular velocity $\omega_b^{d}=\left[\omega_1,\omega_2,\omega_3\right]^{\top}\in\mathbb{R}^{3}$ can be derived as functions of $a$, $b$, $c$, and $\psi$.
\begin{equation}
	\begin{aligned} 
		&q_{u}  = \frac{1}{\sqrt{2(1+c)}}\left[\begin{matrix}1+c&-b&a&0\\\end{matrix}\right]^{\top},\\
		&q_{\psi_d}=\left[\begin{matrix}\cos(\frac{\psi_{d}}{2})&0&0&\sin(\frac{\psi_{d}}{2})\\\end{matrix}\right],\\
		&\omega_1=\sin\left(\psi\right)\dot{a}-\cos\left(\psi\right)\dot{b}-\frac{\dot{c}\left(a\sin\left(\psi\right)-b\cos\left(\psi\right)\right)}{1+c},  \\
		&\omega_2=\cos\left(\psi\right)\dot{a}+\sin\left(\psi\right)\dot{b}-\frac{\dot{c}\left(a\cos\left(\psi\right)+b\sin\left(\psi\right)\right)}{1+c}, \\
		&\omega_3=\frac{b\dot{a}-a\dot{b}}{1+c}+\dot{\psi},
	\end{aligned}  
\end{equation}
where 
\begin{equation}
	\begin{aligned} 
		\aleph &= fb_{3}, \left[\begin{matrix}a&b&c\\\end{matrix}\right]^{\top}=\frac{\aleph}{\Vert \aleph \Vert},  \\
		\left[\begin{matrix}\dot{a}&\dot{b}&\dot{c}\\\end{matrix}\right]^{\top}&=\frac{d}{dt}(\frac{\aleph}{\Vert \aleph \Vert})=\frac{\aleph^{\top}\aleph \bm{I_3} - \aleph\aleph^{\top}}{\Vert \aleph \Vert^{3}}\dot{\aleph}.
	\end{aligned}
\end{equation}

\subsubsection{Rotational loop} In the rotational loop, a NDO-based geometric controller is developed to follow references $\bm{R_d}(q_{d})$ and $\omega^{d}_{b}$. Let $e_R$ and $e_\omega$ denote the attitude tracking error and angular velocity tracking error, respectively.
\begin{equation}
	e_R =
	\frac{1}{2}
	\left(
	\bm{R_b}^{\top}\bm{R_d}
	-
	\bm{R_d}^{\top}\bm{R_b}
	\right)^\vee,
	e_\omega =
	\bm{R_b}^{\top}\bm{R_d}\omega_b^d-\omega_b,
\end{equation}
where $\left[\cdot\right]^\vee$ is the inverse of the cross-map. \citet{kevin2017} provide a rigorous proof that the tracking errors of attitude and angular velocity evolve on the tangent bundle of the Lie group $SO(3)$. Concretely, the following error dynamics can be yielded for any given $\bm{{R}_d}^{\top}\bm{R_b}\in SO(3)$ on account of Rodrigues' formula \citep{5717652}.
\begin{equation}
	\begin{aligned}
		&\dot e_\omega
		=
		-\bm{\omega_b^{\times}}
		\bm{R_b}^{\top}\bm{R_d}\omega_b^d
		+
		\bm{R_b}^{\top}\bm{R_d}\dot\omega_b^d
		-
		\dot\omega_b,
		\\
		&\dot e_R
		=
		\frac{1}{2}
		\left(
		\operatorname{tr}
		\left(\bm{R_b}^{\top}\bm{R_d}\right)\bm{I_3}
		-
		\bm{R_b}^{\top}\bm{R_d}
		\right)e_\omega
		=
		\bm{\Xi}
		\left(\bm{R_b}^{\top}\bm{R_d}\right)e_\omega,
		\\&
		\left\|
		\bm{\Xi}
		\left(\bm{R_b}^{\top}\bm{R_d}\right)
		\right\|^2
		\leq 1.
	\end{aligned}
\end{equation}

\begin{table*}[tp]
	\caption{\textbf{Tracking error statistics of different control schemes (Unit: cm).}}\label{tabtracking}
	\renewcommand{\arraystretch}{1.2}
	\resizebox{\linewidth}{!}{
		\begin{tabular}{ccccccccccccccccccc}
			\hline%
			\hline
			& \multicolumn{9}{@{}c@{}}{Test I} & \multicolumn{6}{@{}c@{}}{Test II} & \multicolumn{3}{@{}c@{}}{Test III}\\	
			\hline	
			\multirow{2}{*}{} & \multicolumn{3}{@{}c@{}}{\multirow{3}{*}{\begin{tabular}[c]{@{}c@{}} $T=12$\end{tabular}}} & \multicolumn{3}{@{}c@{}}{\multirow{3}{*}{\begin{tabular}[c]{@{}c@{}} $T=8$\end{tabular}}}
			& \multicolumn{3}{@{}c@{}}{\multirow{3}{*}{\begin{tabular}[c]{@{}c@{}} $T=6.28$\end{tabular}}}
			& \multicolumn{3}{@{}c@{}}{\multirow{3}{*}{\begin{tabular}[c]{@{}c@{}} $T=12,\omega$=3\end{tabular}}}
			& \multicolumn{3}{@{}c@{}}{\multirow{3}{*}{\begin{tabular}[c]{@{}c@{}} $T=8,\omega$=3\end{tabular}}}
			& \multicolumn{3}{@{}c@{}}{\multirow{3}{*}{\begin{tabular}[c]{@{}c@{}}$T=12,\omega$=3,\\
						\textit{wind disturbance}\end{tabular}}} \\
			\\
			\\
			\hline
			\multicolumn{1}{c|}{Methods} & $\epsilon$ & $\mu$ & \multicolumn{1}{c}{ $\varrho$} & $\epsilon$ & $\mu$ & \multicolumn{1}{c}{ $\varrho$} & $\epsilon$ & $\mu$ & \multicolumn{1}{c|}{ $\varrho$}
			& $\epsilon$ & $\mu$ & \multicolumn{1}{c}{ $\varrho$} & $\epsilon$ & $\mu$ & \multicolumn{1}{c|}{ $\varrho$} & $\epsilon$ & $\mu$ & $\varrho$ \\
			\hline
			\multicolumn{1}{c|}{Baseline I}  & {8.41} & {4.10} & 1.72 & {15.01} & {7.79} & 2.56 & {24.79} & {11.89} & \multicolumn{1}{c|}{4.37} & {11.97} & {4.91} & 2.32 & {15.58} & {8.44} & \multicolumn{1}{c|}{{2.84}} & {13.07} & \multirow{1}{*}{5.26} & {2.40} 
			\\ \hline
			\multicolumn{1}{c|}{Baseline II}  & {6.91} & {3.45} & 1.43 & {12.37} & {6.25} & 2.16 & {23.52} & {13.10} & \multicolumn{1}{c|}{{4.87}} & {10.76} & {3.92} & 1.92 & {17.18} & {7.32} & \multicolumn{1}{c|}{{3.52}} & {11.48} & {4.56} & {2.24} 
			\\ \hline
			\multicolumn{1}{c|}{Baseline III}  & {5.04} & {2.46} & 0.89 & {12.22} & {4.21} & 1.82 & {26.31} & {9.67} & \multicolumn{1}{c|}{{4.26}} & {9.70} & {3.63} & 1.78 & {13.96} & {6.58} & \multicolumn{1}{c|}{{2.78}} & {12.58} & {3.81} & {1.93} 
			\\ \hline
			\multicolumn{1}{c|}{Our scheme}  & {4.85} & {1.86} & 0.96 & {6.38} & {2.32} & 1.21 & {15.59} & {5.75} & \multicolumn{1}{c|}{{2.60}} & {5.79} & {2.43} & 0.99 & {10.66} & {5.44} & \multicolumn{1}{c|}{{1.88}} & {7.31} & {2.84} & {1.16} 
			\\ \hline 
	\end{tabular}}
\end{table*}

\begin{figure*}[!tp]
	\centering
	\includegraphics[width=\linewidth,height=\textheight,keepaspectratio]{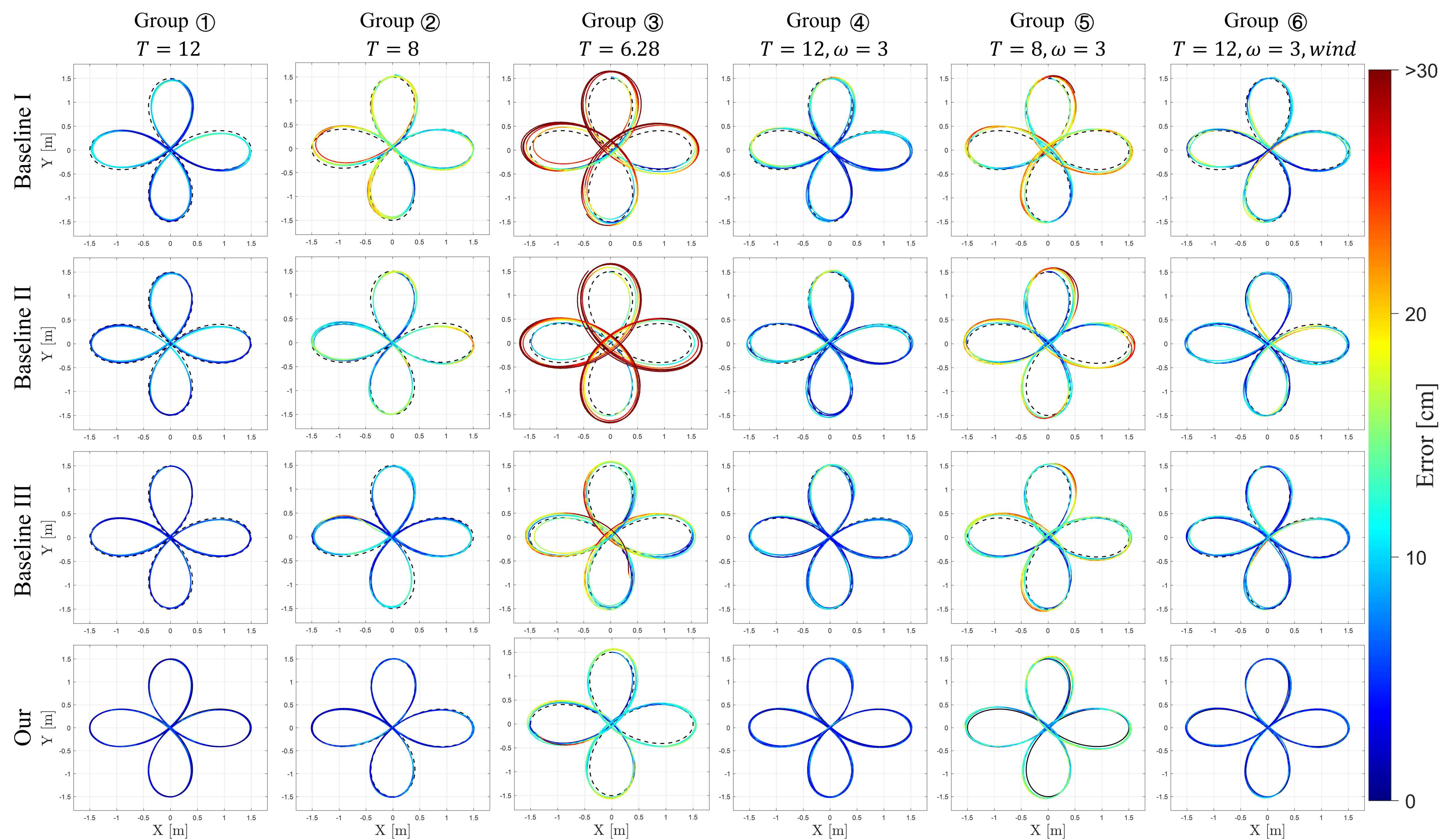}
	\caption{
		Baseline I can track the desired trajectory. However, the tracking performance substantially degrades with
		the increase of the coupling effects. The dynamical performance of Baseline II is enhanced due
		to the compensation of the first-order disturbance terms. Baseline III
		with the offline initialization training process achieves better results than Baseline I and Baseline II. By
		combining the established coupling model and residual observers, our scheme demonstrates the
		excellent dynamic performance among all experiments.}\label{IJRR_All_tracking}
\end{figure*}

By virtue of the established dynamic coupling dynamics, the rotational controller can be developed as:
\begin{equation}
	\begin{aligned}
		\tau
		={}&
		k_Re_R+k_\omega e_\omega
		+\omega_b^\times\bm J\omega_b\\
		&+\bm J
		\left(
		-\omega_b^\times
		\bm{R_b}^\top\bm{R_d}\omega_b^d
		+
		\bm{R_b}^\top\bm{R_d}\dot\omega_b^d
		\right)-\hat\tau_m-\hat\Delta_\tau .
	\end{aligned}
	\label{tau}
\end{equation}
where $k_R\in\mathbb{R}$ and $k_\omega\in\mathbb{R}$ are the designed control parameters. Aided by the auxiliary variable $\zeta$, the NDO for estimating the residual disturbance $\Delta_{\tau}$ is formulated as:
\begin{equation}
	\left\{
	\begin{aligned}
		&\hat\Delta_\tau
		=
		\zeta+\bm{L_r}\bm{J}\omega_b,
		\\
		&\dot\zeta
		=
		\bm{L_r}
		\left(
		\bm{\omega_b^{\times}}\bm{J}\omega_b
		-
		\tau-\hat\tau_m
		\right)
		-
		\bm{L_r}
		\left(
		\zeta+\bm{L_r}\bm{J}\omega_b
		\right).
	\end{aligned}
	\right.
	\label{NDO}
\end{equation}
where $\bm{L_r}\in\mathbb{R}^{3\times3}$ is the positive gain matrix. Let $\tilde{\Delta}_{\tau}=\Delta_{\tau}-\hat{\Delta}_{\tau}$ be the estimation error of the NDO. Then, we have
\begin{equation}
	\begin{aligned}
		\dot{\tilde\Delta}_\tau
		&=
		\dot\Delta_\tau-\dot{\hat\Delta}_\tau
		=
		\dot\Delta_\tau-\dot\zeta
		-
		\bm{L_r}\bm{J}\dot\omega_b
		\\
		&=
		\dot\Delta_\tau+\bm{L_r}\hat\Delta_\tau
		-
		\bm{L_r}
		\Delta_\tau
		\\
		&=
		\dot\Delta_\tau
		-
		\bm{L_r}{\tilde\Delta}_\tau.
	\end{aligned}
\end{equation}

Based on the preceding analysis, the following theorem can be established: The nonlinear controller in Eq. (\ref{tau}) with the NDO in Eq. (\ref{NDO}) can guarantee all signals in the rotational loop to be uniformly ultimately bounded when the following conditions hold:
\begin{equation}
	\begin{aligned}
		&\Theta\left(\bm{R_b},\bm{R_d}\right)
		=
		\frac{1}{2}
		\operatorname{tr}
		\left(
		\bm{I_3}-\bm{R_d}^{\top}\bm{R_b}
		\right),
		\\
		&\Theta
		\left(
		\bm{R_b}(0),\bm{R_d}(0)
		\right)
		<2,
		\\
		&\|e_\omega(0)\|^2
		<
		\frac{2k_R}{\lambda_M(\bm{J})}
		\left[
		2-
		\Theta
		\left(
		\bm{R_b}(0),\bm{R_d}(0)
		\right)
		\right].
	\end{aligned}
\end{equation}

\textit{Proof.} Consider the following auxiliary Lyapunov-function:
\begin{equation}
	V_{4}= k_{R} \Theta(\bm{R_b},\bm{R_d})+\frac{1}{2}e_{\omega}^{\top}\bm{J}e_{\omega}+\varsigma e_R^{\top}e_{\omega}+\frac{1}{2}\tilde{\Delta}_{\tau}^{\top}\tilde{\Delta}_{\tau},
\end{equation}
where $\varsigma$ is the designed non-negative constant and $\Theta(\bm{R_b},\bm{R_d})$ represents the attitude error function. For a given $\bm{R_d}$, $0\leq\Theta<2$ holds within the region where the rotation angle between $\bm{R_d}$ and $\bm{R_b}$ is less than 180$^{\circ}$. By integrating the designed dynamic controller, the derivative of $V_{4}$ can be derived as:
\begin{equation}
	\begin{aligned}
		\dot V_4
		&=
		k_R\dot\Theta(\bm{R_b},\bm{R_d})
		+
		e_\omega^\top\bm{J}\dot e_\omega
		+
		\varsigma
		\left(
		\dot e_R^\top e_\omega
		+
		e_R^\top\dot e_\omega
		\right)
		+
		\tilde\Delta_\tau^\top\dot{\tilde\Delta}_\tau
		\\
		&=
		k_Re_\omega^\top e_R
		-
		e_\omega^\top
		\left(
		k_Re_R+k_\omega e_\omega+\tilde\Delta_\tau
		\right)
		+
		\varsigma e_\omega^\top\bm{\Xi}^\top e_\omega
		\\
		&\quad
		-
		\varsigma e_R^\top\bm{J}^{-1}
		\left(
		k_Re_R+k_\omega e_\omega+\tilde\Delta_\tau
		\right)\\
		&\quad
		+
		\tilde\Delta_\tau^\top
		\left(
		\dot\Delta_\tau-\bm{L_r}\tilde\Delta_\tau
		\right).
	\end{aligned}
\end{equation}

Subsequently, $\dot{V}_{4}$ can be separated into two parts: one with terms containing $\tilde{\Delta}_{\tau}$ and the other without $\tilde{\Delta}_{\tau}$
\begin{equation}
	\begin{aligned} 
		&	\dot{V}_{4} = \dot{V}_{e} + \dot{V}_{\tau} ,\\
		&	\dot{V}_{e} =  -\varsigma k_{R}e_{R}^{\top} \textbf{\textit{J}}^{-1} e_{R}-k_{\omega}e_{\omega}^{\top}e_{\omega} - \varsigma k_{\omega}e_{R}^{\top} \textbf{\textit{J}}^{-1} e_{\omega} + \varsigma e_{\omega}^{\top}\Xi^{\top}e_{\omega}	,\\
		&\dot{V}_{\tau}=\tilde{\Delta}_{\tau}^{\top}(\dot{\Delta}_{\tau}-\textbf{\textit{L}}_\textit{r}\tilde{\Delta}_{\tau}) - (e_{\omega}^{\top} + \varsigma e_{R}^{\top} \textbf{\textit{J}}^{-1})\tilde{\Delta}_{\tau}.
	\end{aligned}  
\end{equation}

Leveraging the methodology delineated in \citet{8619390}, we have
\begin{equation}
	\dot{V}_{e} \leq -\Upsilon_{1}^{\top}\bm{M_1}\Upsilon_{1},
\end{equation}
where $\Upsilon_{1} = \left[ \Vert e_{R}\Vert, \Vert e_{\omega}\Vert \right]^{\top}$. The operator $\Vert\cdot\Vert$ represents the standard Euclidean norm for a vector. $\bm{M_1}$ is
\begin{equation}
	\bm{M_1} = \left[\begin{matrix}\frac{\varsigma k_{R}}{\lambda_{M}(\textbf{\textit{J}})}&-\frac{\varsigma k_{\omega}}{2\lambda_{m}(\textbf{\textit{J}})}		
		\\-\frac{\varsigma k_{\omega}}{2\lambda_{m}(\textbf{\textit{J}})}&k_{\omega}-\varsigma
		\\\end{matrix}\right],
\end{equation}
where $\lambda_M(\cdot)$ represents the maximum eigenvalue of the matrix. Let $V_{5}$ be 
\begin{equation}
	V_{5} = k_{R} \Theta(\bm{R_b},\bm{R_d}) +\frac{1}{2} e_{\omega}^{\top}\bm{J}{e}_{\omega}+\varsigma {e}_R^{\top}e_{\omega},
\end{equation}
such that $	V_{4} = V_{5} + \frac{1}{2}\tilde{\Delta}^{\top}_{\tau}\tilde{\Delta}_{\tau}$. It can be shown that  $V_{5}$ satisfies
\begin{equation}
	\begin{aligned} 
		\Upsilon_{1}^{\top}\underline{\bm{M_1}}\Upsilon_{1} \leq	&{V}_{5}  \leq \Upsilon_{1}^{\top}\overline{\bm{M_1}}\Upsilon_{1},\\
		\underline{\bm{M_1}} = \frac{1}{2}\left[\begin{matrix}k_{R}&-\varsigma	
			\\-\varsigma&\lambda_{m}(\textbf{\textit{J}})
			\\\end{matrix}\right], \quad &\overline{\bm{M_1}} = \frac{1}{2}\left[\begin{matrix}\frac{2k_{R}}{\rho}&\varsigma	
			\\\varsigma&\lambda_{M}(\textbf{\textit{J}})
			\\\end{matrix}\right],
	\end{aligned} \label{V5}
\end{equation}
where $0<\rho<2$. In an attempt to ensure that $\underline{\bm{M_1}}$, $\bm{M_1}$, and $\overline{\bm{M_1}}$ are all positive definite matrices, the parameter $\varsigma$ can be selected as $\varsigma < \min\{k_{\omega},\sqrt{k_R\lambda_m(\textbf{\textit{J}})},\frac{4k_{\omega} k_R\lambda_m^2(\textbf{\textit{J}})}{k_\omega^2\lambda_M(\textbf{\textit{J}})+4k_R\lambda_m^2(\textbf{\textit{J}})}\}$. Subsequently, using Young’s inequality, it can be achieved that
\begin{equation}
	\begin{aligned} 
		-e_{\omega}^{\top}\tilde{\Delta}_{\tau} &\leq e_{\omega}^{\top}e_{\omega}+\frac{1}{4}\tilde{\Delta}_{\tau}^{\top}\tilde{\Delta}_{\tau} ,\\
		-e_{R}^{\top}\bm{J}^{-1}\tilde{\Delta}_{\tau} &\leq \frac{1}{\lambda_{M}(\bm{J})}e_{R}^{\top}e_{R}+\frac{\lambda_{M}(\bm{J})}{4\lambda^{2}_{m}(\bm{J})}\tilde{\Delta}_{\tau}^{\top}\tilde{\Delta}_{\tau} 	,\\
		\tilde{\Delta}_{\tau}^{\top}\dot{\Delta}_{\tau}&\leq \frac{1}{2}\dot{\Delta}_{\tau}^{\top}\dot{\Delta}_{\tau} + \frac{1}{2}\tilde{\Delta}_{\tau}^{\top}\tilde{\Delta}_{\tau}.
	\end{aligned}  
\end{equation}

Therefore, we have
\begin{equation}
	\dot{V}_{4} \leq -\beta_{4}\tilde{\Delta}_{\tau}^{\top}\tilde{\Delta}_{\tau} - \Upsilon_{1}^{\top}\bm{M_{2}}\Upsilon_{1} + \varepsilon_2,
\end{equation}
where $\beta_4=\lambda_m(\bm{L_r})-\frac{\varsigma\lambda_{M}(\bm{J})}{4\lambda^{2}_{m}(\bm{J})}-\frac{3}{4}$. $\varepsilon_2$ is the upper bound of $\frac{1}{2}\dot{\Delta}_{\tau}^{\top}\dot{\Delta}_{\tau}$, and $\bm{M_2}$ is 
\begin{equation}
	\bm{M_2} = \left[\begin{matrix}\varsigma\frac{ k_{R}-1}{\lambda_{M}(\textbf{\textit{J}})}&-\frac{\varsigma k_{\omega}}{2\lambda_{m}(\textbf{\textit{J}})}		
		\\-\frac{\varsigma k_{\omega}}{2\lambda_{m}(\textbf{\textit{J}})}&k_{\omega}-\varsigma-1
		\\\end{matrix}\right].
\end{equation}

Furthermore, in order to guarantee that $\bm{M_2}$ is a positive definite matrix, the parameter $\varsigma$ can be selected as $\varsigma < \min\{k_{\omega}-1,\sqrt{k_R\lambda_m(\textbf{\textit{J}})},\frac{4(k_{\omega}-1) (k_R-1)\lambda_m^2(\textbf{\textit{J}})}{k_\omega^2\lambda_M(\textbf{\textit{J}})+4(k_R-1)\lambda_m^2(\textbf{\textit{J}})}\}$. By combining Eq. (\ref{V5}), we have 
\begin{equation}
	\dot{V}_4 \leq -\sigma_2V_4+\varepsilon_2,
\end{equation}
where $\sigma_2=\min\{\frac{\lambda_m(\bm{M_2})}{\lambda_M(\overline{\bm{M_1}})},2\beta_4\}$. Therefore, it can be given 
\begin{equation}
	0 \leq V_4\leq \frac{\varepsilon_2}{\sigma_2}+\left[V_4(0)-\frac{\varepsilon_2}{\sigma_2}\right]e^{-\sigma_2t}, \forall t>0
\end{equation}
which indicates that $e_{R}$, $e_{\omega}$, and $\tilde{\Delta}_{\tau}$ are uniformly ultimately bounded. 

\begin{Remark}
	As shown in Fig.~\ref{IJRR_setup}, the coaxial quadrotor is selected as the aerial platform to provide high thrust density without increasing the platform size. However, aerodynamic interaction between the upper and lower rotors leads to thrust loss, making conventional inverse-model allocation inaccurate. To account for these effects, a physics-informed optimization-based control allocation method for coaxial multirotors is adopted according to Eq.~\eqref{f_total} and Eq. \eqref{tau} \citep{11618145}. A nonlinear mapping from rotor inputs to the generated thrust and torque is experimentally identified and embedded into the allocation process to explicitly account for coaxial aerodynamic interaction.
\end{Remark}

\subsection{Manipulator kinematic controller}
\begin{figure}[tbp]
	\centering
	\includegraphics[width=\linewidth,height=\textheight,keepaspectratio]{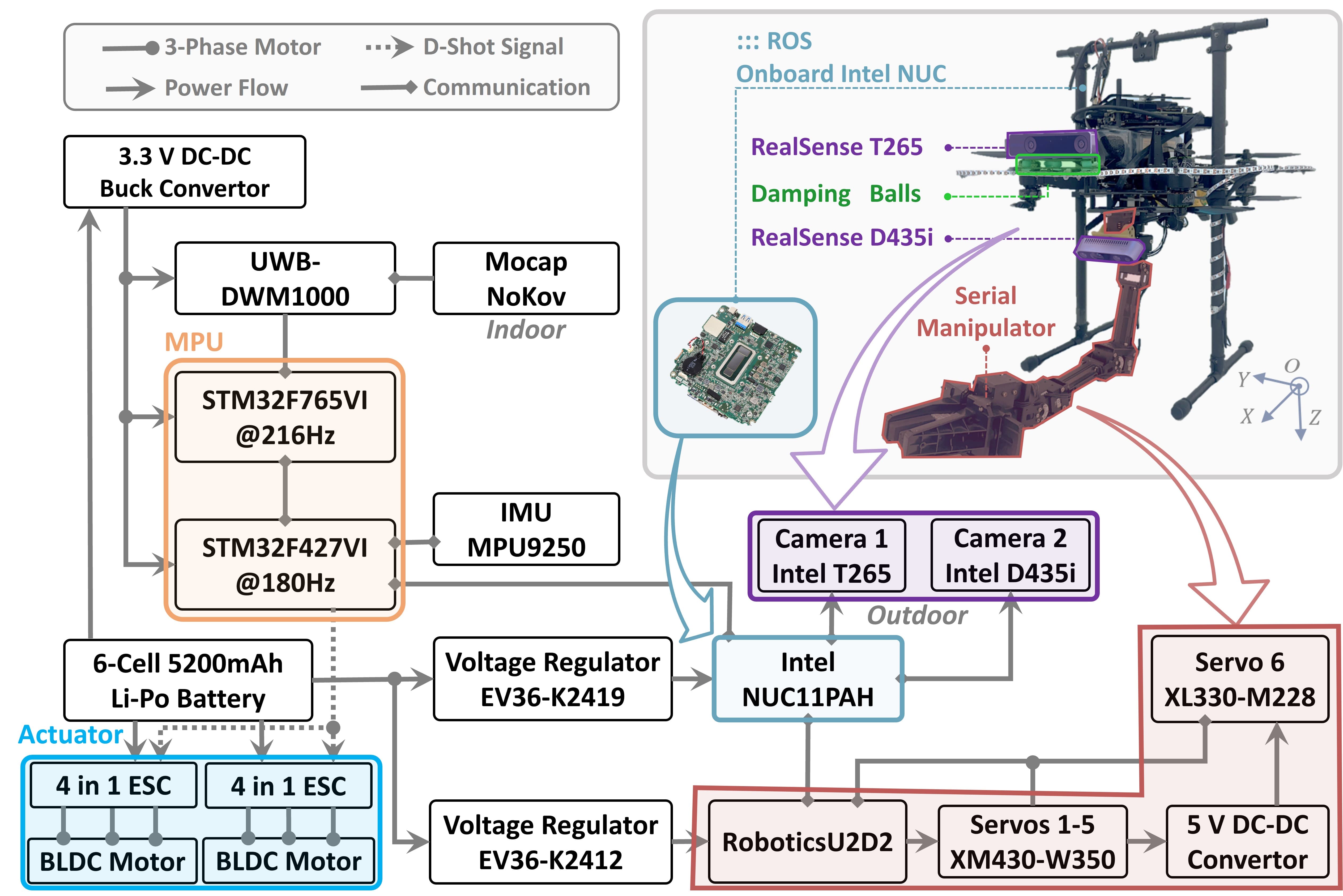}
	\caption{Hardware system of CAPTURER.}\label{IJRR_setup}
\end{figure}

In our work, the manipulator control commands are defined at the kinematic level. The inner control loop then generates the necessary torque to drive the robot, guided by the reference joint position or velocity from the outer kinematic loop. As illustrated in Fig. \ref{IJRR_setup}, a joint velocity planner is presented. 

Let $p_{ed}^{d}\in \mathbb{R}^{3}$ represent the desired state of the end-effector in $\{\mathcal{D}\}$. $\Delta p_{ed} = p_{ed}^{d} - p_{ed}$ denotes the tracking error of the end-effector. Based on the feedback signals, the desired joint velocity $\dot{q}_{re}\in \mathbb{R}^{4}$ is defined as:
\begin{equation}
	\begin{aligned}
		&\dot{q}_{re} = \bm{J_{t}}^{\dagger}\dot{p}_{ed}^{d} + \bm{J_{t}}^{\top}\Delta \varepsilon,
		\Delta \varepsilon = \frac{\partial \Pi\left(\Delta p_{ed}\right)}{\partial \Delta p_{ed}},\\
		&\Pi\left(\Delta p_{ed}\right) = \frac{k_{d}}{N_{m}}\left[\max\left(0,||\Delta  p_{ed}||^{2}-\rho\right)\right]^{N_{m}},
	\end{aligned}	 
\end{equation}
where $\bm{J_{t}}^{\dagger}$ is the pseudo-inverse of $\bm{J_{t}}$.  $\Pi\left(\Delta p_{ed}\right)\in\mathbb{R}$ represents the designed potential energy function for tracking error, which incorporates an energy gain parameter $k_{d}>0$, a positive bound $\rho>0$, and a constant $N_{m}\in\mathbb{N}, N_{m}\geq 2$. $\Delta \varepsilon\in\mathbb{R}^{3}$ denotes the gradient of $\Pi(\Delta p_{ed})$. 

\subsection{Controller comparison}\label{CC}
Three scenarios are designed to quantitatively evaluate the tracking performance of the developed anti-disturbance control scheme. In addition, three different controllers are selected for comparison: Baseline I (nonlinear baseline controller) vs. Baseline II (observer-based controller \citep{10083097}) vs. Baseline III (learning-based controller \citep{CAO2022107607,10547187,MA2022212}) vs. Our method. Concretely, our scheme without the disturbance rejection strategy is defined as the nonlinear baseline controller. In \citet{10083097}, two ESOs are employed to estimate and compensate the ignored higher-order coupling terms. In Baseline III, the dynamic coupling disturbances are approximated by the NN using available state information \citep{CAO2022107607,10547187,MA2022212}. 

The given reference of the aerial platform is a four-leaf trajectory, which is defined as:
$$[1.5\cos(\frac{4\pi}{T}t)\sin(\frac{2\pi}{T}t), 1.5\cos(\frac{4\pi}{T}t)\cos(\frac{2\pi}{T}t), -1] \quad m $$  with the lap time $T$. Each experiment consists of one warm-up lap followed by three recorded laps. For $e_i=\|p_b^d(i)-p_b(i)\|$, the maximum tracking
error $\epsilon$, mean absolute tracking error $\mu$, and standard deviation $\varrho$ are adopted as the performance metrics:
\begin{equation}
	\epsilon=\max_i e_i, \mu=\frac1n\sum_i e_i,
	\varrho=\sqrt{\frac1{n-1}\sum_i(e_i-\mu)^2}.
\end{equation}

\textit{\textbf{Test I:}} Test I evaluates the tracking performance at different flight speeds. The manipulator is kept fixed relative to the aerial platform, while three experiments are conducted with different lap times. As shown in Fig.~\ref{IJRR_All_tracking}, the proposed anti-disturbance control scheme maintains satisfactory tracking performance under all tested conditions by fully utilizing all available information. Taking $T=8$ s as an example, the proposed scheme reduces $\mu$ by 70\%, 63\%, and 45\% compared with Baselines I, II, and III, respectively, from 7.79~cm, 6.25~cm, and 4.21~cm to 2.32~cm. Similarly, $\varrho$ is reduced by 53\%, 44\%, and 34\%, from 2.56~cm, 2.16~cm, and 1.82~cm to 1.21~cm. Regarding $\epsilon$, the error decreases from 15.01~cm, 12.37~cm, and 12.22~cm under Baselines I, II, and III to 6.38~cm with the proposed scheme, corresponding to improvements of 58\%, 48\%, and 48\%, respectively.

\textit{\textbf{Test II:}} Test II explores the capability of achieving dynamic operations. The manipulator tracks the periodic trajectory defined in Section~\ref{coupling_analysis} with \(\omega=3\) rad/s, while the aerial platform follows the same four-leaf trajectory used in Test I. The prime purpose is to reproduce the strong coupling effects caused by dynamic operations. As reported in Table \ref{tabtracking}, due to the involvement of the dynamic coupling, the $\epsilon$ metric using compared methods deteriorates by 42\%, 56\%, and 93\%, respectively (from 8.41 cm to 11.97 cm, from 6.91 cm to 10.76 cm, and from 5.04 cm to 9.70 cm). In sharp contrast, the control performance of our method is decreased by approximately 19\%. Moreover, the $\mu$ metric of our proposed scheme improves by 51\%, 38\%, and 33\% with the assistance of the estimated coupling disturbances in comparison with three baselines (from 4.91 cm, 3.92 cm, and 3.63 cm to 2.43 cm). The improvement rates of $\varrho$ and $\epsilon$ can be found in Table \ref{tabtracking}.
\textit{\textbf{Test III:}} More importantly, Test III evaluates the robustness of the proposed anti-disturbance scheme against unknown external disturbances. A 380 W fan is positioned at [-2,-2,-1] m to generate the disturbance field. The maximum wind speed reaches 8~m/s, as measured using an AS8556 digital anemometer. As shown in Table \ref{tabtracking}, the tracking performance of Baseline I declines by 7\% due to insufficient resistance to external disturbances (mean absolute error $\mu$ from 4.91 cm to 5.26 cm). Moreover, the peak error $\epsilon$ of Baseline III increases by 30\% (from 9.70 cm to 12.58 cm) due to the fact that the learning-based adaptive controller requires several periods to stabilize when addressing unknown external disturbances. Moreover, compared with Baseline II, the proposed scheme reduces $\epsilon$, $\mu$, and $\varrho$ by 36\%, 38\%, and 48\%, respectively (from 11.48 cm, 4.56 cm, and 2.24 cm to 7.31 cm, 2.84 cm, and 1.16 cm). The tracking performance under various scenarios is demonstrated in Fig. \ref{IJRR_All_tracking}. More experimental details can be found in the Appendix.

%% file: sections/07_grasping.tex
\section{Aerial grasping across different environments}\label{AGADE}
\subsection{Hardware}
Fig. \ref{IJRR_setup} shows the physical architecture and electronic interconnections of CAPTURER. CAPTURER is built on a customized X300-C8 carbon-fiber frame, powered by a 6S 5200 mAh Li-Po battery. Four dedicated voltage regulators provide independent power supplies for the onboard modules. High-level computation is centralized on an Intel NUC11PAH (16GB RAM), which executes cooperative planning and perception algorithms within the Robot Operating System (ROS Noetic) framework. Visual perception is facilitated by a dual-camera configuration: a RealSense T265 provides platform state feedback, while a RealSense D435i handles target detection. To maintain target visibility during aerial grasping, the D435i pitch angle is dynamically regulated by a dedicated XL330-M228 servo actuator.

\begin{figure*}[!tbp]
	\centering
	\includegraphics[width=0.95\textwidth]{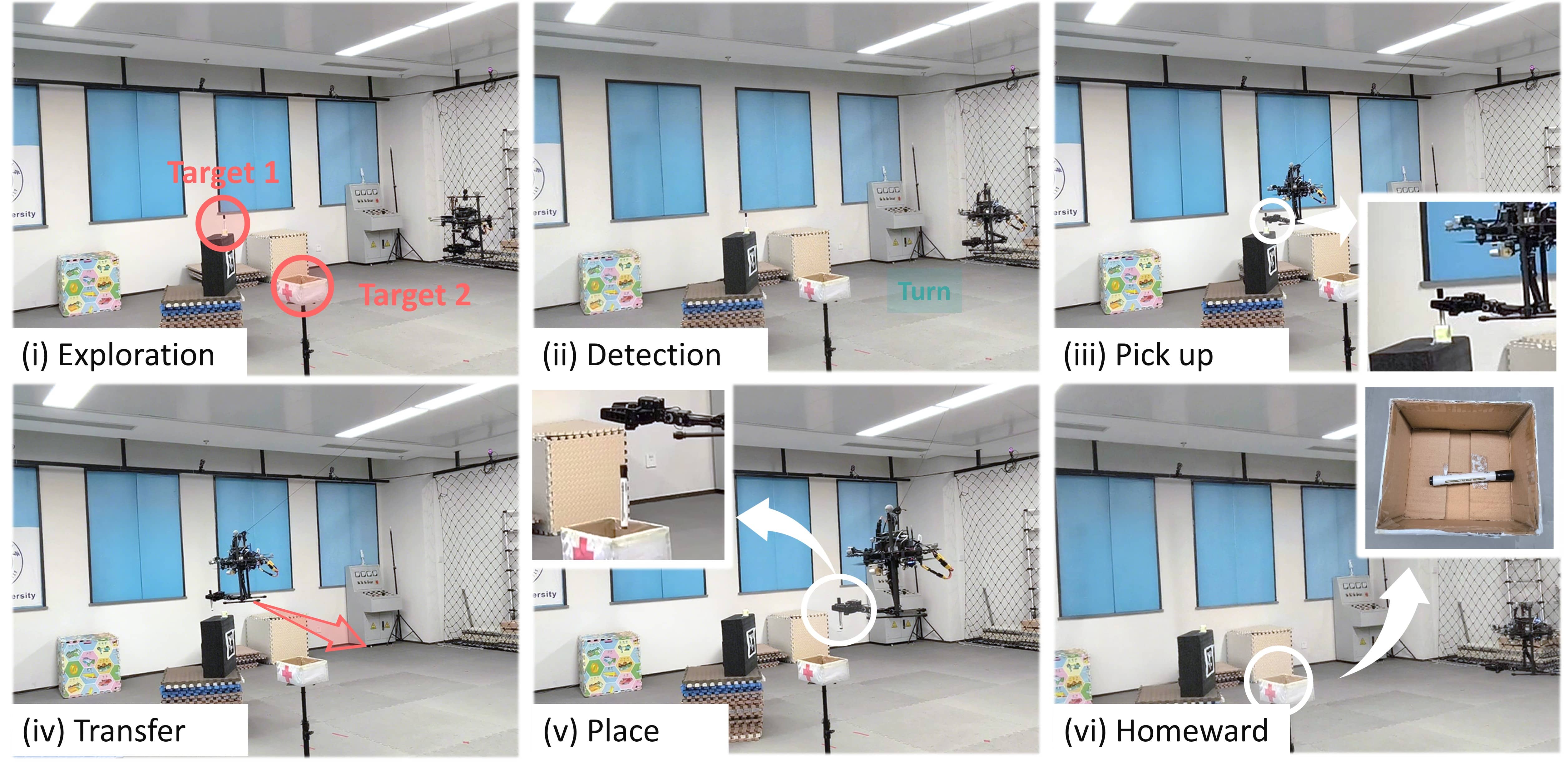}
	\captionsetup{font={stretch=1.0},justification=justified,singlelinecheck=false}
	\caption{ CAPTURER actively detects and grasps a target, then precisely places it into the specified area. The experimental video can be found in the Supplementary Movie S2.}
	\label{IJRR_grasp_V1}
\end{figure*}
\begin{figure*}[!tp]
	\centering
	\includegraphics[width=0.95\textwidth]{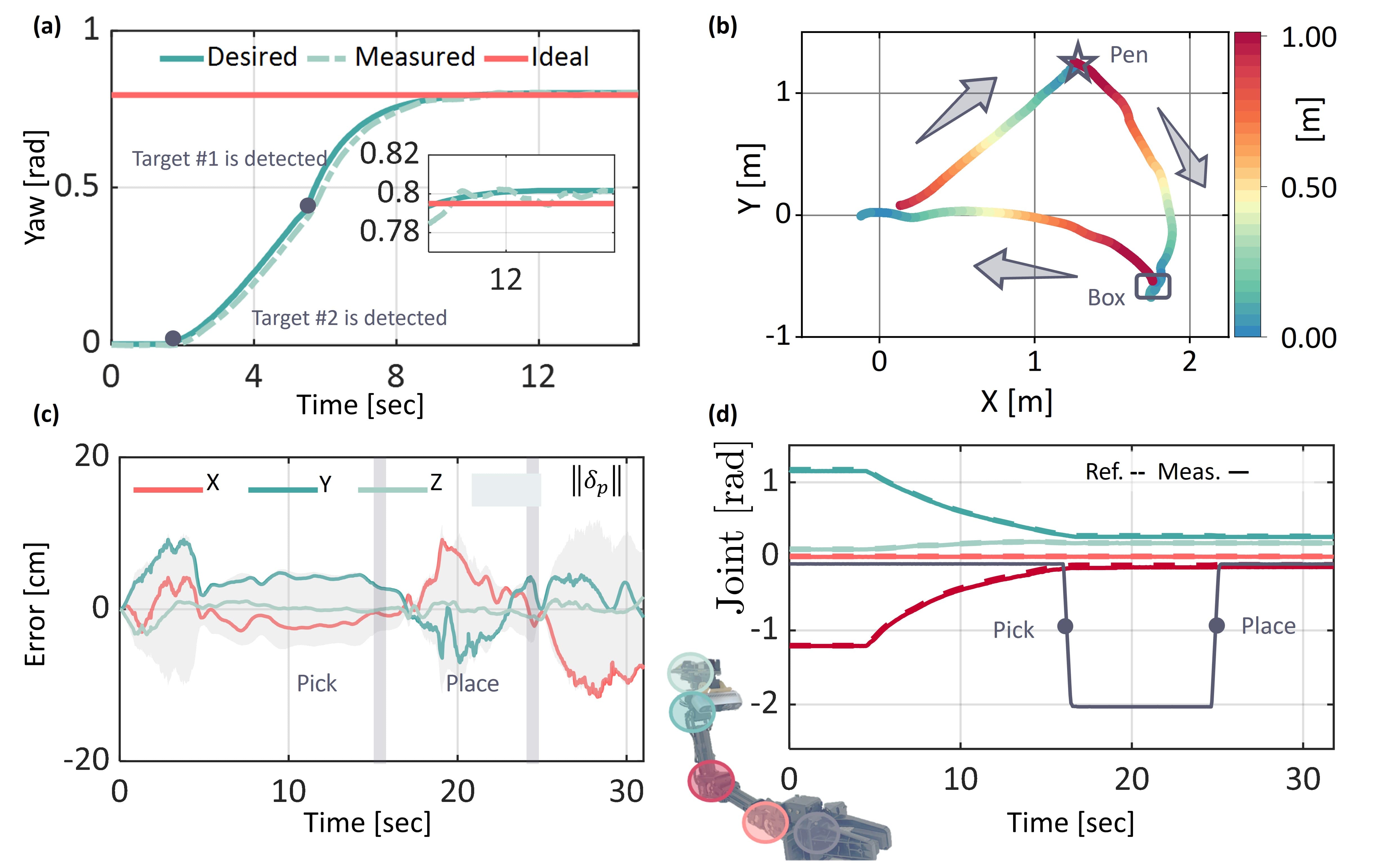}
	\captionsetup{font={stretch=1.0},justification=justified,singlelinecheck=false}
	\caption{ (a) Process of detecting unknown objects. (b) Process of aerial grasp and place. (c)  Position tracking error of the aerial platform. (d) The manipulator joints are regulated to grasp the target.}
	\label{IJRR_grasp_V1_5}
\end{figure*}
The serial manipulator comprises five XM430-W350 actuators, interfaced via a U2D2 converter that translates high-level UART commands into 4 Mbps TTL signals.  An STM32F427VI MCU (180 MHz) interfaces with an MPU9050 IMU (500 Hz) and a UWB-D100 sensor (50 Hz), utilizing the MAVLink protocol for synchronized communication with the host computer. The dynamic flight controller and motor allocation scheme are implemented on a separate STM32F765VI MCU, operating with multi-rate loops for position (100 Hz) and attitude (500 Hz) regulation. 

\subsection{Experimental details}
Our experiments involve grasping objects under different operating conditions. The effective grasping ranges of the gripper along the \(X\)- and \(Y\)-axes are 4~cm and 5~cm, respectively. In contrast, the allowable error along the \(Z\)-axis depends on the target dimension. For the marker pen, modeled as a cylinder with a diameter of 1.5~cm and a height of 10~cm, the corresponding maximum geometric tolerances are 2.5~cm, 3.5~cm, and 10~cm along the \(X\), \(Y\), and \(Z\) directions, respectively. The 10~cm value is specific to the marker pen. For smaller objects, such as the servo actuator evaluated later, the allowable errors decrease according to the object dimensions. Therefore, in the marker-pen experiments, grasping success is most sensitive to the \(X\)-axis tracking error. 

The gripper is commanded to close when the position error between the end-effector and the target falls within the prescribed region.
After receiving the command, the gripper takes approximately 0.2~s to close from its fully open configuration. Compared with quasi-static
grasping, dynamic grasping imposes a tighter timing constraint. The end-effector must achieve the required grasping alignment before closure
is completed, despite ongoing relative motion. The limited time available for adjustment makes close coordination between the aerial platform, manipulator, and gripper essential.

Unlike methods that increase the allowable grasping tolerance through soft grippers, compliant structures, or magnetic end effectors \citep{8594330,9196576,10224341,npjrobotics}, CAPTURER relies on the local adjustment of a rigid serial manipulator to directly grasp centimeter-scale objects. The framework preserves high positioning precision without relying on a task-specific attachment mechanism, making it suited to dynamic aerial manipulation.
\subsection{Indoor grasping of a stationary object}\label{Indoor}
An object detection and pick-and-place task is conducted to evaluate the aerial manipulation capability of CAPTURER. CAPTURER is required to actively detect a target, grasp it, and subsequently place it in a specified area. This experiment evaluates the complete perception, planning, and control pipeline during the representative physical interaction task. The key snapshots are illustrated in Fig. \ref{IJRR_grasp_V1}. Two capabilities are evaluated: 1) the ability to detect multiple objects is demonstrated via yaw modulation; 2) the ability to generate the cooperative trajectory and conduct aerial operations is highlighted during the object pick-and-place process. 

\begin{figure*}[!tp]
	\centering
	\includegraphics[width=\textwidth,height=0.8\textheight,keepaspectratio]{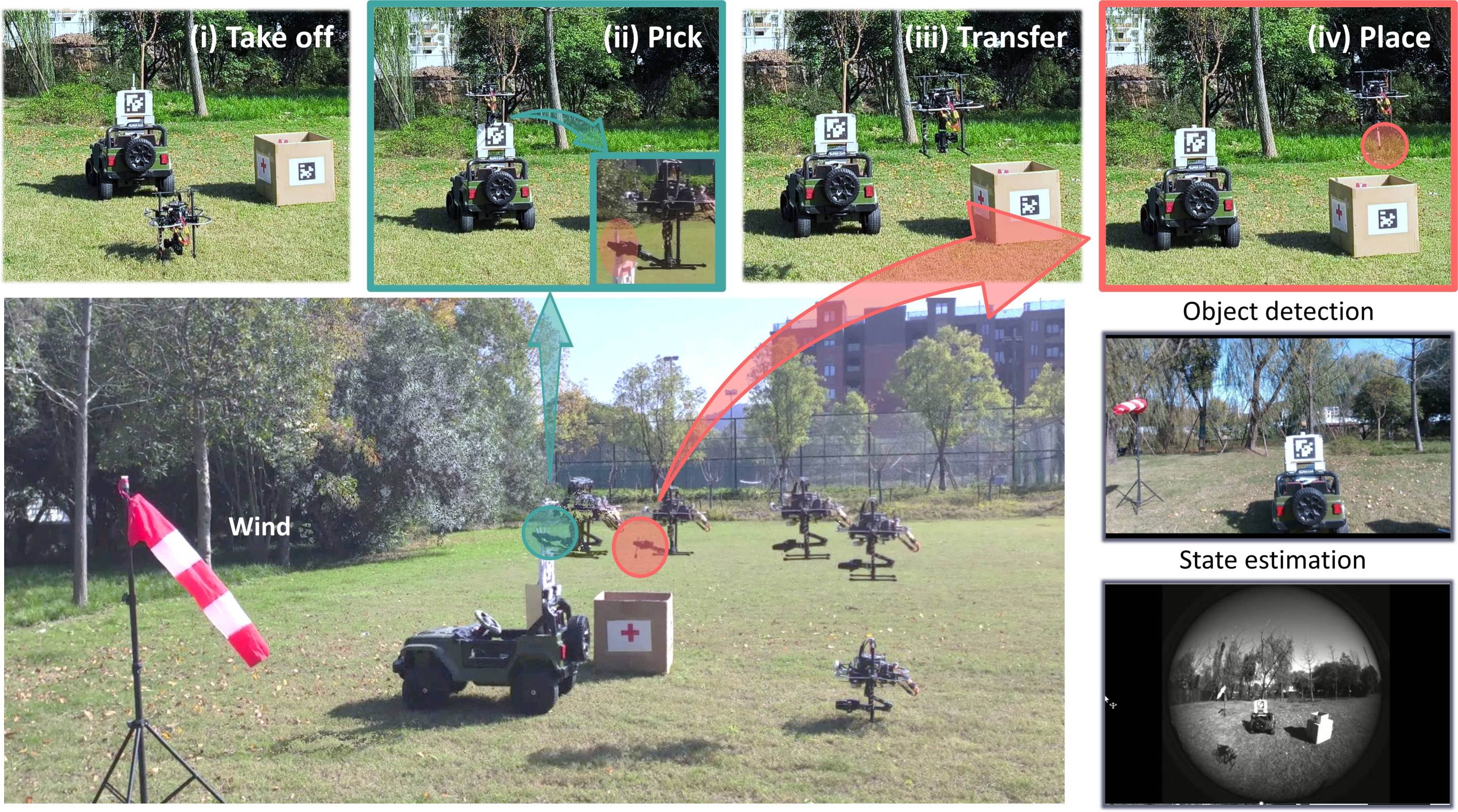}
	\captionsetup{font={stretch=1.0}}
	\caption{Sequential snapshots of the aerial grasping in an open lawn. The experimental video can be found in the Supplementary Movie S3.}
	\label{IJRR_grasp_V2}
\end{figure*}
\begin{figure*}[!tp]
	\centering
	\includegraphics[width=0.9\textwidth,height=\textheight,keepaspectratio]{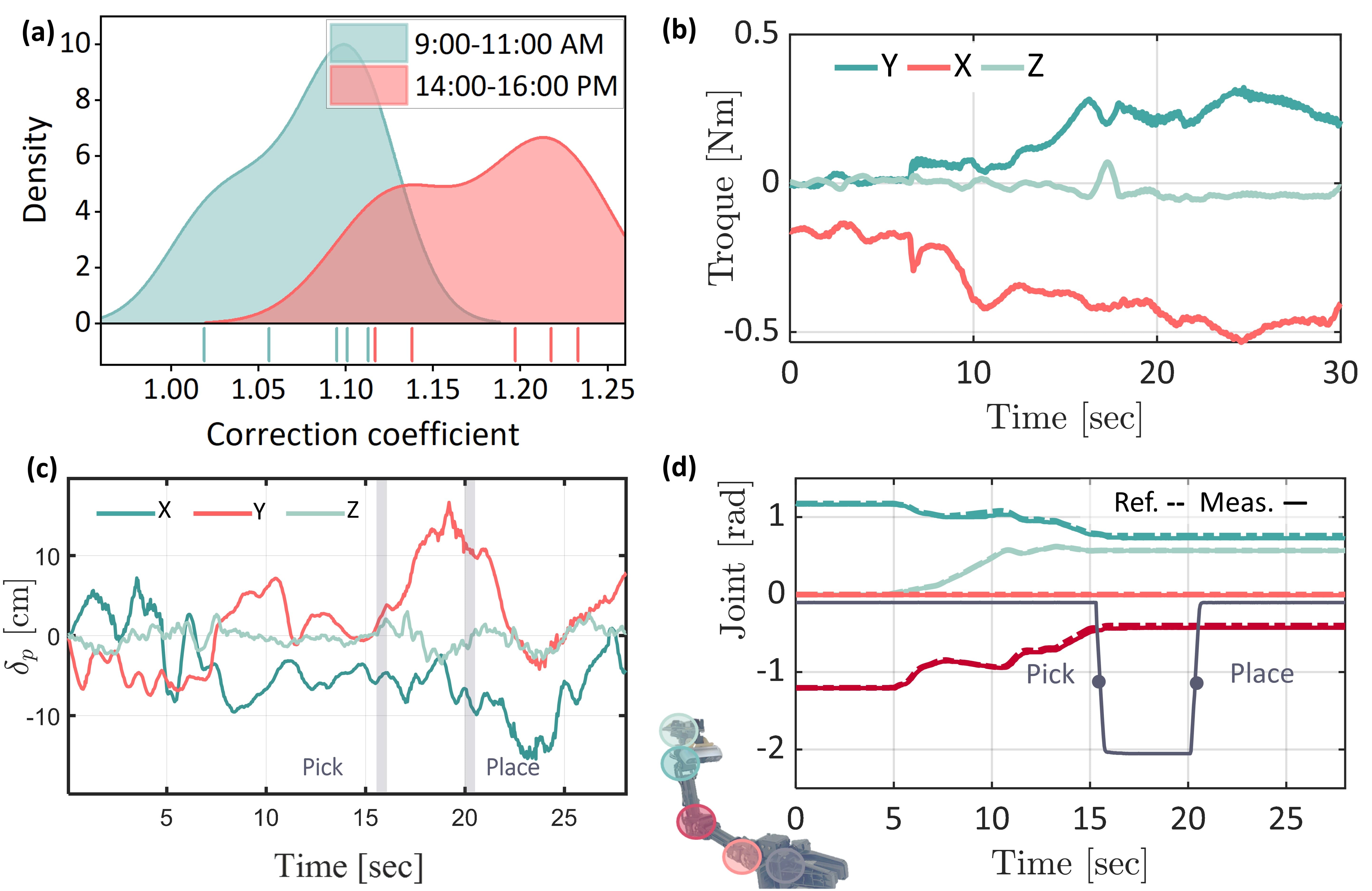}
	\captionsetup{font={stretch=1.0}}
	\caption{(a) The distribution of state drift across different times. (b) Estimation results of residual disturbance torque. (c) Position tracking error of the aerial platform along three axes. (d) The manipulator state is continuously regulated to compensate for the UAV tracking error.}
	\label{IJRR_grasp_V2_5}
\end{figure*}
To evaluate the onboard target detection capability, the marker pen and medical box are detected solely using the onboard camera. Fig. \ref{IJRR_grasp_V1}(a) illustrates that only target 2 appears in the camera's FoV during the initial phase after takeoff. Therefore, CAPTURER actively rotates clockwise to search for target 1 after acquiring the filtered position of target 2. Once target 1 appears in the camera's FoV, the desired yaw angle is immediately computed according to the detected target position, allowing the aerial manipulator to continuously orient toward target 1. A first-order filter is employed to generate the desired yaw angle to avoid abrupt attitude changes. Fig. \ref{IJRR_grasp_V1_5}(a) illustrates the entire process of detecting two targets, ending with CAPTURER aligned with target 1 for grasping. More details of the presented detection strategy can be found in Section \ref{m_ODMP}.

In comparison with existing studies where the trajectory is  pre-defined offline \citep{9834073,7989751,9196576,10495172,9981668,10339889}, CAPTURER generates cooperative motion online according to the relative target state. When the target lies outside the effective workspace of the manipulator, the aerial platform first approaches the marker pen. Once the distance between the end-effector and the target decreases below the switching threshold $d_0=0.20$~m, the manipulator joint configuration is optimized online to achieve the precise grasp. After placing the object in the specified area, CAPTURER returns to its initial position. During the entire task, the motion references of the UAV and the manipulator are updated online. As can be seen from Fig. \ref{IJRR_grasp_V1_5}(b), the motion characteristics of the aerial manipulator are incorporated. The key flight data of the aerial platform are displayed in Fig. \ref{IJRR_grasp_V1_5}(c). Fig. \ref{IJRR_grasp_V1_5}(d) illustrates the optimized joint state of the manipulator and its dynamic control performance. The tracking error of CAPTURER is illustrated in Fig. \ref{IJRR_endeffector_1}.

\subsection{Grasping a stationary object in an open lawn}\label{lawn}
CAPTURER is further evaluated in an outdoor open lawn using onboard perception. As illustrated in Fig.~\ref{IJRR_grasp_V2}, CAPTURER successfully grasps a static target in an open lawn under natural wind conditions. Compared with the structured indoor environment, outdoor aerial grasping
introduces additional visual uncertainties, including illumination
variations and cluttered backgrounds. These factors induce inevitable deviations in vision-based state estimation. To maintain manipulation success, a drift correction approach is implemented to reduce  these perception errors (Section \ref{m_S_dri}). Fig. \ref{IJRR_grasp_V2_5}(a) indicates that CAPTURER can compensate for varying drift magnitudes across different environmental trials.

The robustness of CAPTURER is further evaluated under aerodynamic disturbances. Natural wind causes noticeable motion fluctuations of the aerial platform, as reflected by the estimated residual torques in Fig.~\ref{IJRR_grasp_V2_5}(b). Meanwhile, the aerial platform position tracking errors are shown in Fig.~\ref{IJRR_grasp_V2_5}(c). Despite these disturbances, the manipulator continuously adjusts its configuration according to the cooperative planner to compensate for the platform motion. This regulation continues until the end-effector enters the predefined grasping tolerance region. Fig.~\ref{IJRR_grasp_V2_5}(d) shows the manipulator regulation process. 

The outdoor manipulation would be difficult for conventional aerial robots lacking active motion regulation. By jointly optimizing the platform and manipulator states, CAPTURER successfully grasps the object when the target enters its dexterous workspace. The online adaptation demonstrates the applicability of the proposed framework to both indoor and outdoor manipulation.

\begin{figure*}[!tp]
	\centering
	\includegraphics[width=\textwidth,height=0.8\textheight,keepaspectratio]{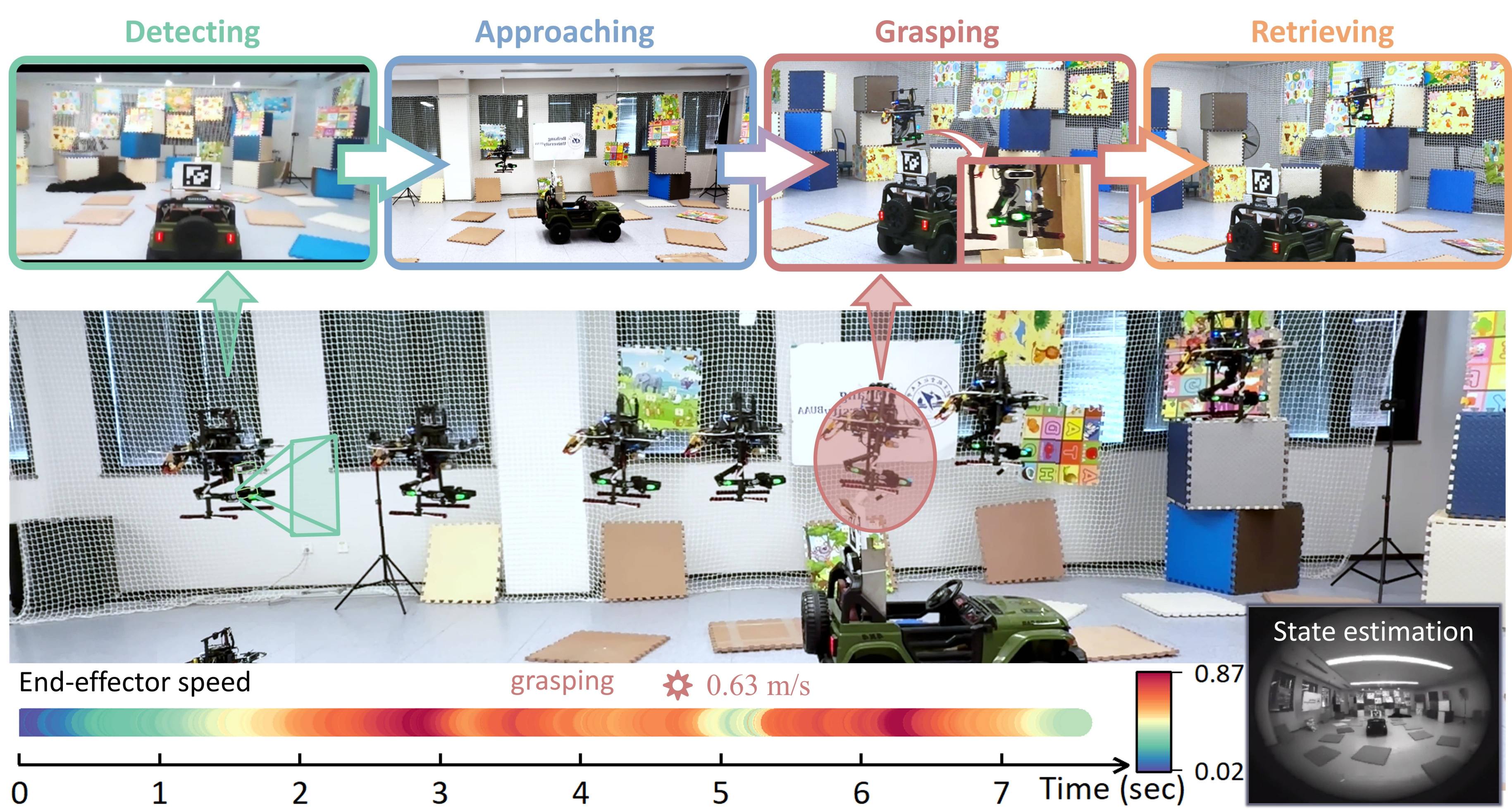}
	\captionsetup{font={stretch=1.0}}
	\caption{Key snapshots of the dynamic aerial grasping. The experimental video can be found in the Supplementary Movie S4.}
	\label{IJRR_grasp_V3}
\end{figure*}

\begin{figure*}[!tp]
	\centering
	\includegraphics[width=0.9\textwidth,height=\textheight,keepaspectratio]{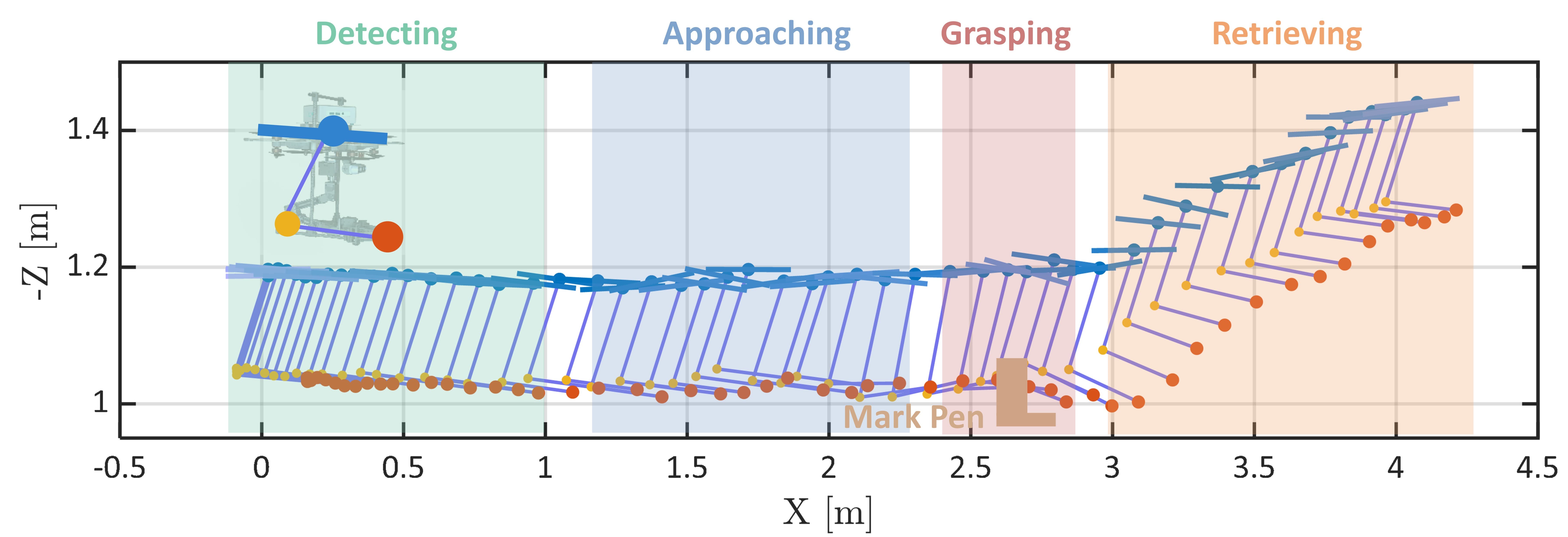}
	\captionsetup{font={stretch=1.0}}
	\caption{Poses of CAPTURER in the $X$-$Z$ plane, where the brown box
		indicates the captured object.}
	\label{IJRR_grasp_V3_1}
\end{figure*}

\subsection{Dynamic grasping of a stationary object} \label{DGSO}
We evaluate the dynamic grasping capability of CAPTURER by significantly compressing the manipulation window. Unlike conventional quasi-static aerial manipulation \citep{pioneer,5980314,7989751,9789094,10486851,Ryll20196DIC}, CAPTURER performs target grasping at a high relative velocity, an essential capability  for time-critical missions.
Nevertheless, higher grasping speeds place greater demands on perception, motion coordination, and tracking accuracy.  In comparison with existing systems that require predefined initial configurations \citep{10224341}, the presented online replanning framework allows the grasping task to be initiated from different initial positions. In this experiment, the manipulation is initiated at a distance of approximately 3.0~m, close to the effective range of the onboard target detection system. More details can be found in Fig.~\ref{IJRR_Detection}.

Another  challenge in dynamic grasping arises from the non-negligible actuation latency during gripper configuration transitions. At high relative velocity, the configuration delay can induce additional  relative displacement between the end-effector and target. CAPTURER addresses this issue by coordinating the platform motion, manipulator joint, and gripper timing. The entire grasping sequence is illustrated in Fig. \ref{IJRR_grasp_V3}. During the initial approach, the manipulator remains retracted to reduce the CoM shift and the resulting dynamic coupling effects. The switching distance $d_0$ is adjusted according to the aerial platform velocity $|v_b|$. Once the target enters this range, the aerial platform maintains its forward motion, while the manipulator is actively regulated to compensate for the remaining relative position error and complete the grasp.

Upon successful grasping, CAPTURER moves away from the target region while maintaining the grasped object. Unlike previous methods that require active deceleration \citep{10224341} or manual trajectory tuning \citep{npjrobotics}, CAPTURER successfully captures the target at a relative velocity of 0.63~m/s using onboard perception. Fig. \ref{IJRR_grasp_V3_1} illustrates the motion process. The end-effector tracking error is shown in Fig. \ref{IJRR_endeffector_3}.

\subsection{Dynamic grasping of a moving object}\label{outdoor_moving}

\begin{figure*}[!tp]
	\centering
	\includegraphics[width=\textwidth,height=0.8\textheight,keepaspectratio]{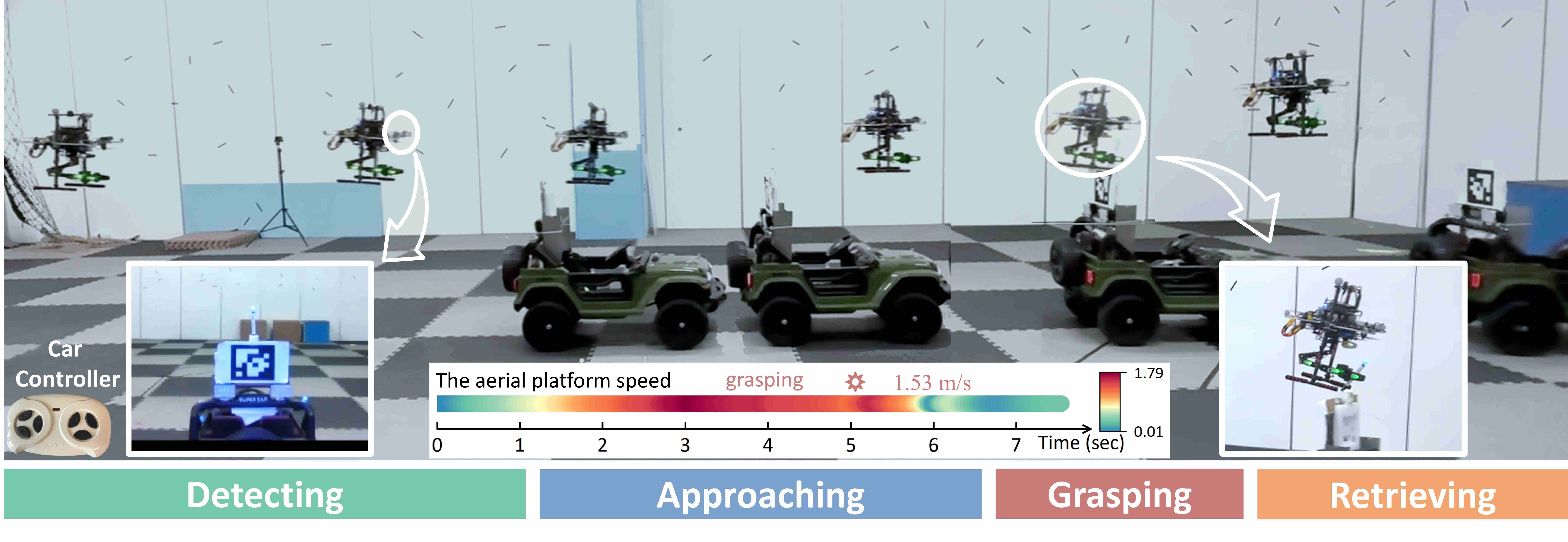}
	\captionsetup{font={stretch=1.0}}
	\caption{Key snapshots of the dynamic grasping of a moving object. The experimental video can be found in the Supplementary Movie S5.}
	\label{IJRR_grasp_V4}
\end{figure*}

To evaluate CAPTURER under a more challenging dynamic scenario, we require the aerial system to grasp a moving target within a spatially constrained environment, as shown in Fig. \ref{IJRR_grasp_V4}. A marker pen mounted on a mobile vehicle is driven toward the workspace boundary. This setting creates a limited capture window and requires the aerial manipulator to complete the grasp before the target leaves the reachable region. In aerial robotics, dexterous grasping of moving targets removes the need to immobilize the target before interaction and extends aerial manipulation toward dynamic tasks. However, maintaining centimeter-level precision during dynamic interaction with a moving target introduces substantial challenges. Due to motion blur and sensor synchronization latency, visual measurements alone are insufficient for maintaining continuous state information during rapid relative motion. To maintain perception continuity, target motion is extrapolated via dynamic state propagation, with step sizes adaptively coupled to the relative system states. The predicted target state provides a continuous reference for online cooperative planning when current measurements are delayed or temporarily degraded.

\begin{figure*}[!tp]
	\centering
	\includegraphics[width=0.9\linewidth,height=\textheight,keepaspectratio]{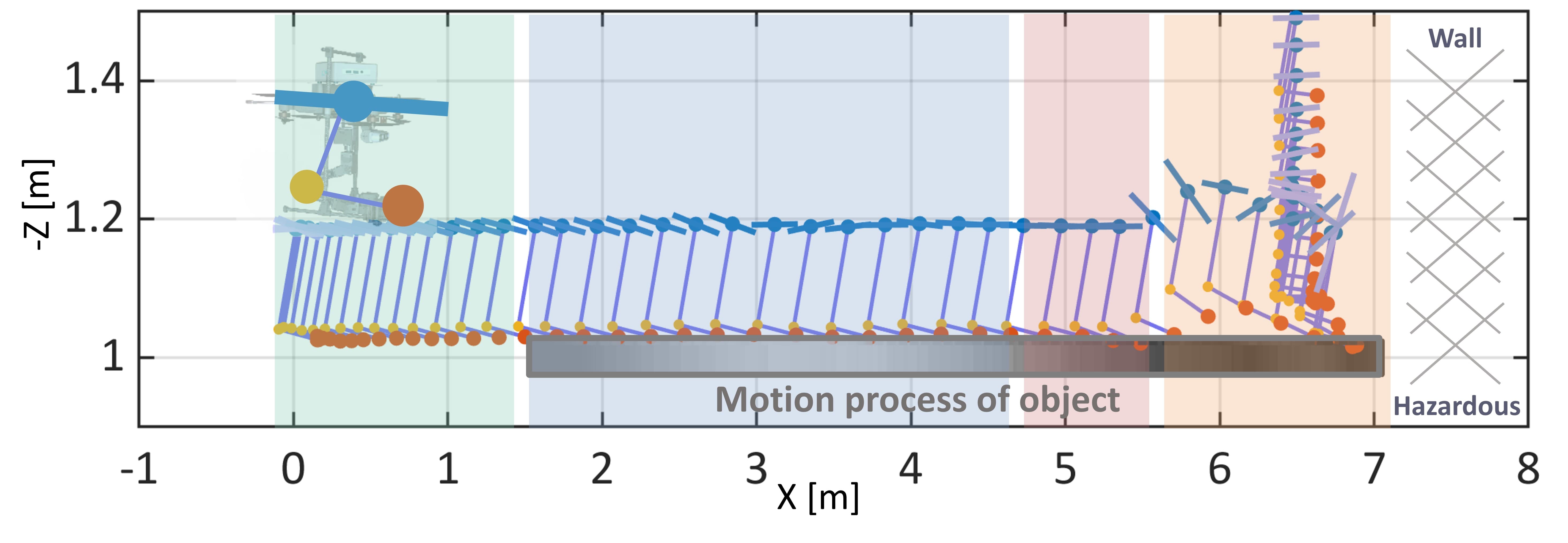}
	\caption{Poses of CAPTURER in the $X$-$Z$ plane, where the gray box indicates the target motion process.}
	\label{IJRR_grasp_V4_1}
\end{figure*}
\begin{figure}[!tbp]
	\centering
	\includegraphics[width=\linewidth,height=\textheight,keepaspectratio]{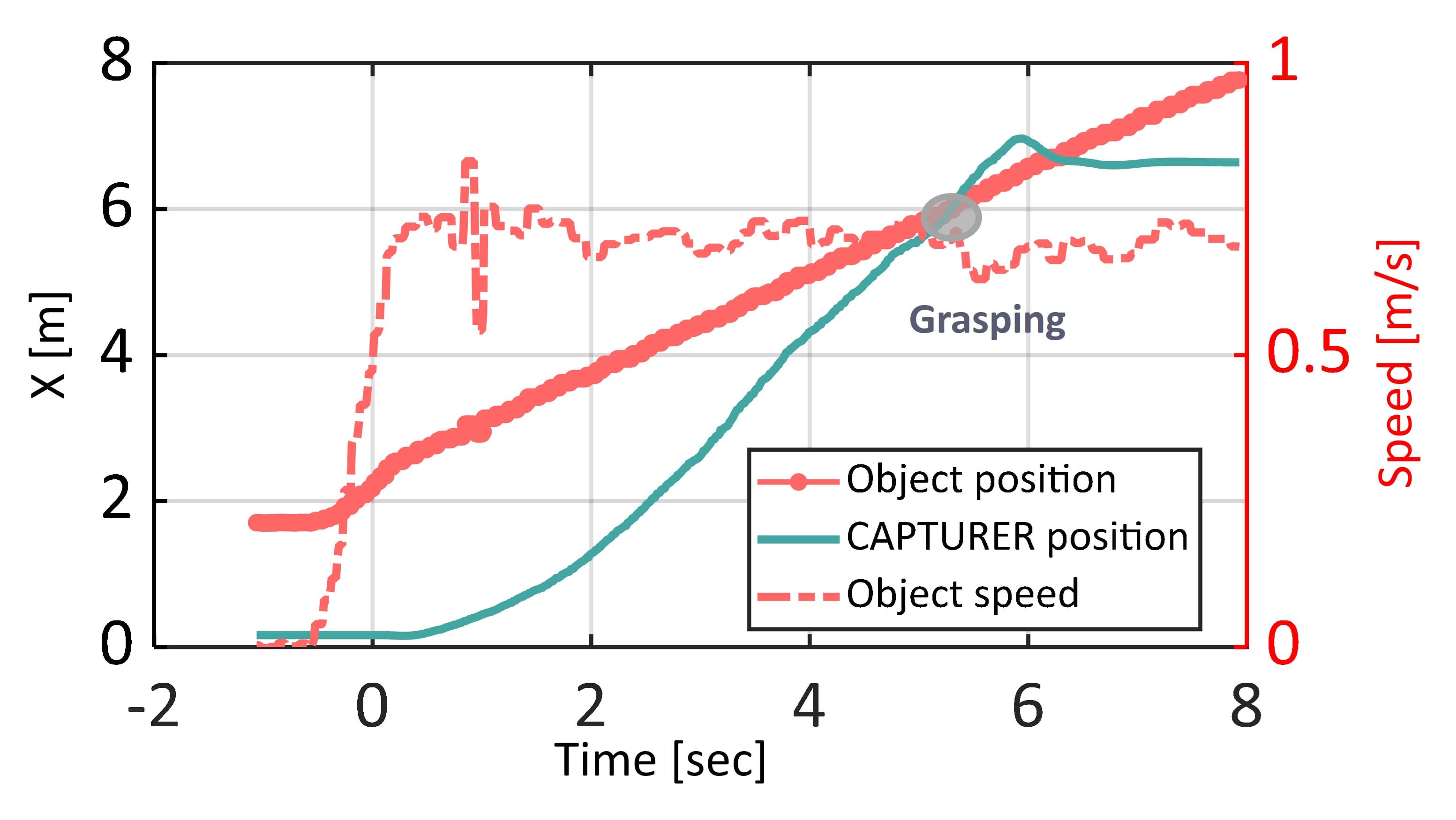}
	\caption{The relative motion between CAPTURER and the target. The instant of starting to close the gripper is denoted by a gray circle. The red dashed line denotes the target velocity.}\label{IJRR_grasp_V4_2}
\end{figure}
\begin{figure}[!tbp]
	\centering
	\includegraphics[width=\linewidth,height=\textheight,keepaspectratio]{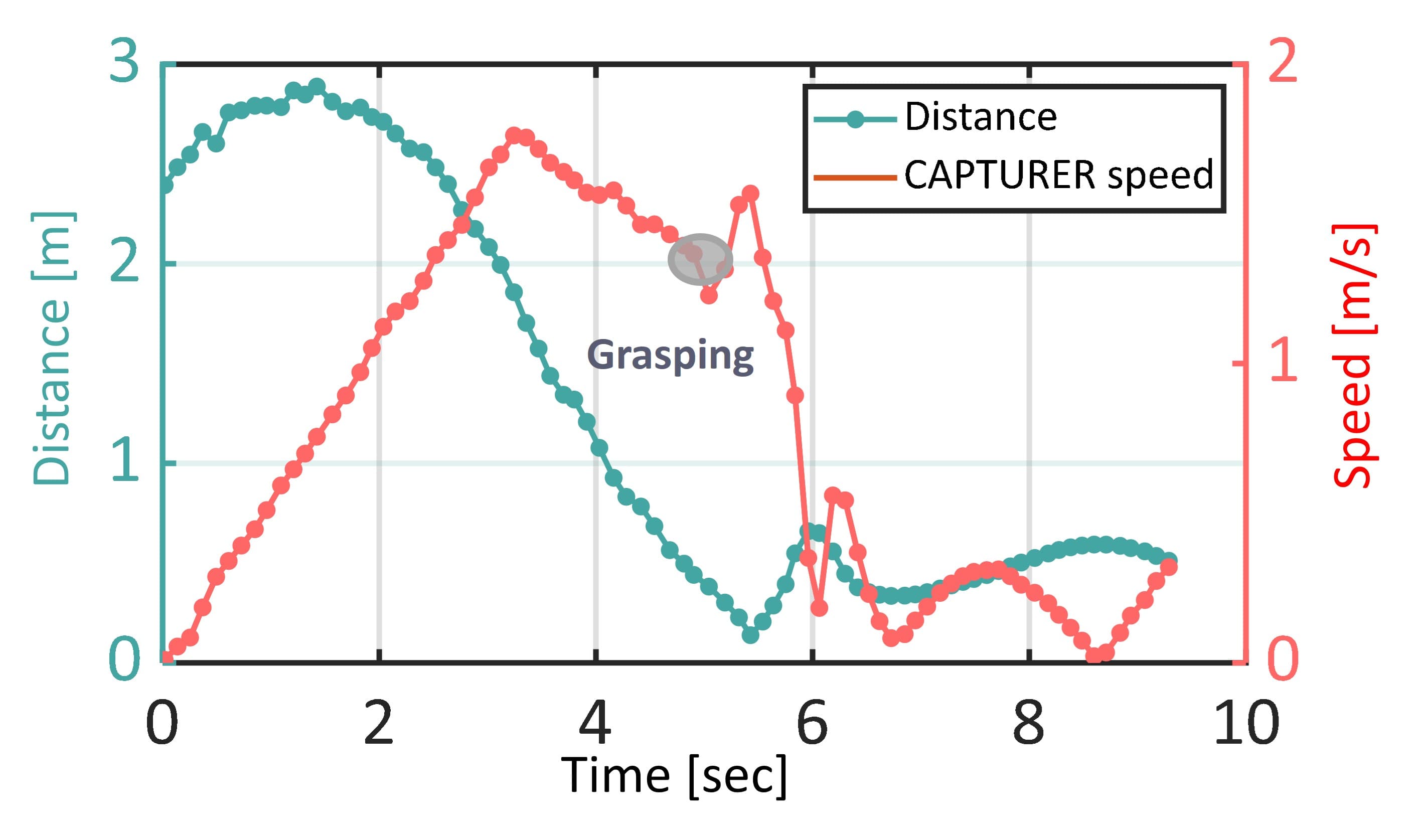}
	\caption{Evolution of the relative distance and motion velocity.}\label{IJRR_grasp_V4_3}
\end{figure}

The aerial manipulation task is further complicated by deviations in the target trajectory during manual operation, as the target does not follow a perfect path. We conduct hardware experiments using two speed settings of the remote controller, corresponding to maximum target velocities of 0.45 m/s and 0.75 m/s. CAPTURER successfully generates online cooperative trajectories and grasps the moving target under both conditions. At higher velocities, the grasping window is further reduced. As the interaction occurs near the operational boundary, CAPTURER executes a reactive vertical retreat immediately after capture. The relative motion between CAPTURER and the target is shown in Fig. \ref{IJRR_grasp_V4_2}. Fig. \ref{IJRR_grasp_V4_3} demonstrates the temporal evolution of the relative distance and motion velocity. Initially, the target vehicle accelerates toward the boundary, briefly widening the gap. CAPTURER then detects the target motion and executes a swift approach to grasp the target. The end-effector tracking error is shown in Fig. \ref{IJRR_endeffector_4}.

Across the conducted experiments, CAPTURER achieves a success rate of 6/6 at both 0.45~m/s and 0.75~m/s. The critical transition threshold $d_0$ is set to 0.40 m to ensure sufficient regulation time before the target reaches the boundary. These results represent an advancement from stationary manipulation to the agile grasping of moving targets, underscoring the feasibility of CAPTURER in complex environments. 

\begin{figure*}[!tp]
	\centering
	\includegraphics[width=\textwidth,height=0.8\textheight,keepaspectratio]{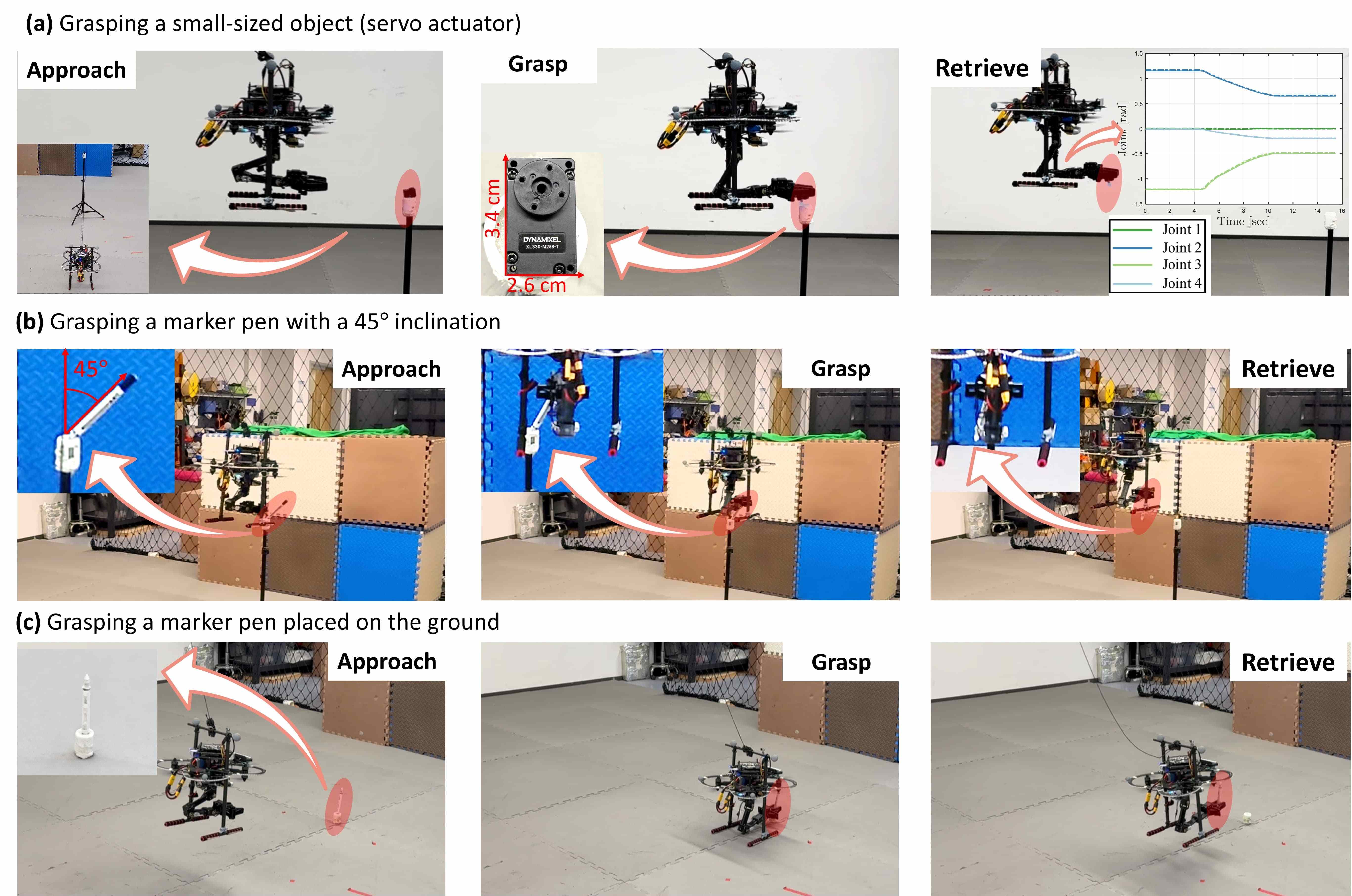}
	\captionsetup{font={stretch=1.0}}
	\caption{Representative scenarios that evaluate object adaptability, angular tolerance, and vertical reachability. The experimental video can be found in the Supplementary Movie S6.}
	\label{IJRR_Capability}
\end{figure*}

\subsection{Capability analysis of CAPTURER}\label{Other_tests}
The employed 5-DoF serial manipulator serves as the primary tool for high-precision aerial interaction, with kinematic redundancy enabling joint-space optimization under practical constraints. Due to the prioritized workspace along the  Y axis, the gripper exhibits inherent geometric limitations when interacting with elongated objects. To systematically evaluate the interaction capability under such constraints, Fig. \ref{IJRR_Capability} illustrates three representative scenarios that evaluate object adaptability, angular tolerance, and vertical reachability.

1) \textbf{Square object grasping}: CAPTURER grasps a servo actuator (2.1 × 2.4 × 3.4 cm), demonstrating the capacity to secure common non-cylindrical components within the gripper’s aperture. During manipulation, the fourth joint dynamically modulates the end-effector pitch angle to maintain vertical alignment and compensate for platform motion. The successful retrieval of functional hardware highlights the potential of CAPTURER for dexterous maintenance and assembly operations.

2) \textbf{Angular tolerance evaluation}: CAPTURER is required to capture a marker pen positioned at a 45$^\circ$ inclination. Despite the object’s longitudinal dimension exceeding the gripper’s nominal aperture, CAPTURER accommodates the orientation mismatch by adjusting the approach direction and manipulator configuration. The experiment demonstrates that CAPTURER can exploit its local dexterity to grasp elongated objects whose orientations are not aligned with the primary grasping direction.

3) \textbf{Vertical reachability assessment}: CAPTURER further demonstrates ground-level target grasping. This experiment confirms the system’s vertical reachability and its capacity to accommodate significant variations in target elevation. The results highlight the robustness of the cooperative planning framework during ground-level interaction, where aerodynamic ground effects and kinematic limits are most pronounced.

%% file: sections/08_discussion.tex
\section{Conclusion and limitations}
In this study, we develop CAPTURER, an autonomous aerial manipulator for centimeter-level aerial grasping across diverse operating conditions. CAPTURER integrates onboard perception, cooperative planning, and disturbance-rejection control within a unified framework. This integration enables the aerial platform and manipulator to perform complementary roles during grasping.

Comprehensive grasping experiments with stationary and moving targets under indoor and outdoor conditions validate the capability of CAPTURER. CAPTURER achieves centimeter-level grasping accuracy and successfully grasps a marker pen only 1.5 cm in diameter at speeds exceeding 1.5 m/s. These results demonstrate that CAPTURER enables precise aerial grasping under different target motions and environmental conditions.

The current perception pipeline relies on prior knowledge of target geometry and fiducial-based detection, which limits its applicability to unknown or unstructured objects. Future work will investigate category-level perception, markerless target tracking, and online scene reconstruction to relax these assumptions. In addition, the current manipulator has a preferred interaction direction and a limited local workspace. Incorporating an omnidirectional aerial platform or a manipulator with improved directional dexterity may further extend the reachable interaction space. Future studies will also examine more general coordination criteria that explicitly consider target motion, manipulator reachability, sensing uncertainty, and grasp timing.

\begin{figure*}[!tp]\centering
	\includegraphics[width=\textwidth,height=.78\textheight,keepaspectratio]{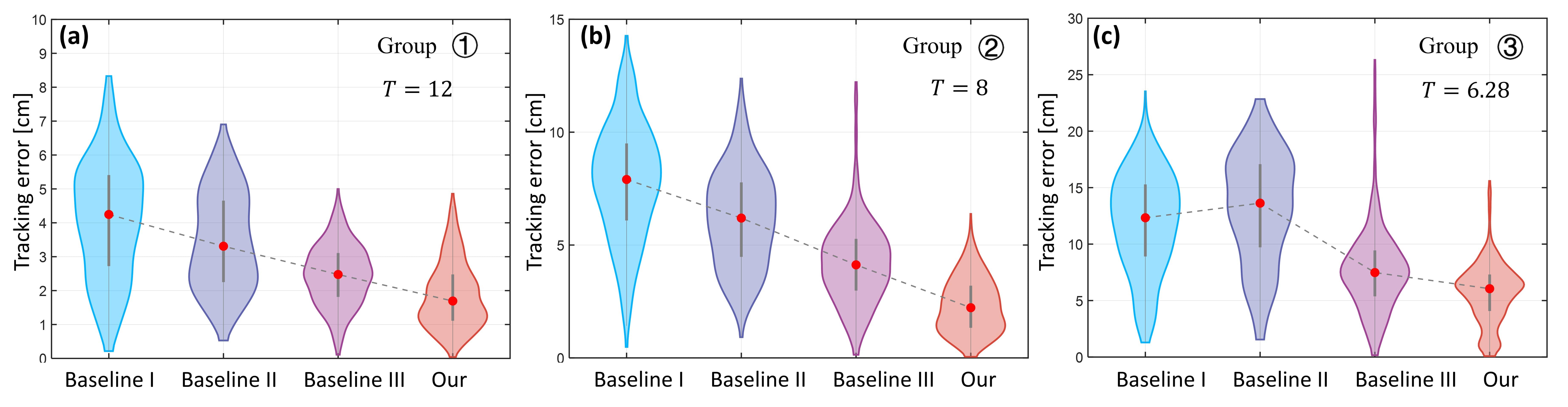}
	\caption{The aerial platform is required to track a four-leaf
		trajectory with different lap times while the manipulator is fixed to the underside of the system. (a-c) Tracking error distribution of all methods in Group \circled{1}, \circled{2}, and \circled{3}.}
	\label{IJRR_Control_1}
\end{figure*}

\begin{figure*}[!tp]\centering
	\includegraphics[width=\textwidth,height=.78\textheight,keepaspectratio]{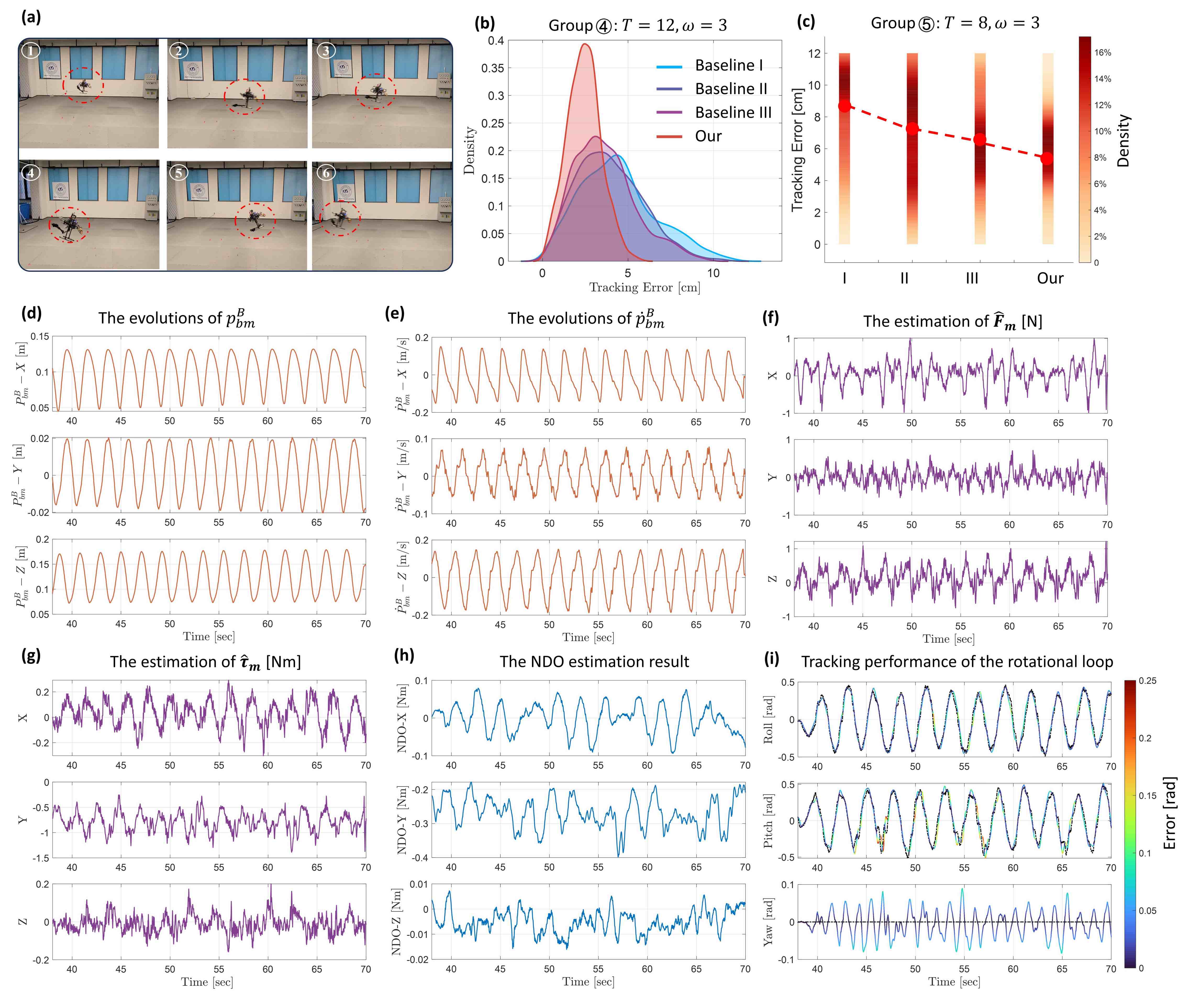}
	\caption{The aerial platform is required to track a four-leaf trajectory with different lap times while the manipulator follows a prescribed trajectory.
		(a) Key snapshots of the flight setup.
		(b) Tracking error distribution of all methods in Group \circled{4}.
		(c) Tracking error distribution of all methods in Group \circled{5}.
		(d)-(e) The evolution of $P_{bm}^{B}$ and $\dot{P}_{bm}^{B}$.
		(f)-(g) The estimation of $\hat{F}_m$ and $\hat{\tau}_m$.
		(h) Estimation result of disturbance torque in the rotational loop.
		(i) The dynamic tracking performance of the rotational loop.}
	\label{IJRR_Control_2}
\end{figure*}

\begin{figure*}[!tp]\centering
	\includegraphics[width=\textwidth,height=.78\textheight,keepaspectratio]{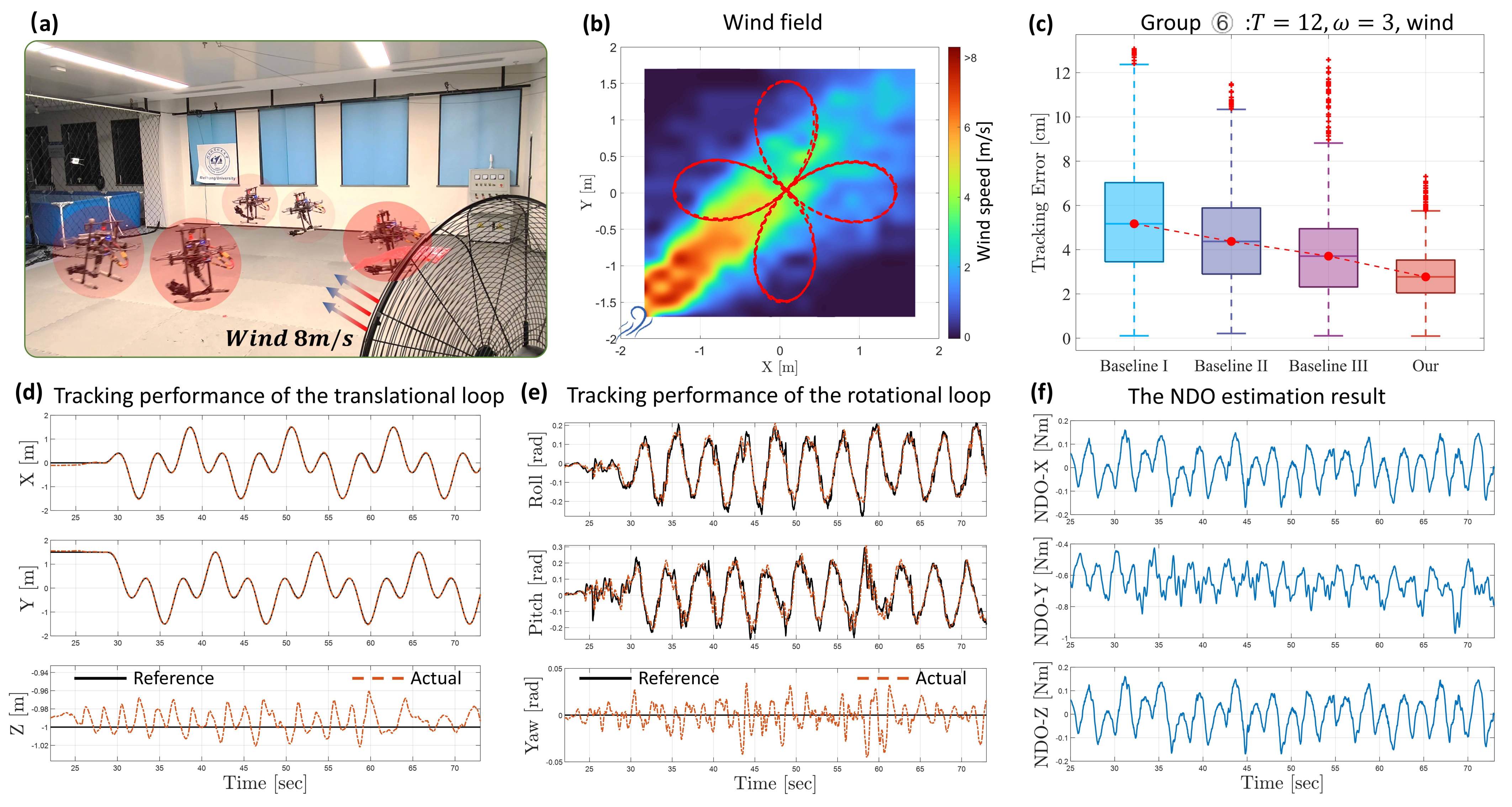}
	\caption{(a) Experimental snapshot. (b) The wind field. (c) Tracking error distribution. (d) The position tracking performance. (e) The attitude tracking performance. (f) Estimation results of disturbance torque in the rotational loop.}
	\label{IJRR_Control_3}
\end{figure*}
\begin{figure}[tbp]
	\centering
	\includegraphics[width=\linewidth,height=\textheight,keepaspectratio]{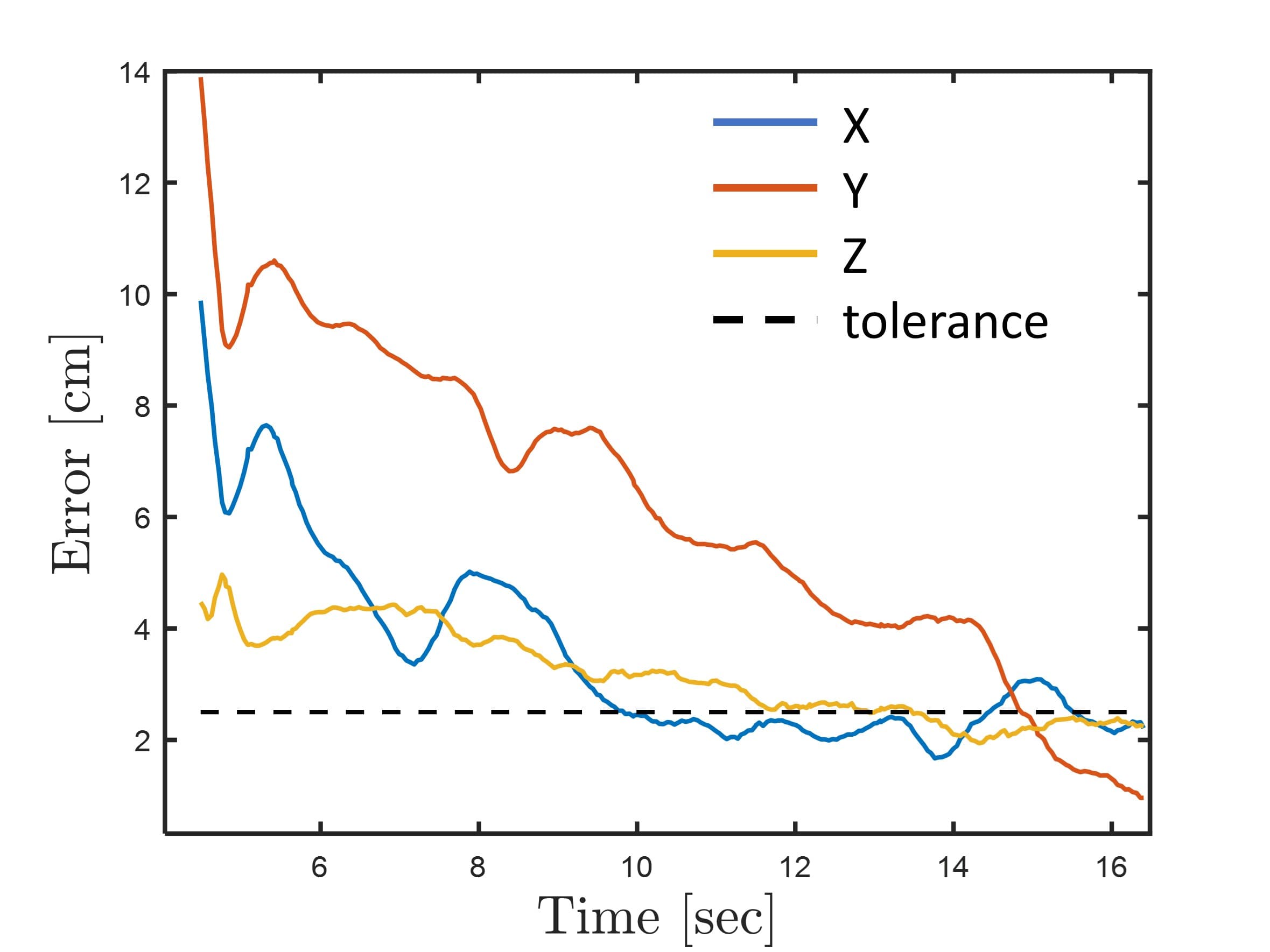}
	\caption{The end-effector position tracking errors in Section \ref{Indoor}.}\label{IJRR_endeffector_1}
\end{figure}
\begin{figure}[tbp]
	\centering
	\includegraphics[width=\linewidth,height=\textheight,keepaspectratio]{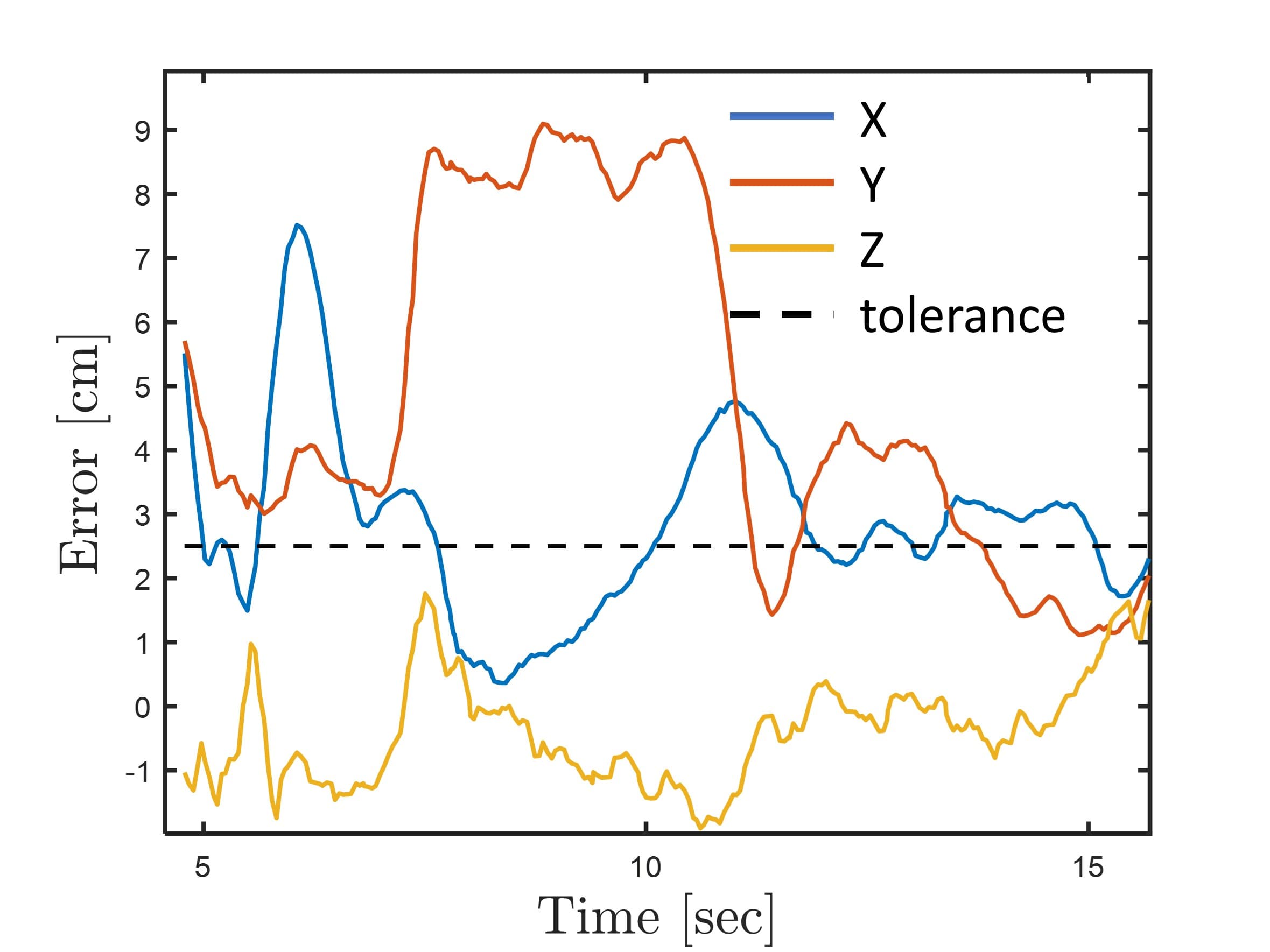}
	\caption{The end-effector position tracking errors in Section \ref{lawn}.}\label{IJRR_endeffector_2}
\end{figure}
\begin{figure}[tbp]
	\centering
	\includegraphics[width=\linewidth,height=\textheight,keepaspectratio]{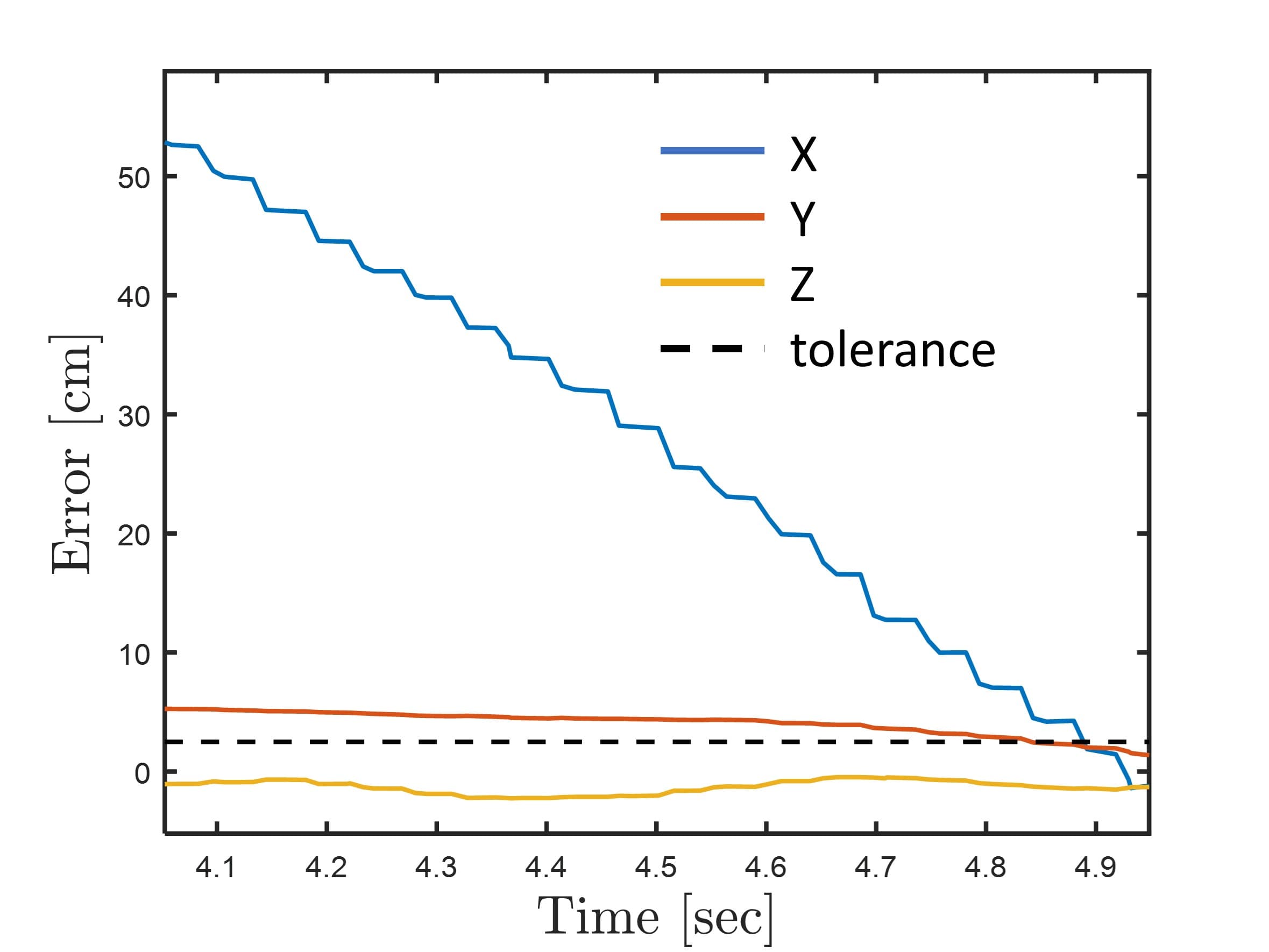}
	\caption{The end-effector position tracking errors in Section \ref{DGSO}.}\label{IJRR_endeffector_3}
\end{figure}
\begin{figure}[tbp]
	\centering
	\includegraphics[width=\linewidth,height=\textheight,keepaspectratio]{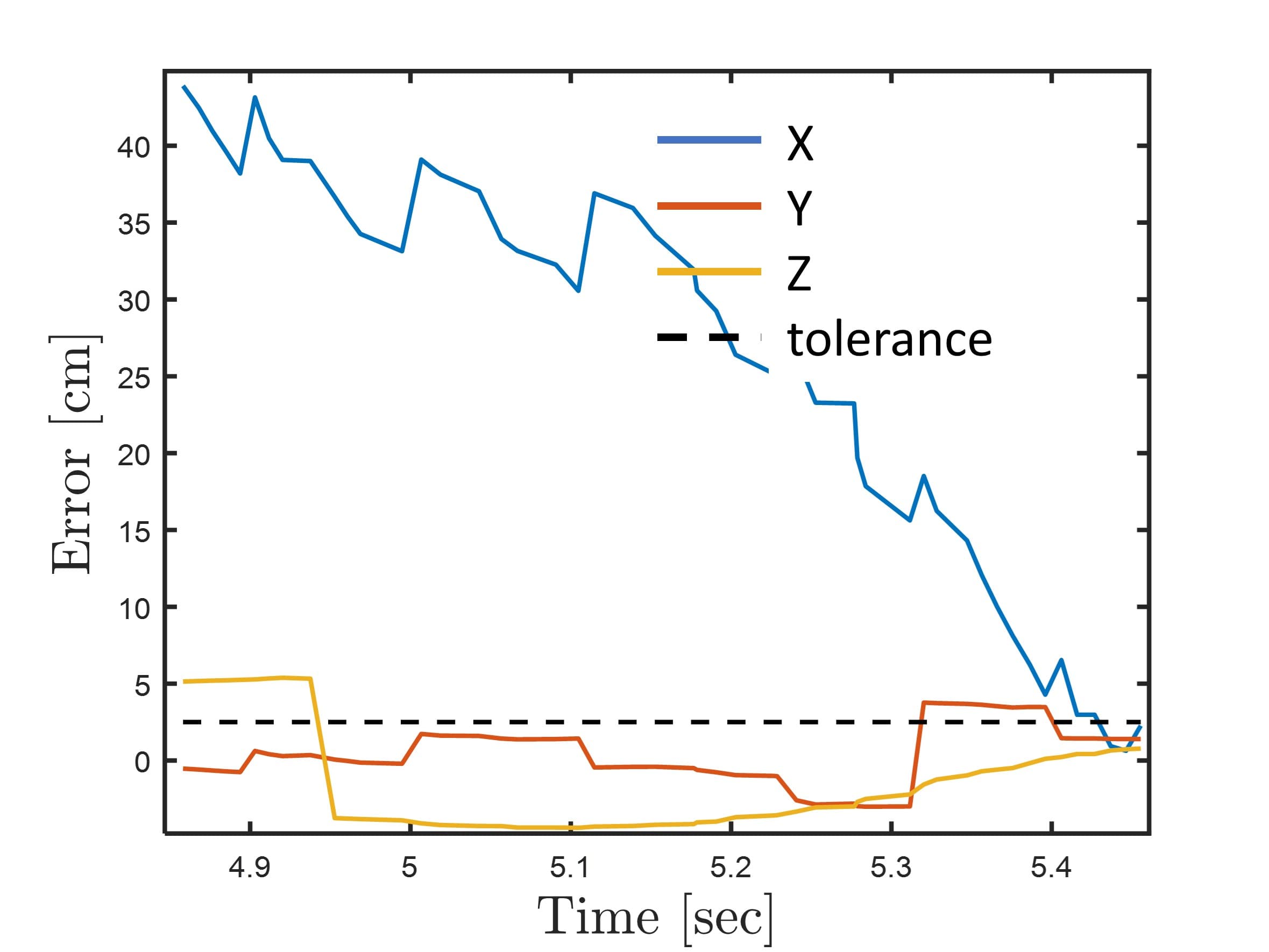}
	\caption{The end-effector position tracking errors in Section \ref{outdoor_moving}.}\label{IJRR_endeffector_4}
\end{figure}
\begin{figure}[tbp]
	\centering
	\includegraphics[width=\linewidth,height=\textheight,keepaspectratio]{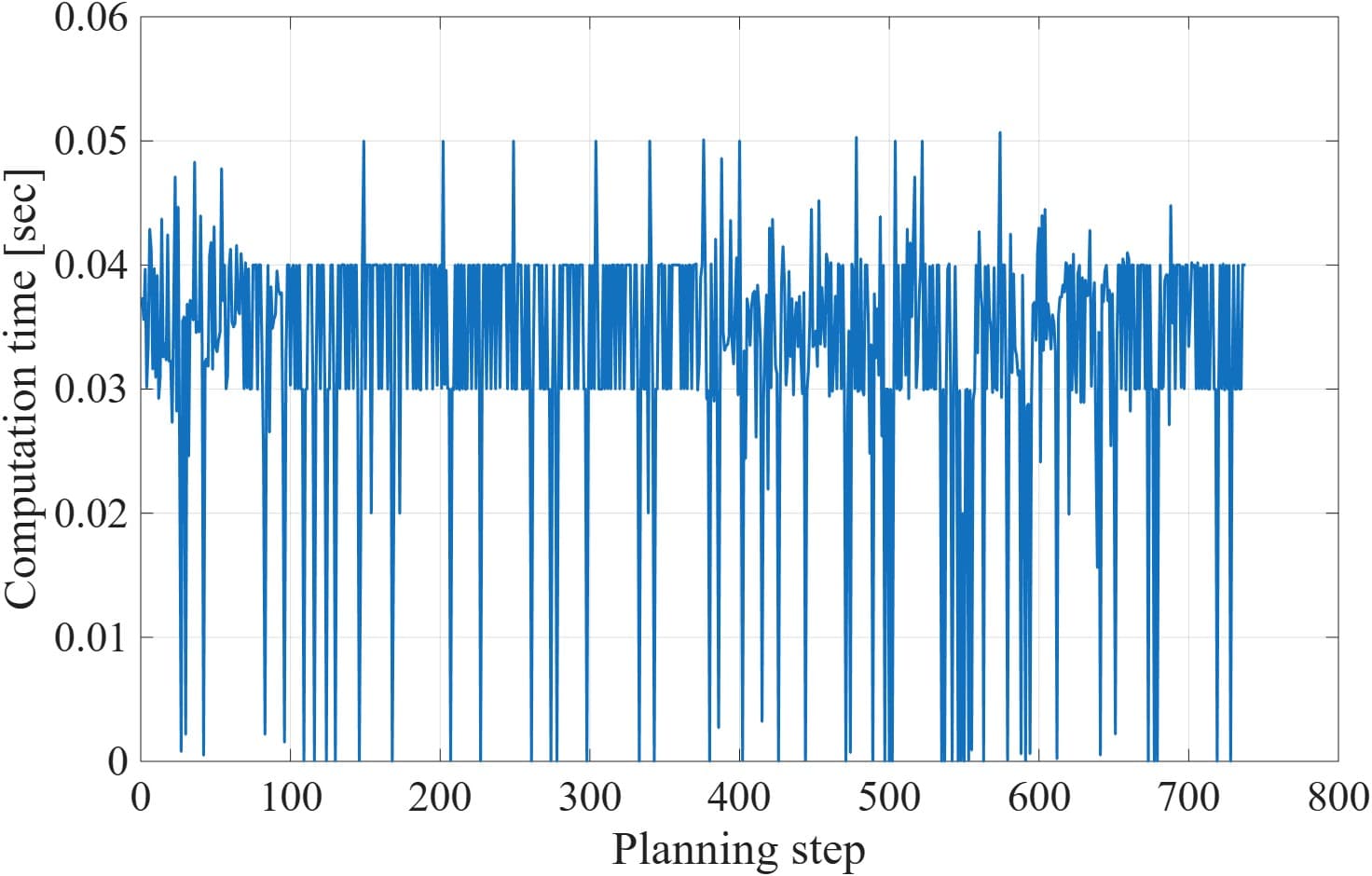}
	\caption{Online optimization time at each replanning step during aerial grasping.}\label{IJRR_planning_time}
\end{figure}

\section{Appendix}

\subsection{Experimental results of Section \ref{CC}}\label{EDCC}
Figs. \ref{IJRR_Control_1}-\ref{IJRR_Control_3} present the detailed experimental results of the dynamic controller in Test I, Test II and Test III, respectively.

\subsection{Definition of propagation matrices in Section \ref{CMP}}\label{DPM}
The propagation matrices are formulated as:
\begin{equation} 
	\begin{split} 
		\bm{A_p}&=\begin{bmatrix} 
			\bm{I_3} & \delta t\bm{I_3} & \bm{0_3} & \bm{0_3} \\ 
			\bm{0_3} & \bm{I_3} & \bm{0_3} & \bm{0_3} \\ 
			\bm{0_3} & \bm{0_3} & \bm{I_3} & \delta t\bm{I_3} \\ 
			\bm{0_3} & \bm{0_3} & \bm{0_3} & \bm{I_3} 
		\end{bmatrix}, \\ 
		\bm{B_p}&=\begin{bmatrix} 
			0.5\delta t^{2}\bm{I_3} & \bm{0_3} \\ 
			\delta t\bm{I_3} & \bm{0_3} \\ 
			\bm{0_3} & 0.5\delta t^{2}\bm{I_3} \\ 
			\bm{0_3} & \delta t\bm{I_3} 
		\end{bmatrix}, 
		\bm{C_p}=\bm{I_{12}}, 
	\end{split} 
	\label{eq:prop_mat} 
\end{equation}

\subsection{Planning parameters of Section \ref{AGADE}}\label{}
Here, \(\otimes\) denotes the Kronecker product and \(\bm{I_N}\in\mathbb{R}^{N\times N}\) is the identity matrix. Figs.  \ref{IJRR_endeffector_1}-\ref{IJRR_endeffector_4} illustrate the tracking error of the end-effector under different grasping conditions. Fig. \ref{IJRR_planning_time} shows the computation time of the proposed cooperative planning method during aerial grasping.

\textbf{Indoor grasping of a stationary object:}\\
$\bm{\Lambda_{1}}
=\bm{I_N}\otimes\operatorname{diag}\{6000,6000,9000\}$,\\
$\bm{\Lambda_{2}^{u}}
=\bm{I_N}\otimes\operatorname{diag}\{50,50,50,800,800,800\}$,\\
$\bm{\Lambda_{2}^{m}}
=\bm{I_N}\otimes\operatorname{diag}\{800,800,800,1,1,1\}$,\\
$\bm{\Lambda_{3}}
=\bm{I_N}\otimes\operatorname{diag}\{30,30,30,15,15,15\}$,\\
$\Delta u_{\max}
=\bm{1_N}\otimes[0.15,0.15,0.15,6,6,6]^{\top}$,\\
$\Delta u_{\min}=-\Delta u_{\max}$,
$d_{0}=0.20$,
$\lambda_{4,u}=50$,
$\lambda_{4,m}=50$,
$\lambda_{5}=300$,
$d_{m}=0.03$,
$\dot{p}_{b\max}=3$,
$\dot{p}_{ed\max}=0.4$,
$\ddot{p}_{b\max}=3$, and
$\ddot{p}_{ed\max}=9$.

\textbf{Grasping a stationary object in an open lawn:}\\
$\bm{\Lambda_{1}}
=\bm{I_N}\otimes\operatorname{diag}\{6000,6000,9000\}$,\\
$\bm{\Lambda_{2}^{u}}
=\bm{I_N}\otimes\operatorname{diag}\{50,50,50,800,800,800\}$,\\
$\bm{\Lambda_{2}^{m}}
=\bm{I_N}\otimes\operatorname{diag}\{800,800,800,1,1,1\}$,\\
$\bm{\Lambda_{3}}
=\bm{I_N}\otimes\operatorname{diag}\{30,30,30,15,15,15\}$,\\
$\Delta u_{\max}
=\bm{1_N}\otimes[0.15,0.15,0.15,6,6,6]^{\top}$,\\
$\Delta u_{\min}=-\Delta u_{\max}$,
$d_{0}=0.20$,
$\lambda_{4,u}=50$,
$\lambda_{4,m}=50$,
$\lambda_{5}=300$,
$d_{m}=0.03$,
$\dot{p}_{b\max}=3$,
$\dot{p}_{ed\max}=0.4$,
$\ddot{p}_{b\max}=3$, and
$\ddot{p}_{ed\max}=9$.

\textbf{Dynamic grasping of a stationary object:}\\
$\bm{\Lambda_{1}}
=\bm{I_N}\otimes\operatorname{diag}\{6000,6000,9000\}$,\\
$\bm{\Lambda_{2}^{u}}
=\bm{I_N}\otimes\operatorname{diag}\{50,50,50,800,800,800\}$,\\
$\bm{\Lambda_{2}^{m}}
=\bm{I_N}\otimes\operatorname{diag}\{50,50,50,1,1,1\}$,\\
$\bm{\Lambda_{3}}
=\bm{I_N}\otimes\operatorname{diag}\{30,30,30,15,15,15\}$,\\
$\Delta u_{\max}
=\bm{1_N}\otimes[0.3,0.3,0.3,9,9,9]^{\top}$,\\
$\Delta u_{\min}=-\Delta u_{\max}$,
$d_{0}=\|v_b\|$,
$\lambda_{4,u}=50$,
$\lambda_{4,m}=50$,
$\lambda_{5}=300$,
$d_{m}=0.03$,
$\dot{p}_{b\max}=3$,
$\dot{p}_{ed\max}=0.5$,
$\ddot{p}_{b\max}=4$, and
$\ddot{p}_{ed\max}=9$.

\textbf{Dynamic grasping of a moving object:}\\
$\bm{\Lambda_{1}}
=\bm{I_N}\otimes\operatorname{diag}\{6000,6000,9000\}$,\\
$\bm{\Lambda_{2}^{u}}
=\bm{I_N}\otimes\operatorname{diag}\{30,30,30,800,800,800\}$,\\
$\bm{\Lambda_{2}^{m}}
=\bm{I_N}\otimes\operatorname{diag}\{30,30,30,1,1,1\}$,\\
$\bm{\Lambda_{3}}
=\bm{I_N}\otimes\operatorname{diag}\{30,30,30,15,15,15\}$,\\
$\Delta u_{\max}
=\bm{1_N}\otimes[0.65,0.65,0.65,9,9,9]^{\top}$,\\
$\Delta u_{\min}=-\Delta u_{\max}$,
$d_{0}=0.40$,
$\lambda_{4,u}=50$,
$\lambda_{4,m}=50$,
$\lambda_{5}=300$,
$d_{m}=0.03$,
$\dot{p}_{b\max}=4$,
$\dot{p}_{ed\max}=0.5$,
$\ddot{p}_{b\max}=5.5$, and
$\ddot{p}_{ed\max}=9$.

%% file: references.bib
@article{BONYANKHAMSEH2018221,
	author = {Hossein Bonyan Khamseh and Farrokh Janabi-Sharifi and Abdelkader Abdessameud},
	title = {Aerial manipulation—A literature survey},
	journal = {Robotics and Autonomous Systems},
	year = {2018},
	volume = {107},
	pages = {221--235},
}

@article{9462539,
	author = {Ollero, Anibal and Tognon, Marco and Suarez, Alejandro and Lee, Dongjun and Franchi, Antonio},
	title = {Past, present, and future of aerial robotic manipulators},
	journal = {IEEE Transactions on Robotics},
	year = {2022},
	volume = {38},
	number = {1},
	pages = {626--645},
}

@article{nature,
	author = {Zhang, Ketao and Chermprayong, Pisak and Xiao, Feng and Tzoumanikas, Dimos and Dams, Barrie and Kay, Sebastian and Kocer, Basaran Bahadir and Burns, Alec and Orr, Lachlan and Alhinai, Talib and Choi, Christopher and Darekar, Durgesh Dattatray and Li, Wenbin and Hirschmann, Steven and Soana, Valentina and Awang Ngah, Shamsiah and Grillot, Clément and Sareh, Sina and Choubey, Ashutosh and Margheri, Laura and Pawar, Vijay M. and Ball, Richard J. and Williams, Chris and Shepherd, Paul and Leutenegger, Stefan and Stuart-Smith, Robert and Kovac, Mirko},
	title = {Aerial additive manufacturing with multiple autonomous robots},
	journal = {Nature},
	year = {2022},
	volume = {609},
	pages = {709--717},
}

@article{Proximal,
	author = {Cao, Huazi and Shen, Jiahao and Zhang, Yin and Fu, Zheng and Liu, Cunjia and Sun, Sihao and Zhao, Shiyu},
	title = {Proximal cooperative aerial manipulation with vertically stacked drones},
	journal = {Nature},
	year = {2025},
	volume = {646},
	pages = {576--583},
}

@article{10083097,
	author = {Cao, Huazi and Li, Yongqi and Liu, Cunjia and Zhao, Shiyu},
	title = {{ESO}-based robust and high-precision tracking control for aerial manipulation},
	journal = {IEEE Transactions on Automation Science and Engineering},
	year = {2024},
	volume = {21},
	number = {2},
	pages = {2139--2155},
}

@article{10339889,
	author = {Wang, Meng and Chen, Zeshuai and Guo, Kexin and Yu, Xiang and Zhang, Youmin and Guo, Lei and Wang, Wei},
	title = {Millimeter-level pick and peg-in-hole task achieved by aerial manipulator},
	journal = {IEEE Transactions on Robotics},
	year = {2024},
	volume = {40},
	pages = {1242--1260},
}

@article{8059875,
	author = {Orsag, Matko and Korpela, Christopher and Bogdan, Stjepan and Oh, Paul},
	title = {Dexterous Aerial Robots—{M}obile Manipulation Using Unmanned Aerial Systems},
	journal = {IEEE Transactions on Robotics},
	year = {2017},
	volume = {33},
	number = {6},
	pages = {1453--1466},
}

@inproceedings{04,
	author = {Garimella, Gowtham and Kobilarov, Marin},
	title = {Towards model-predictive control for aerial pick-and-place},
	booktitle = {2015 IEEE International Conference on Robotics and Automation (ICRA)},
	year = {2015},
	pages = {4692--4697},
}

@article{9834073,
	author = {Chen, Tony G. and Hoffmann, Kenneth A. W. and Low, Jun En and Nagami, Keiko and Lentink, David and Cutkosky, Mark R.},
	title = {Aerial grasping and the velocity sufficiency region},
	journal = {IEEE Robotics and Automation Letters},
	year = {2022},
	volume = {7},
	number = {4},
	pages = {10009--10016},
}

@inproceedings{9981668,
	author = {Appius, Aurel X. and Bauer, Erik and Blöchlinger, Marc and Kalra, Aashi and Oberson, Robin and Raayatsanati, Arman and Strauch, Pascal and Suresh, Sarath and von Salis, Marco and Katzschmann, Robert K.},
	title = {{RAPTOR}: Rapid aerial pickup and transport of objects by robots},
	booktitle = {2022 IEEE/RSJ International Conference on Intelligent Robots and Systems (IROS)},
	year = {2022},
	pages = {349--355},
}

@article{10224341,
	author = {Luo, Wei and Chen, Jingshan and Ebel, Henrik and Eberhard, Peter},
	title = {Time-optimal handover trajectory planning for aerial manipulators based on discrete mechanics and complementarity constraints},
	journal = {IEEE Transactions on Robotics},
	year = {2023},
	volume = {39},
	number = {6},
	pages = {4332--4349},
}

@inproceedings{7989751,
	author = {Seo, Hoseong and Kim, Suseong and Kim, H. Jin},
	title = {Aerial grasping of cylindrical object using visual servoing based on stochastic model predictive control},
	booktitle = {2017 IEEE International Conference on Robotics and Automation (ICRA)},
	year = {2017},
	pages = {6362--6368},
}

@article{9789094,
	author = {Lai, Ningbin and Chen, Yanjie and Liang, Jiacheng and He, Bingwei and Zhong, Hang and Zhang, Hui and Wang, Yaonan},
	title = {Image dynamics-based visual servo control for unmanned aerial manipulator with a virtual camera},
	journal = {IEEE/ASME Transactions on Mechatronics},
	year = {2022},
	volume = {27},
	number = {6},
	pages = {5264--5274},
}

@article{9866833,
	author = {Bucki, Nathan and Tang, Jerry and Mueller, Mark W.},
	title = {Design and control of a midair-reconfigurable quadcopter using unactuated hinges},
	journal = {IEEE Transactions on Robotics},
	year = {2023},
	volume = {39},
	number = {1},
	pages = {539--557},
}

@inproceedings{8594330,
	author = {Zhao, Na and Luo, Yudong and Deng, Hongbin and Shen, Yantao and Xu, Hao},
	title = {The deformable quad-rotor enabled and wasp-pedal-carrying inspired aerial gripper},
	booktitle = {2018 IEEE/RSJ International Conference on Intelligent Robots and Systems (IROS)},
	year = {2018},
	pages = {1--9},
}

@inproceedings{9196576,
	author = {Shi, Fan and Zhao, Moju and Murooka, Masaki and Okada, Kei and Inaba, Masayuki},
	title = {Aerial regrasping: Pivoting with transformable multilink aerial robot},
	booktitle = {2020 IEEE International Conference on Robotics and Automation (ICRA)},
	year = {2020},
	pages = {200--207},
}

@article{10495172,
	author = {Xu, Mengxin and De, Qixin and Yu, Dafang and Hu, An and Liu, Zhe and Wang, Hesheng},
	title = {Biomimetic morphing quadrotor inspired by eagle claw for dynamic grasping},
	journal = {IEEE Transactions on Robotics},
	year = {2024},
	volume = {40},
	pages = {2513--2528},
}

@article{10044987,
	author = {Xu, Mengxin and Huang, Siyuan and He, Ruokun and Yu, Dafang and Wang, Hesheng},
	title = {Aerial shooting manipulator for distant grasping},
	journal = {IEEE Robotics and Automation Letters},
	year = {2023},
	volume = {8},
	number = {4},
	pages = {1991--1998},
}

@article{Champion_level,
	author = {Elia Kaufmann and Leonard Bauersfeld and Antonio Loquercio and Matthias Muller and Vladlen Koltun and Davide Scaramuzza},
	title = {Champion-level drone racing using deep reinforcement learning},
	journal = {Nature},
	year = {2023},
	volume = {620},
	pages = {982--987},
}

@article{abh1221,
	author = {Philipp Foehn and Angel Romero and Davide Scaramuzza},
	title = {Time-optimal planning for quadrotor waypoint flight},
	journal = {Science Robotics},
	year = {2021},
	volume = {6},
	number = {56},
	pages = {eabh1221},
}

@inproceedings{6608869,
	author = {Arleo, G. and Caccavale, F. and Muscio, G. and Pierri, F.},
	title = {Control of quadrotor aerial vehicles equipped with a robotic arm},
	booktitle = {2013 Mediterranean Conference on Control and Automation (MED)},
	year = {2013},
	pages = {1174--1180},
}

@article{Behavioral,
	author = {K. Baizid and G. Giglio and F. Pierri and M. A. Trujillo and G. Antonelli and F. Caccavale and A. Viguria and S. Chiaverini and A. Ollero},
	title = {Behavioral control of unmanned aerial vehicle manipulator systems},
	journal = {Autonomous Robots},
	year = {2017},
	volume = {41},
	pages = {1203--1220},
}

@article{7962217,
	author = {Muscio, G. and Pierri, F. and Trujillo, M. A. and Cataldi, E. and Antonelli, G. and Caccavale, F. and Viguria, A. and Chiaverini, S. and Ollero, A.},
	title = {Coordinated control of aerial robotic manipulators: Theory and experiments},
	journal = {IEEE Transactions on Control Systems Technology},
	year = {2018},
	volume = {26},
	number = {4},
	pages = {1406--1413},
}

@article{LAI2022102724,
	author = {Ningbin Lai and Yanjie Chen and Jiacheng Liang and Bingwei He and Hang Zhong and Yaonan Wang},
	title = {An onboard-eye-to-hand visual servo and task coordination control for aerial manipulator based on a spherical model},
	journal = {Mechatronics},
	year = {2022},
	volume = {82},
	pages = {102724},
}

@inproceedings{7017672,
	author = {Santamaria-Navarro, Angel and Lippiello, Vincenzo and Andrade-Cetto, Juan},
	title = {Task priority control for aerial manipulation},
	booktitle = {2014 IEEE International Symposium on Safety, Security, and Rescue Robotics (SSRR)},
	year = {2014},
	pages = {1--6},
}

@inproceedings{8796272,
	author = {Imanberdiyev, Nursultan and Monica, Josephine and Kayacan, Erdal},
	title = {A multi-task velocity-based redundancy resolution strategy for unmanned aerial manipulators},
	booktitle = {2019 18th European Control Conference (ECC)},
	year = {2019},
	pages = {1130--1135},
}

@article{9194145,
	author = {Bordalba, Ricard and Ros, Lluís and Porta, Josep M.},
	title = {A randomized kinodynamic planner for closed-chain robotic systems},
	journal = {IEEE Transactions on Robotics},
	year = {2021},
	volume = {37},
	number = {1},
	pages = {99--115},
}

@article{8283709,
	author = {Tognon, Marco and Cataldi, Elisabetta and Chavez, Hermes Amadeus Tello and Antonelli, Gianluca and Cortés, Juan and Franchi, Antonio},
	title = {Control-aware motion planning for task-constrained aerial manipulation},
	journal = {IEEE Robotics and Automation Letters},
	year = {2018},
	volume = {3},
	number = {3},
	pages = {2478--2484},
}

@article{IJARS,
	author = {Alvaro Caballero and Manuel Bejar and Angel Rodriguez-Castaño and Anibal Ollero},
	title = {Motion planning with dynamics awareness for long reach manipulation in aerial robotic systems with two arms},
	journal = {International Journal of Advanced Robotic Systems},
	year = {2018},
	volume = {15},
	number = {3},
	pages = {1729881418770525},
}

@article{CAO2022107607,
	author = {Huazi Cao and Yu Wu and Lixin Wang},
	title = {Adaptive {NN} motion control and predictive coordinate planning for aerial manipulators},
	journal = {Aerospace Science and Technology},
	year = {2022},
	volume = {126},
	pages = {107607},
}

@book{shabana2020dynamics,
	author = {Shabana, Ahmed A.},
	title = {Dynamics of Multibody Systems},
	publisher = {Cambridge University Press},
	year = {2020},
}

@article{7762762,
	author = {Rossi, Roberto and Santamaria-Navarro, Angel and Andrade-Cetto, Juan and Rocco, Paolo},
	title = {Trajectory generation for unmanned aerial manipulators through quadratic programming},
	journal = {IEEE Robotics and Automation Letters},
	year = {2017},
	volume = {2},
	number = {2},
	pages = {389--396},
}

@inproceedings{10156288,
	author = {Chen, Rui and Liu, Qianyuan and Chen, Zeshuai and Guo, Kexin and Yu, Xiang},
	title = {Online trajectory generation for aerial manipulator subject to multi-tasks and inequality constraints},
	booktitle = {2023 International Conference on Unmanned Aircraft Systems (ICUAS)},
	year = {2023},
	pages = {933--939},
}

@article{9919418,
	author = {Huang, Yanjiang and Ke, Jianhong and Zhang, Xianmin and Ota, Jun},
	title = {Dynamic parameter identification of serial robots using a hybrid approach},
	journal = {IEEE Transactions on Robotics},
	year = {2023},
	volume = {39},
	number = {2},
	pages = {1607--1621},
}

@article{8772145,
	author = {Gaz, Claudio and Cognetti, Marco and Oliva, Alexander and Robuffo Giordano, Paolo and De Luca, Alessandro},
	title = {Dynamic identification of the franka emika panda robot with retrieval of feasible parameters using penalty-based optimization},
	journal = {IEEE Robotics and Automation Letters},
	year = {2019},
	volume = {4},
	number = {4},
	pages = {4147--4154},
}

@article{SEEBER2021109858,
	author = {Richard Seeber and Hernan Haimovich and Martin Horn and Leonid M. Fridman and Hernán {De Battista}},
	title = {Robust exact differentiators with predefined convergence time},
	journal = {Automatica},
	year = {2021},
	volume = {134},
	pages = {109858},
}

@inproceedings{7158852,
	author = {Bellicoso, Carmine Dario and Buonocore, Luca Rosario and Lippiello, Vincenzo and Siciliano, Bruno},
	title = {Design, modeling and control of a 5-DoF light-weight robot arm for aerial manipulation},
	booktitle = {2015 Mediterranean Conference on Control and Automation (MED)},
	year = {2015},
	pages = {853--858},
}

@inproceedings{7353623,
	author = {Jimenez-Cano, A.E and Braga, J. and Heredia, G. and Ollero, A.},
	title = {Aerial manipulator for structure inspection by contact from the underside},
	booktitle = {2015 IEEE/RSJ International Conference on Intelligent Robots and Systems (IROS)},
	year = {2015},
	pages = {1879--1884},
}

@article{8924899,
	author = {Zhang, Guangyu and He, Yuqing and Dai, Bo and Gu, Feng and Han, Jianda and Liu, Guangjun},
	title = {Robust Control of an Aerial Manipulator Based on a Variable Inertia Parameters Model},
	journal = {IEEE Transactions on Industrial Electronics},
	year = {2020},
	volume = {67},
	number = {11},
	pages = {9515--9525},
}

@article{9507086,
	author = {Bodie, Karen and Tognon, Marco and Siegwart, Roland},
	title = {Dynamic End Effector Tracking With an Omnidirectional Parallel Aerial Manipulator},
	journal = {IEEE Robotics and Automation Letters},
	year = {2021},
	volume = {6},
	number = {4},
	pages = {8165--8172},
}

@article{EMAMI2021106573,
	author = {Seyyed Ali Emami and Afshin Banazadeh},
	title = {Simultaneous trajectory tracking and aerial manipulation using a multi-stage model predictive control},
	journal = {Aerospace Science and Technology},
	year = {2021},
	volume = {112},
	pages = {106573},
}

@article{10478625,
	author = {Wang, Haokun and Li, Haojia and Zhou, Boyu and Gao, Fei and Shen, Shaojie},
	title = {Impact-Aware Planning and Control for Aerial Robots With Suspended Payloads},
	journal = {IEEE Transactions on Robotics},
	year = {2024},
	volume = {40},
	pages = {2478--2497},
}

@article{9345436,
	author = {Liu, Yen-Chen and Huang, Chi-Yu},
	title = {{DDPG}-Based Adaptive Robust Tracking Control for Aerial Manipulators With Decoupling Approach},
	journal = {IEEE Transactions on Cybernetics},
	year = {2022},
	volume = {52},
	number = {8},
	pages = {8258--8271},
}

@article{8047452,
	author = {Kim, Suseong and Choi, Seungwon and Kim, Hyeonggeun and Shin, Jongho and Shim, Hyungbo and Kim, H. Jin},
	title = {Robust control of an equipment-added multirotor using disturbance observer},
	journal = {IEEE Transactions on Control Systems Technology},
	year = {2018},
	volume = {26},
	number = {4},
	pages = {1524--1531},
}

@article{9309359,
	author = {Lee, Dongjae and Seo, Hoseong and Jang, Inkyu and Lee, Seung Jae and Kim, H. Jin},
	title = {Aerial manipulator pushing a movable structure using a {DOB}-based robust controller},
	journal = {IEEE Robotics and Automation Letters},
	year = {2021},
	volume = {6},
	number = {2},
	pages = {723--730},
}

@article{10547187,
	author = {Wang, Meng and Lyu, Shangke and Liu, Qianyuan and Yang, Ziqi and Guo, Kexin and Yu, Xiang},
	title = {Precise end-effector control for an aerial manipulator under composite disturbances: Theory and experiments},
	journal = {IEEE Transactions on Automation Science and Engineering},
	year = {2025},
	volume = {22},
	pages = {4006--4021},
}

@article{VRBA2022103970,
	author = {Matou\v{s} Vrba and Yurii Stasinchuk and Tom\'{a}\v{s} B\'{a}\v{c}a and Vojt\v{e}ch Spurn\'{y} and Mat\v{e}j Petrlík and Daniel He\v{r}t and David \v{Z}aitlík and Martin Saska},
	title = {Autonomous capture of agile flying objects using {UAVs}: The {MBZIRC} 2020 challenge},
	journal = {Robotics and Autonomous Systems},
	year = {2022},
	volume = {149},
	pages = {103970},
}

@article{limit,
	author = {Yunlong Song and Angel Romero and Matthias Müller and Vladlen Koltun and Davide Scaramuzza},
	title = {Reaching the limit in autonomous racing: Optimal control versus reinforcement learning},
	journal = {Science Robotics},
	year = {2023},
	volume = {8},
	number = {82},
	pages = {eadg1462},
}

@inproceedings{7759617,
	author = {Wang, John and Olson, Edwin},
	title = {{AprilTag} 2: Efficient and robust fiducial detection},
	booktitle = {2016 IEEE/RSJ International Conference on Intelligent Robots and Systems (IROS)},
	year = {2016},
	pages = {4193--4198},
}

@article{631234,
	author = {Swevers, J. and Ganseman, C. and Tukel, D.B. and de Schutter, J. and Van Brussel, H.},
	title = {Optimal robot excitation and identification},
	journal = {IEEE Transactions on Robotics and Automation},
	year = {1997},
	volume = {13},
	number = {5},
	pages = {730--740},
}

@inproceedings{5717652,
	author = {Lee, Taeyoung and Leok, Melvin and McClamroch, N. Harris},
	title = {Geometric tracking control of a quadrotor {UAV} on {SE(3)}},
	booktitle = {49th IEEE Conference on Decision and Control (CDC)},
	year = {2010},
	pages = {5420--5425},
}

@book{kevin2017,
	author = {Lynch, Kevin M. and Park, Frank C.},
	title = {Modern Robotics: Mechanics, Planning, and Control},
	publisher = {Cambridge University Press},
	year = {2017},
}

@article{9697912,
	author = {Xu, Wei and Cai, Yixi and He, Dongjiao and Lin, Jiarong and Zhang, Fu},
	title = {{FAST-LIO2}: Fast direct {LiDAR}-inertial odometry},
	journal = {IEEE Transactions on Robotics},
	year = {2022},
	volume = {38},
	number = {4},
	pages = {2053--2073},
}

@article{9615117,
	author = {Lyu, Pin and Bai, Shiyu and Lai, Jizhou and Wang, Bingqing and Sun, Xin and Huang, Kai},
	title = {Optimal time difference-Based {TDCP-GPS/IMU} navigation using graph optimization},
	journal = {IEEE Transactions on Instrumentation and Measurement},
	year = {2021},
	volume = {70},
	pages = {1--10},
}

@article{Ryll20196DIC,
	author = {Markus Ryll and Giuseppe Muscio and Francesco Pierri and Elisabetta Cataldi and Gianluca Antonelli and Fabrizio Caccavale and Davide Bicego and Antonio Franchi},
	title = {6{D} interaction control with aerial robots: The flying end-effector paradigm},
	journal = {The International Journal of Robotics Research},
	year = {2019},
	volume = {38},
	pages = {1045--1062},
}

@article{9781617,
	author = {Hu, An and Xu, Mengxin and Wang, Hesheng and Castañeda, Herman},
	title = {Vision-based impedance control of an aerial manipulator using a nonlinear observer},
	journal = {IEEE Transactions on Automation Science and Engineering},
	year = {2023},
	volume = {20},
	number = {2},
	pages = {1441--1451},
}

@article{10598388,
	author = {Duong, Thai and Altawaitan, Abdullah and Stanley, Jason and Atanasov, Nikolay},
	title = {{Port-Hamiltonian} Neural {ODE} Networks on Lie Groups for Robot Dynamics Learning and Control},
	journal = {IEEE Transactions on Robotics},
	year = {2024},
	volume = {40},
	pages = {3695--3715},
}

@inproceedings{NEURIPS2018_69386f6b,
	author = {Chen, Ricky T. Q. and Rubanova, Yulia and Bettencourt, Jesse and Duvenaud, David K},
	title = {Neural Ordinary Differential Equations},
	booktitle = {Advances in Neural Information Processing Systems},
	year = {2018},
	volume = {31},
	pages = {6571--6583},
	editor = {S. Bengio and H. Wallach and H. Larochelle and K. Grauman and N. Cesa-Bianchi and R. Garnett},
}

@article{npjrobotics,
	author = {Samuel Ubellacker and Aaron Ray and James M. Bern and Jared Strader and Luca Carlone},
	title = {High-speed aerial grasping using a soft drone with onboard perception},
	journal = {npj Robotics},
	year = {2024},
	volume = {2},
	pages = {5},
}

@article{10486851,
	author = {Cao, Huazi and Shen, Jiahao and Liu, Cunjia and Zhu, Bo and Zhao, Shiyu},
	title = {Motion planning for aerial pick-and-place with geometric feasibility constraints},
	journal = {IEEE Transactions on Automation Science and Engineering},
	year = {2025},
	volume = {22},
	pages = {2577--2594},
}

@article{10722859,
	author = {Li, Zhan and Li, Hai and Xu, Quman and Yu, Xinghu and Basin, Michael V.},
	title = {Coupling disturbance modeling and compensation for aerial manipulator in highly dynamic motion},
	journal = {IEEE Transactions on Cybernetics},
	year = {2025},
	volume = {55},
	number = {1},
	pages = {124--135},
}

@article{7747236,
	author = {Cadena, Cesar and Carlone, Luca and Carrillo, Henry and Latif, Yasir and Scaramuzza, Davide and Neira, José and Reid, Ian and Leonard, John J.},
	title = {Past, present, and future of simultaneous localization and mapping: Toward the robust-perception age},
	journal = {IEEE Transactions on Robotics},
	year = {2016},
	volume = {32},
	number = {6},
	pages = {1309--1332},
}

@article{Swarm,
	author = {Xin Zhou and Xiangyong Wen and Zhepei Wang and Yuman Gao and Haojia Li and Qianhao Wang and Tiankai Yang and Haojian Lu and Yanjun Cao and Chao Xu and Fei Gao},
	title = {Swarm of micro flying robots in the wild},
	journal = {Science Robotics},
	year = {2022},
	volume = {7},
	number = {66},
	pages = {eabm5954},
}

@article{10328688,
	author = {Gao, Yuman and Ji, Jialin and Wang, Qianhao and Jin, Rui and Lin, Yi and Shang, Zhimeng and Cao, Yanjun and Shen, Shaojie and Xu, Chao and Gao, Fei},
	title = {Adaptive tracking and perching for quadrotor in dynamic scenarios},
	journal = {IEEE Transactions on Robotics},
	year = {2024},
	volume = {40},
	pages = {499--519},
}

@article{8931787,
	author = {Zhong, Hang and Miao, Zhiqiang and Wang, Yaonan and Mao, Jianxu and Li, Ling and Zhang, Hui and Chen, Yanjie and Fierro, Rafael},
	title = {A practical visual servo control for aerial manipulation using a spherical projection model},
	journal = {IEEE Transactions on Industrial Electronics},
	year = {2020},
	volume = {67},
	number = {12},
	pages = {10564--10574},
}

@article{9102390,
	author = {Gao, Fei and Wang, Luqi and Zhou, Boyu and Zhou, Xin and Pan, Jie and Shen, Shaojie},
	title = {Teach-repeat-replan: A complete and robust system for aggressive flight in complex environments},
	journal = {IEEE Transactions on Robotics},
	year = {2020},
	volume = {36},
	number = {5},
	pages = {1526--1545},
}

@inproceedings{Hopf,
	author = {Watterson, Michael and Kumar, Vijay},
	title = {Control of Quadrotors Using the {Hopf} Fibration on {SO}(3)},
	booktitle = {2019 International Symposium of Robotics Research (ISRR)},
	year = {2019},
	pages = {199--215},
}

@article{MA2022212,
	author = {Le Ma and Yiming Yan and Zhiwei Li and Jie Liu},
	title = {Neural-embedded learning control for fully-actuated flying platform of aerial manipulation system},
	journal = {Neurocomputing},
	year = {2022},
	volume = {482},
	pages = {212--223},
}

@article{scirobotics_Ladar,
	author = {Ren, Yunfan and Zhu, Fangcheng and Lu, Guozheng and Cai, Yixi and Yin, Longji and Kong, Fanze and Lin, Jiarong and Chen, Nan and Zhang, Fu},
	title = {Safety-assured high-speed navigation for {MAVs}},
	journal = {Science Robotics},
	year = {2025},
	volume = {10},
	number = {98},
	pages = {eado6187},
}

@article{10943237,
	author = {Zhang, Zhaopeng and Yu, Hai and Chai, Yi and Yang, Zhichao and Liang, Xiao and Fang, Yongchun and Han, Jianda},
	title = {An End-Effector-Oriented Coupled Motion Planning Method for Aerial Manipulators in Constrained Environments},
	journal = {IEEE/ASME Transactions on Mechatronics},
	year = {2025},
	volume = {30},
	number = {6},
	pages = {6027--6037},
}

@article{SR_PHRI,
	author = {Emanuele Aucone and Stefano Mintchev},
	title = {Embodied aerial physical interaction: Combining body and brain for robust interaction with unstructured environments},
	journal = {Science Robotics},
	year = {2025},
	volume = {10},
	number = {102},
	pages = {eads0200},
}

@article{nerual_fly,
	author = {Michael O’Connell and Guanya Shi and Xichen Shi and Kamyar Azizzadenesheli and Anima Anandkumar and Yisong Yue and Soon-Jo Chung},
	title = {{Neural-Fly} enables rapid learning for agile flight in strong winds},
	journal = {Science Robotics},
	year = {2022},
	volume = {7},
	number = {66},
	pages = {eabm6597},
}

@article{ado6251,
	author = {Yusuf Kaya and Lachlan Orr and Basaran Kocer and Vijay Pawar and Robert Smith and Mirko Kovač},
	title = {Aerial additive manufacturing: Toward on-site building construction with aerial robots},
	journal = {Science Robotics},
	year = {2025},
	volume = {10},
	number = {101},
	pages = {eado6251},
}

@inproceedings{pioneer,
	author = {Quentin Lindsey and Daniel Mellinger and Vijay Kumar},
	title = {Construction of cubic structures with quadrotor teams},
	booktitle = {Proceedings of Robotics: Science and Systems},
	year = {2011},
}

@inproceedings{5980314,
	author = {Pounds, Paul and Bersak, Daniel and Dollar, Aaron},
	title = {Grasping from the air: Hovering capture and load stability},
	booktitle = {2011 IEEE International Conference on Robotics and Automation (ICRA)},
	year = {2011},
	pages = {2491--2498},
}

@article{Kumar2014,
	author = {Justin Thomas and Giuseppe Loianno and Joseph Polin and Koushil Sreenath and Vijay Kumar},
	title = {Toward autonomous avian-inspired grasping for micro aerial vehicles},
	journal = {Bioinspiration \& Biomimetics},
	year = {2014},
	volume = {9},
	number = {2},
	pages = {025010},
}

@article{rodriguezramos,
	author = {Rodriguez-Ramos, Alejandro and Alvarez-Fernandez, Adrian and Bavle, Hriday and Rodriguez-Vazquez, Javier and Lu, Liang and Fernandez-Cortizas, Miguel and Suarez Fernandez, Ramon A. and Rodelgo, Alberto and Santos, Carlos and Molina, Martin and Merino, Luis and Caballero, Fernando and Campoy, Pascual},
	title = {Autonomous aerial robot for high-speed search and intercept applications},
	journal = {Field Robotics},
	year = {2022},
	volume = {2},
	number = {1},
	pages = {1320--1350},
}

@inproceedings{8619390,
	author = {Bisheban, Mahdis and Lee, Taeyoung},
	title = {Geometric adaptive control for a quadrotor {UAV} with wind disturbance rejection},
	booktitle = {2018 IEEE Conference on Decision and Control (CDC)},
	year = {2018},
	pages = {2816--2821},
}

@article{Kim_IJRR,
	author = {Lee, Dongjae and Kim, Byeongjun and Kim, Hyoun Jin},
	title = {Autonomous aerial manipulation at arbitrary pose in {SE(3)} with robust control and whole-body planning},
	journal = {The International Journal of Robotics Research},
	year = {2026},
	volume = {45},
	number = {7},
	pages = {1021--1045},
}

@article{9929266,
	author = {Mishra, Hrishik and Garofalo, Gianluca and Giordano, Alessandro Massimo and De Stefano, Marco and Ott, Christian and Kugi, Andreas},
	title = {Reduced {Euler-Lagrange} Equations of Floating-Base Robots: Computation, Properties, \& Applications},
	journal = {IEEE Transactions on Robotics},
	year = {2023},
	volume = {39},
	number = {2},
	pages = {1439--1457},
}

@misc{quaternionkinematicserrorstatekalman,
	author = {Joan Solà},
	title = {Quaternion kinematics for the error-state {Kalman} filter},
	year = {2017},
	note = {arXiv:1711.02508},
}

@inproceedings{11618145,
	author = {Wang, Meng and Chen, Zeshuai and Guo, Yuxin and Guo, Kexin and Yu, Xiang},
	title = {A Physics-Informed Optimization Control Allocation Strategy for Coaxial Multirotor},
	booktitle = {2026 IEEE 20th International Conference on Control and Automation (ICCA)},
	year = {2026},
	pages = {1406--1411},
}

@article{IJRR_Jia,
	author = {Jindou Jia and Kexin Guo and Yuyang Wang and Sicheng Zhou and Jiayi Zhang and Yuhang Liu and Xiang Yu and Yang Shi and Lei Guo},
	title = {{FORESEER}: Recognize and utilize uncertainties by integrating data-based learning and symbolic feedback},
	journal = {The International Journal of Robotics Research},
	year = {2025},
	note = {OnlineFirst},
}

@article{IJRR_ChenWang,
	author = {Chen Wang and Kaiyi Ji and Junyi Geng and Zhongqiang Ren and Taimeng Fu and Fan Yang and Yifan Guo and Haonan He and Xiangyu Chen and Zitong Zhan and Qiwei Du and Shaoshu Su and Bowen Li and Yuheng Qiu and Yi Du and Qihang Li and Yifan Yang and Xiao Lin and Zhipeng Zhao},
	title = {Imperative learning: A self-supervised neuro-symbolic learning framework for robot autonomy},
	journal = {The International Journal of Robotics Research},
	year = {2025},
	note = {OnlineFirst},
}

@article{IJRR_FuZhang,
	author = {Guozheng Lu and Yixi Cai and Nan Chen and Fanze Kong and Yunfan Ren and Fu Zhang},
	title = {Trajectory generation and tracking control for aggressive tail-sitter flights},
	journal = {The International Journal of Robotics Research},
	year = {2024},
	volume = {43},
	number = {3},
	pages = {241--280},
}

@article{IJRR_Zhao,
	author = {Zian Ning and Yin Zhang and Jianan Li and Zhang Chen and Shiyu Zhao},
	title = {A bearing-angle approach for unknown target motion analysis based on visual measurements},
	journal = {The International Journal of Robotics Research},
	year = {2024},
	volume = {43},
	number = {8},
	pages = {1228--1249},
}

@article{IJRR_kummar,
	author = {Daniel Mellinger and Nathan Michael and Vijay Kumar},
	title = {Trajectory generation and control for precise aggressive maneuvers with quadrotors},
	journal = {The International Journal of Robotics Research},
	year = {2012},
	volume = {31},
	number = {5},
	pages = {664--674},
}

@article{jia2026physics,
	author = {Jia, Jindou and Han, Shixuan and Wang, Meng and Li, Gen and Yang, Zihan and Zhou, Sicheng and Guo, Kexin and Yang, Jianfei and Yu, Xiang and Wang, Wei and Guo, Lei},
	title = {Physics filtering favors the generalization of robot learning},
	journal = {npj Robotics},
	year = {2026},
	volume = {4},
	pages = {48},
}
